%% file: sample.tex
\documentclass[twoside,11pt]{article}

\usepackage{jmlr2e_backup}

\usepackage[utf8]{inputenc} 
\usepackage[T1]{fontenc} 
\usepackage{hyperref}  
\usepackage{url}   
\usepackage{booktabs}  
\usepackage{amsfonts}  
\usepackage{nicefrac}  
\usepackage{microtype}  
\usepackage{amsmath}
\usepackage{float}
\usepackage{graphicx}
\usepackage{subcaption}
\usepackage{amsthm}
\usepackage[capitalise,noabbrev]{cleveref}
\usepackage{verbatim}
\usepackage{csquotes}
\usepackage{amssymb}
\usepackage{float}
\usepackage{graphicx}
\usepackage{subcaption}
\usepackage{enumerate}
\usepackage{enumitem}
\usepackage{natbib}
\usepackage{bbm}
\usepackage{bigints}
\usepackage{soul}
\usepackage{pgffor}
\usepackage{tikz}

\usepackage{algorithm}
\usepackage{algorithmic}
\usepackage{cases}
\usepackage{ifthen}
\theoremstyle{plain}

\newtheorem{theorem}{Theorem}[section]

\newtheorem{definition}[theorem]{Definition}

\makeatletter
\renewcommand*{\ALG@name}{Algorithm}
\makeatother

\usepackage[colorinlistoftodos]{todonotes}

\newcommand{\pb}{P_{\textrm{B}}(f|S)}
\newcommand{\psgd}{P_{\textrm{SGD}}(f|S)}
\newcommand{\popt}{P_{\textrm{OPT}}(f|S)}
\newcommand{\padam}{P_{\textrm{Adam}}(f|S)}
\newcommand{\padagrad}{P_{\textrm{Adagrad}}(f|S)}
\newcommand{\padadelta}{P_{\textrm{Adadelta}}(f|S)}
\newcommand{\prmsprop}{P_{\textrm{RMSprop}}(f|S)}
\newcommand{\pntk}{P_{\textrm{NTK}}(f|S)}
\newcommand{\eg}{\langle \epsilon_G \rangle}

\firstpageno{1}

\begin{document}

\title{Is SGD a Bayesian sampler? Well, almost.}

\author{\name Chris Mingard \email christopher.mingard@chem.ox.ac.uk \\
  \addr Department of Chemistry\\
  University of Oxford
  \AND
  \name Guillermo Valle-P\'erez \email guillermo.valle@dtc.ox.ac.uk \\
  \addr Department of Physics\\
  University of Oxford
  \AND
  \name Joar Skalse \email joar.skalse@cs.ox.ac.uk \\
  \addr Department of Computer Science\\
  University of Oxford
  \AND
  \name Ard A. Louis \email ard.louis@physics.ox.ac.uk \\
  \addr Department of Physics\\
  University of Oxford
}

\editor{}

\maketitle
\begin{abstract}

Deep neural networks (DNNs) generalise remarkably well in the overparameterised regime, suggesting a strong inductive bias towards functions with low generalisation error. 
We empirically investigate this bias 
by calculating, for a range of architectures and datasets, the probability $\psgd$ that an overparameterised DNN, trained with stochastic gradient descent (SGD) 
or one of its variants, 
converges on a function $f$ consistent with a training set $S$. We also use Gaussian processes to estimate the Bayesian posterior probability $\pb$ that the DNN expresses $f$ upon random sampling of its parameters, conditioned on $S$. 

Our main findings are that $\psgd$ correlates remarkably well with $\pb$ and that $\pb$ is strongly biased towards low-error and low complexity functions. These results imply that strong inductive bias in the parameter-function map (which determines $\pb$), rather than a special property of SGD, is the primary explanation for why DNNs generalise so well in the overparameterised regime. 

While our results suggest that the Bayesian posterior $\pb$ is the first order determinant of $\psgd$, there remain second order differences that are sensitive to hyperparameter tuning. A function probability picture, based on $\psgd$ and/or $\pb$, can shed light on the way that variations in architecture or hyperparameter settings such as batch size, learning rate, and optimiser choice, affect DNN performance. 
\end{abstract}

\section{Introduction}\label{sec1:intro}

While deep neural networks (DNNs) have revolutionised modern machine learning~\citep{lecun2015deep,schmidhuber2015deep}, a solid theoretical understanding of why they work so well is still lacking. One surprising property is that they typically perform best in the overparameterised regime, with many more parameters than data points. Standard learning theory approaches~\citep{shalev2014understanding}, based for example on model capacity, suggest that such highly expressive~\citep{cybenko1989approximation,hornik1991approximation,hanin2019universal} DNNs should heavily over-fit in this regime, and therefore not generalise at all.

Stochastic gradient descent (SGD)~\citep{bottou2018optimization} is one of the key technical innovations allowing large DNNs to be efficiently trained in the highly overparameterised regime. 
In supervised learning, SGD allows the user to efficiently find sets of parameters that lead to zero training error. The power of SGD as an optimiser for DNNs was demonstrated in an influential paper~\citep{zhang2016understanding}, which showed that zero training error solutions for CIFAR-10 image data with randomised labels can be found with a relatively moderate increase in computational effort over that needed for a correctly labelled dataset. These experiments also framed the conundrum of generalisation in the overparameterised regime as follows: Given that DNNs can memorise randomly labelled image datasets, which leads to poor generalisation, why do they behave so differently on correctly labelled datasets and select for functions that generalise well? The solution to this conundrum must be that SGD-trained DNNs have an inductive bias towards functions that generalise well (on structured data).

The possibility that SGD is not just good for optimisation, but is also a key source of inductive bias, has generated an extensive literature. One major theme concerns the effect of SGD on the flatness of the minima found, typically expressed in terms of eigenvalues of a local Hessian or related measures. A link between better generalisation and flatter minima has been widely reported \citep{hochreiter1997flat,keskar2016large,jastrzebski2018finding,wu2017towards,zhang2018energy,wei2019noise}, but see also~\citep{dinh2017sharp}. 
Theoretical work on SGD has also generated a large and sophisticated literature. For example, 
in~\citep{soudry2018implicit} it was demonstrated that SGD finds the max-margin solution in unregularised logistic regression, whilst it was shown in~\citep{brutzkus2017sgd} that overparameterised DNNs trained with SGD avoid over-fitting on linearly separable data. Recently, \citep{NIPS2019_8847} proved agnostic generalisation bounds of SGD-trained neural networks. Other recent work~\citep{poggio2020complexity} suggests that gradient descent performs a hidden regularisation in normalised weights, but a different analysis suggests that such implicit regularisation may well be very hard to prove in a more general setting for SGD~\citep{dauber2020can}. 
Overall, while SGD and its related algorithms are excellent optimisers, there is as yet no consensus on what inductive bias SGD provides for DNNs. For a more detailed discussion of this SGD-related literature see \cref{rel:optimiser_trained}.


An alternative approach is to consider the inductive properties of \emph{random neural networks}, that is \textit{untrained} DNNs with weights sampled from a (typically i.i.d.) distribution. Recent theoretical and empirical work~\citep{valle2018deep,de2018random,mingard2019neural} suggests that the (prior) probability $P(f)$ that an untrained DNN outputs a function $f$ upon random sampling of its parameters (typically the weights and biases) is strongly biased towards ``simple'' functions with low Kolmogorov complexity (see also~\cref{subapp:simplicity_bias}). A widely held assumption is that such simple hypotheses will generalise well -- think Occam's razor. Indeed, many processes modelled by DNNs are simple~\citep{lin2017does,goldt2019modelling,spigler2019asymptotic}. For more on these topics see~\cref{subapp:simplicity_bias} and~\cref{rel:simplicity_bias_generalisation}.

If the inductive bias towards simplicity described above for untrained networks is preserved throughout training, then this could help explain the DNN generalisation conundrum. Again, there is an extensive literature relevant to this topic. For example, a number of papers~\citep{poole2016exponential,lee2018deep,valle2018deep,yang2019scaling,mingard2019neural,cohen2019learning,wilson2020bayesian} employ arguments on heuristic grounds that the bias in untrained random neural networks could be used to study the inductive bias of optimiser-trained DNNs. Optimiser-trained DNNs have also been directly compared to their Bayesian counterparts (c.f.\ \cref{sec:heuristic_arguments,rel:opt_vs_bayes} for more detailed discussions). In an important development, \citet{lee2017deep,matthews2018gaussian,novak2018bayesian} used the Gaussian process (GP) approximation to Bayesian DNNs, which is exact in the limit of infinite width, and found that the generalisation performance of Bayesian DNNs and SGD-trained DNNs was relatively similar for standard deep learning datasets such as CIFAR-10, though \cite{wenzel2020good} found more significant differences when using Monte Carlo to approximate finite-width Bayesian DNNs. Others have used either Monte Carlo methods~\citep{stephan2017stochastic} or the GP approximation~\citep{matthewssample,g.2018gaussian,lee2019wide,wilson2020bayesian} to examine how similar the Bayesian posterior is to the sampling distribution of SGD (whether in parameter or function space), albeit on relatively low dimensional systems compared to conventional DNNs. 

In this paper 
we perform extensive computations, for a series of standard DNNs and datasets, of the probability $\psgd$ that a DNN trained with SGD (or one of its variants) to zero error on training set $S$, converges on a function $f$. We then compare these results to the Bayesian posterior probability $\pb$, for these same functions, conditioned on achieving zero training error on $S$. 

The \textbf{main question} we explore here is: \textit{How similar is $\pb$ to $\psgd$}? If the two are significantly different, then we may conclude that SGD provides an important source of inductive bias. If the two are broadly similar over a wide range of architectures, datasets, and optimisers, 
then the inductive bias is primarily determined by the prior $P(f)$ of the untrained DNN.

\subsection{Main results summary}\label{sec:our_contribution}

We carried out extensive sampling experiments to estimate 
$\psgd$. Functions are distinguished by the way they classify elements on a 
test set $E$.  We use the Gaussian process (GP) approximation to estimate $\pb$ for the same systems. Our main findings are: 

\textbf{(1)} $\psgd \approx \pb$ for a range of architectures, including FCNs, CNNs and LSTMs, applied to datasets such as MNIST, Fashion-MNIST, an IMDb movie review database and an ionosphere dataset. This agreement also holds for variants of SGD, including Adam~\citep{kingma2014adam}, Adagrad~\citep{duchi2011adaptive}, Adadelta~\citep{zeiler2012adadelta} and RMSprop~\citep{tieleman2012lecture}. 

\textbf{(2)} The $\pb$ of functions $f$ that achieve zero-error on the training set $S$ can vary over hundreds of orders of magnitude, with a strong bias towards a set of low generalisation/low complexity functions. This tiny fraction of high probability functions also dominate what is found by DNNs trained with SGD. It is striking that even within this subset of functions, $\psgd$ and $\pb$ correlate so well. Our empirical results suggest that, \emph{for DNNs with large bias in $\pb$}, SGD behaves \emph{to first order} like a Bayesian optimiser and is therefore exponentially biased towards simple functions with better generalisation. Thus, SGD is not itself the primary source of inductive bias for DNNs.

\textbf{(3)} A function-based picture can also be fruitful for illustrating \textit{second order} effects where an optimiser-trained DNN differs from the Bayesian prediction. For example, training an FCN with different optimisers (OPT) such as Adam, Adagrad, Adadelta and RMSprop on MNIST generates slight but measurable variations in the distributions of $P_{\textrm{OPT}}(f|S)$. Such information can be used to analyse differences in performance.
For instance, we find that changing batch size affects $\padam$ but, as was found for generalisation error~\citep{keskar2016large,goyal2017accurate,hoffer2017train,smith2017don}, this effect can be compensated by changes in learning rate. 
Architecture changes can also be examined in this picture. For example, adding max-pooling to a CNN trained with Adam on Fashion-MNIST increases both $\pb$ and $\padam$ for the lowest-error function $f$ found.

\section{Preliminaries}\label{sec:preliminary_definitions}

We first introduce a key definition from~\citep{valle2018deep} needed to specify $P(f)$ and $P_B(f|S)$.
\begin{definition}[Parameter-function map]\label{def:PFM}
Consider a parameterised supervised model, and let the input space be $\mathcal{X}$ and the output space be $\mathcal{Y}$. The space of functions the model can express is a set $\mathcal{F} \subseteq \mathcal{Y}^{\mathcal{X}}$. If the model has some number of parameters, taking values within a set $\Theta \subseteq \mathbb{R}^p$, then the parameter-function map $\mathcal{M}$ is defined by
\begin{align*}
 \begin{aligned}
  \mathcal{M}:\Theta &\to \mathcal{F} \\
  \theta &\mapsto f_{\theta}
 \end{aligned}
 \end{align*}
 where $f_\theta$ is the function corresponding to parameters $\theta \in \Theta$.
\end{definition}

The function space $\mathcal{F}$ of a DNN $\mathcal{N}$ could in principle be considered to be the entire space of functions that $\mathcal{N}$ can express on the input vector space $\mathcal{X}$, but it could also be taken to be the set of partial functions $\mathcal{N}$ can express on some subset of $\mathcal{X}$. For example, $\mathcal{F}$ could be taken to be the set of possible classifications of images in MNIST. In this paper we always take $\mathcal{F}$ to be the set of possible outputs of $\mathcal{N}$ for the instances in some dataset.

\subsection{The Bayesian prior probability, $P(f)$}
\label{prelim:prior}

Given a distribution $P_{\textrm{par}}(\theta)$ over the parameters,
we define the $P(f)$ over functions as
\begin{align}\label{eqn:Pf_vol}
P(f)=\bigintssss \mathbbm{1}[\mathcal{M}(\theta)=f]P_{\textrm{par}}(\theta)d\theta,
\end{align}
where $\mathbbm{1}$ is an indicator function ($1$ if its argument is true, and $0$ otherwise). This is the probability that the model expresses $f$ upon random sampling of parameters over a parameter initialisation distribution $P_{\textrm{par}}(\theta)$, which 
is typically taken to have a simple form such as a (truncated) Gaussian. $P(f)$ can also be interpreted as the probability that the DNN expresses $f$ upon initialisation before an optimisation process. It was shown in~\citep{valle2018deep} that the exact form of $P_{\textrm{par}}(\theta)$ (for reasonable choices) does not affect $P(f)$ much (at least for ReLU networks). 
If we condition on functions that obtain zero generalisation error on a dataset $S$, then the procedure above can also be used to generate the posterior
$P_B(f|S)$ which we describe next.

\subsection{The Bayesian posterior probability, $\pb$}\label{sec:PB}

Here, we describe the Bayesian formalism we use, and show how bias in the prior affects the posterior. Consider a supervised learning problem with training data $S$ corresponding to the exact values of the function which we wish to infer (i.e.\ no noise). This formulation corresponds to a $0$-$1$ \textit{likelihood} $P(S|f)$, indicating whether the data is consistent with the function. Formally, if $S=\{(x_i,y_i)\}_{i=1}^m$ corresponds to the set of training pairs, then we let
\begin{equation*}
\label{01likelihood}
  P(S|f) = \begin{cases} 1\textrm{ if } \forall i,\;\; f(x_i)=y_i \\
       0\textrm{ otherwise }.
   \end{cases}
\end{equation*}
Note that in our calculations, this quantity is technically $P(S|f;\{x_i\})$, but we denote it as $P(S|f)$ to simplify notation. We will use a similar convention throughout, whereby the input points are (implicitly) conditioned over. We then assume the prior $P(f)$ corresponds to the one defined in \cref{prelim:prior}. Bayesian inference then assigns a \emph{Bayesian posterior probability} $\pb$ to each $f$ by conditioning on the data according to Bayes rule
\begin{equation}
\label{eqn:bayesian_update}
 \pb := \frac{P(S|f)P(f)}{P(S)},
\end{equation}
where $P(S)$ is also called the \emph{marginal likelihood} or \emph{Bayesian evidence}. It is the total probability of all functions compatible with the training set. For discrete functions, $P(S) = \sum_f P(S|f) P(f)=\sum_{f\in C(S)} P(f)$, with $C(S)$ the set of all functions compatible with the training set. For a fixed training set, all the variation in $P_B(f|S)$ for $f \in C(S)$ comes from the prior $P(f)$ of the untrained network since $P(S)$ is constant. Thus, \textit{the bias in the prior is essentially translated over to the posterior}. 



Thus, $\pb$ is the distribution over functions that would be obtained by randomly sampling parameters according to $P_{\textrm{par}}(\theta)$ and selecting only those that are compatible with $S$.

\subsection{The optimiser probability, $\popt$}

But DNNs are not normally trained by randomly sampling parameters: They are trained by an optimiser. The probability that the optimiser OPT (e.g.\ SGD) finds a function $f$ with zero error on $S$ can be defined as:
\begin{align}\label{eqn:basin_of_attraction}
\popt :=\bigintssss \mathbbm{1}[\mathcal{M}(\theta_f)=f] P_{\textrm{OPT}}(\theta_f| \theta_i,S) \tilde{P}_{\textrm{par}}(\theta_i)d\theta_i d\theta_f
\end{align}
where $P_{\textrm{OPT}}(\theta_t| \theta_i,S)$ denotes the probability that OPT, initialised with parameters $\theta_i$ on a DNN, converges to parameters $\theta_f$ when training is halted after the first epoch where zero classification error is achieved on $S$\footnote{In the special  case where we specify the experiment as `overtraining', then we take the parameters after $p$ epochs with $0$ classification error.}, if such a condition is achieved in a number of iterations less than the maximum number which we allow for the experiments. The initialisation distribution $\tilde{P}_{\textrm{par}}(\theta_i)$ is defined analogously to $P_{\textrm{par}}(\theta)$ in \cref{eqn:Pf_vol} (though it need not be exactly the same). $\popt$ is, therefore, a measure of the `size' of $f$'s `basin of attraction', which intuitively refers to the set of initial parameters that converge to $f$ upon training. 

\section{Methodology, Datasets and DNNs}\label{sec:methodology}

\subsection{Methodology}\label{subsec:methods}

\subsubsection{Definition of functions}\label{subsubsec:def_functions}

For a specific DNN, training set $S=\{(x_i,y_i)\}_{i=1}^{|S|}$ and test set $E=\{(x'_i,y'_i)\}_{i=1}^{|E|}$, we define a function $f$ as a labelling of the inputs in $S$ concatenated with the inputs in $E$.\footnote{Formally then, our space of functions is $\mathcal{F}=\mathcal{Y}^{\mathcal{X}}$, where $\mathcal{X}=\{x_i\}_{i=1}^{|S|}\cup\{x'_i\}^{|E|}$} We will only look at functions which have $0$ error on $S$, so that, for a particular experiment with fixed $S$ and $E$, the functions are distinguished only by the predictions they make on $E$. Furthermore we only consider binary classification tasks (c.f. \cref{sec:datasets}), so that our output space\footnote{Note that for training with MSE loss, we centered the output so the loss is measured with respect to target values in $\{-1,1\}$. This is so thresholding can occur at a value of the last layer preactivation of $0$, which is the same as for cross-entropy loss on logits.} is $\mathcal{Y}=\{0,1\}$. Therefore, we will represent functions by a binary string of length $|E|$ representing the labels on $E$; the $i$th character representing the label on the $i$th input of $E$, $x'_i$. For the sake of simplicity, we will not make a distinction between this \emph{representation} of $f$ and the function $f$ itself, as they are related one-to-one for any particular experiment (with fixed $S$ and $E$).


Restricting the input space where functions are defined can be thought of as a coarse-graining of the functions on the full input space (e.g.\ the space of all possible images for image classification), which allows us to estimate their probabilities from sampling. In the following subsections we explain how the main experimental quantities are computed. Further detail can be found in \cref{app:exampleresults}.

\subsubsection{Calculating $\popt$}\label{popt_computation}
For a given optimiser OPT (SGD or one of its variants), a DNN architecture, loss function (cross-entropy (CE) or mean-square error (MSE)), a training set $S$, and test set $E$, we repeat the following procedure $n$ times: We sample initial parameters $\theta_i$, from an i.i.d.\ truncated Gaussian distribution $\tilde{P}_{\textrm{par}}(\theta_i)$, and train with the optimiser until the first epoch where the network has $100\%$ training classification accuracy (except for experiments labelled ``overtraining,'' where we halt training after $p$ further epochs with $0$ training error have occured, for some specified $p$) \footnote{If SGD fails to achieve 100\% accuracy on $S$ in a maximum number of iterations, we discard the run.}. We then compute the function $f$ found by evaluating the network on the inputs in $E$, as described before.

Note that during training, the network outputs are taken to be the pre-activations of the output layer, which are fed to either the MSE loss, or as logits for the CE loss. At evaluation (to compute $f$), the pre-activations are passed through a threshold function so that positive pre-activations output $1$ and non-positive pre-activations output $0$.

We chose sample sizes between $n=10^4$ and $n=10^6$. In other words, we typically sample over $n=10^4$ to $n=10^6$ different trained DNNS, and count how many times each function $f$ appears in the sample  to generate the estimates of $P_{\textrm{OPT}}(f|S)$. We leave the dependence of $P_{\textrm{OPT}}(f|S)$ on $E$ implicit.
This method of estimating $\popt$ is described more formally in~\cref{subapp:exp1}.

\subsubsection{Calculating $\pb$ with Gaussian Processes}\label{pb_computation}
We use neural network Gaussian processes (NNGPs)~\citep{lee2017deep,matthews2018gaussian,garriga-alonso2018deep,novak2018bayesian} to approximate $\pb$, for some training set $S$ and test set $E$. NNGPs have been shown to accurately approximate the prior over functions $P(f)$ of finite-width Bayesian DNNs~\citep{valle2018deep,g.2018gaussian}.  We use DNNs with relatively wide intermediate layers, relative to the input dimension, to ensure that we are close to the infinite layer-width NNGP limit.  Depending on the loss function, we estimate the posterior $\pb$ as follows:

\begin{itemize}
 \item \textbf{Classifiation as regression with MSE loss}. As has been done in previous work on NNGPs, we consider the classification labels as regression targets with an MSE loss\footnote{As for $\popt$ with MSE loss, we take the regression targets to be $\{-1,1\}$, so thresholding occurs at 0}. We compute the analytical posterior for the NNGP with Gaussian likelihood. This is the posterior over the real-valued outputs at the test points on $E$, which correspond to the pre-activations of the output layer of the DNN. We sample from this posterior, and threshold the real-values like we do for DNNs (positive becomes $1$ and otherwise it becomes $0$) to obtain labels on $E$, and thus a function $f$. We then \textbf{estimate $\pb$ by counting how many times each $f$ is obtained from a set of $n$ independent samples from the posterior}, similar to what we did for $\popt$. For more details on GP computations with MSE loss, see \cref{app:GP_exaplanation:MSE}. 
 We describe this method more formally in \cref{subapp:exp2}. We use this technique when comparing $\pb$ with $\popt$ for MSE loss (e.g.\ \cref{fig:summary_fig_exp1_mnist_mse}).
 
 \item \textbf{Classification with CE loss}. In several experiments, we approximate the NNGP posterior using a 0-1 misclassification loss, which is more justified for classification, and can be thought of as a ``low temperature'' version of the CE loss. Since, in contrast to the MSE case, this posterior is not analytically tractable, we use the expectation propagation (EP) approximation  to estimate probabilities~\citep{rasmussen2004gaussian}. In particular, we \textbf{ estimate $\pb$ via ratio of EP-approximated likelihoods}. The EP approximation can be used  to estimate the marginal likelihood of any labelling over any set of inputs, given a GP prior. As shown in \cref{eqn:bayesian_update}, we can use Bayes theorem to express $\pb$ as a ratio of $P(f)$ and $P(S)$ (which is valid for functions with $0$ error on $S$, and the $0-1$ likelihood), and then use EP approximation to obtain both of these probabilities.  In the text, when we refer to the EP approximation for calculating $\pb$, we are using it as described above.
 
 
 \end{itemize}

\subsubsection{Comparing $\popt$ to $\pb$}\label{pb_comparing}
We note that for MSE loss, we can sample to accurately estimate function probabilities, whereas for the CE loss, we must use the EP approximation to calculate the probability of functions.\footnote{See \cref{app:GP_exp_0-1} for more details.}
When we compare $\pb$ to $\popt$ for CE loss, we take the functions found by the optimiser, which are obtained as described in \cref{popt_computation}, and calculate their $\pb$  using the EP approximation. For MSE loss, both the $\pb$ and the $\popt$ are sampled independently, and probabilities are compared for functions found by both methods.

\subsubsection{Calculating $\pb$ for functions with generalisation error from $0\%$ to $100\%$} 
 \label{pb_vs_error_computation}
 
For the zero training error case studied here, we define functions by their particular labelling on the test set $E$ (as described in \cref{subsubsec:def_functions}).
A function can be generated by picking a certain labelling. Subsequently $\pb$ for CE loss can be calculated using the EP approximation as described above.  To study how $\pb$ varies with generalisation error $\epsilon_G$ on $E$ (the fraction of missclasified inputs on $E$), we perform the following procedure.  For each value of $\epsilon_G$ chosen, typically 10 functions are uniformly sampled by randomly selecting $\epsilon_G |E|$ bits in $E$ and flipping them. EP is then used to calculate $\pb$ for those functions.   The probabilities $\pb$ can range over many orders of magnitude. The low probability functions cannot be obtained by direct sampling, so that a full comparison with $\popt$ is not feasible.  This is more formally described in~\cref{subapp:exp3}.

\subsubsection{CSR complexity}\label{sec:CSR}

The critical sample ratio (CSR) is a measure of complexity of functions expressed by DNNs \citep{krueger2017deep}. It is defined with respect to a sample of inputs as the fraction of those samples which are critical samples. A critical sample is defined to be an input such that there is another input within a box of side $2r$ centred around the input, producing a different output (for discrete classification outputs).
See \cref{app:CSR} for further details.

\subsection{Data sets}\label{sec:datasets}

To efficiently sample functions, we use relatively small test sets (typically $|E| = 100$) and, as is often done in the theoretical literature, binarise our classification datasets.
We define the datasets used below:

\noindent\textbf{MNIST}:
The MNIST database of handwritten numbers \citep{lecun1999object} was binarised with even numbers classified as 0 and odd numbers as 1. Unless otherwise specified, we used $|S|=10000$ and $|E|=100$.

\noindent \textbf{Fashion-MNIST}:
The Fashion-MNIST database \citep{xiao2017fashion} was binarised with T-shirts, coats, pullovers, shirts and bags classified as 0 and trousers, dresses, sandals, trainers and ankle boots classified as 1. Unless otherwise specified, we used $|S|=10000$ and $|E|=100$.

\noindent\textbf{IMDb movie review dataset}: We take the IMDb movie review dataset from Keras. The task is to correctly classify each review as positive or negative given the text of the review. We preprocess the set by removing the most common words and normalising.\footnote{We used the version of the dataset and preprocessing technique given here: \url{https://www.kaggle.com/drscarlat/imdb-sentiment-analysis-keras-and-tensorflow}} This procedure was employed to make sure there are functions with high enough probability to be sampled multiple times with Experiments 1 and 2 above. Used with $|S|=45000$ and $|E|=50$.

\noindent \textbf{Ionosphere Dataset}:
This is a small non-image dataset with 34 features\footnote{\url{https://archive.ics.uci.edu/ml/datasets/Ionosphere}} aimed at identifying structure in the ionosphere~\citep{sigillito1989classification}. Used with $|S|=301$ and $|E|=50$.

\noindent For image datasets, we will typically use normalised data (pixel values in range [0,1]) for MSE loss, and unnormalised data for CE loss (pixel values in range [0,255]).

\subsection{Architectures} \label{sec:architetures}
We used the following standard architectures. 

\noindent\textbf{FCN:} 2 hidden layer, 1024 node vanilla fully connected network (FCN) with ReLU activations.

\noindent\textbf{CNN + (Max Pooling) + [BatchNorm]:} Layer 1: Convolutional Layer with $32$ features size $3\times3$. (Layer 1a: Max Pool $2\times2$). [Layer 1b: Batch Norm]. Layer 2: Flatten. Layer 3: FCN with width 1024. [Layer 3a: Batch Norm]. Layer 4: FCN, 1 output with ReLU activations.

\noindent\textbf{LSTM:} Layer 1: Embedding layer. Layer 2: LSTM, 256 outputs. Layer 3: FCN, 512 outputs. Layer 4: Fully-Connected, 1 output with ReLU activations for the fully connected layers.

\noindent Hyperparameters are, unless otherwise specified, the default values in Keras 2.3.0. See \cref{subapp:exp1} for details on the parameter initialisation.


\section{Empirical results for $\pb$ v.s.\ $\popt$ for different architectures and datasets }

In this first of two main results sections, we focus on testing our hypothesis that $\pb \approx \popt$ for FCN, CNN and LSTM architectures on MNIST, Fashion-MNIST, the IMDb review, and the Ionosphere datasets, using several variants of the SGD optimiser.  In the following subsection will describe the main results in detail for an FCN on MNIST. The experiments in the next sections will be the same except for the architecture, dataset, or optimiser which will be varied.

\subsection{Comparing $\pb$ to $\popt$ for an FCN on MNIST} 

\begin{figure}
\centering

\begin{subfigure}[b]{0.31\textwidth}
 \tikz[remember picture] \node[anchor=south, inner sep=0] (model) {\includegraphics[width=\textwidth]{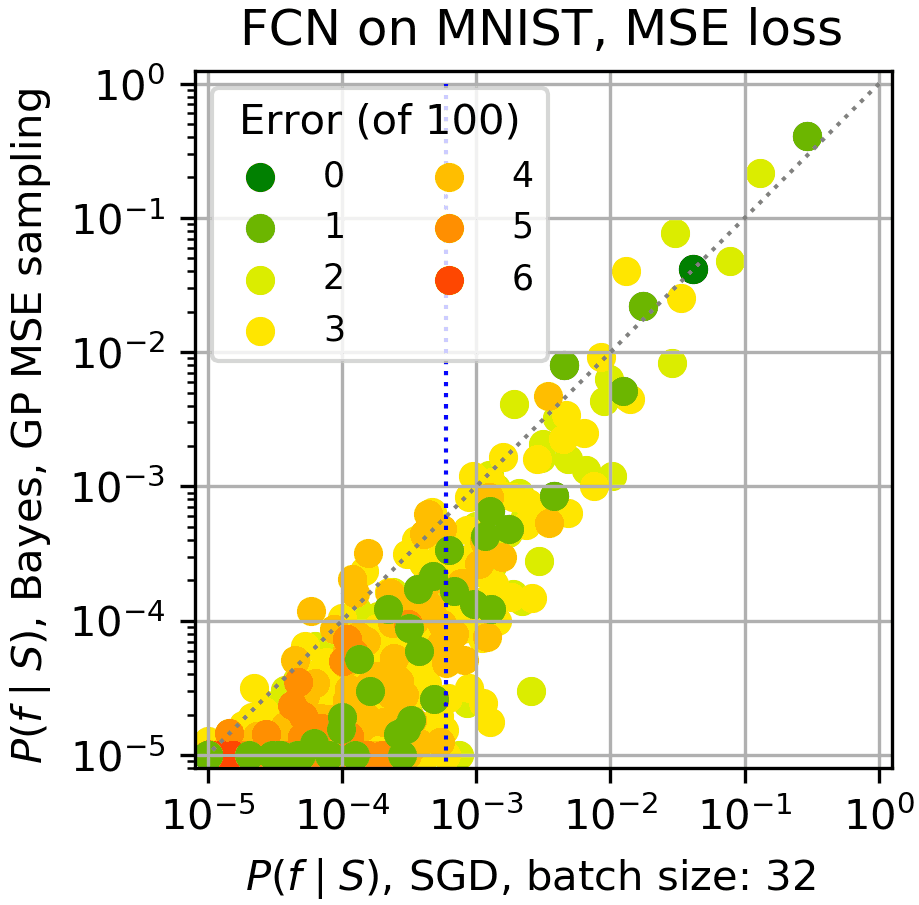}};
 \caption{$\pb \textrm{\,v.s.} \psgd$\\$\quad$}
 \label{fig:summary_fig_exp1_mnist_mse}
\end{subfigure}
~~
\begin{subfigure}[b]{0.31\textwidth}
 \tikz[remember picture] \node[anchor=south, inner sep=0] (replica) {\includegraphics[width=\textwidth]{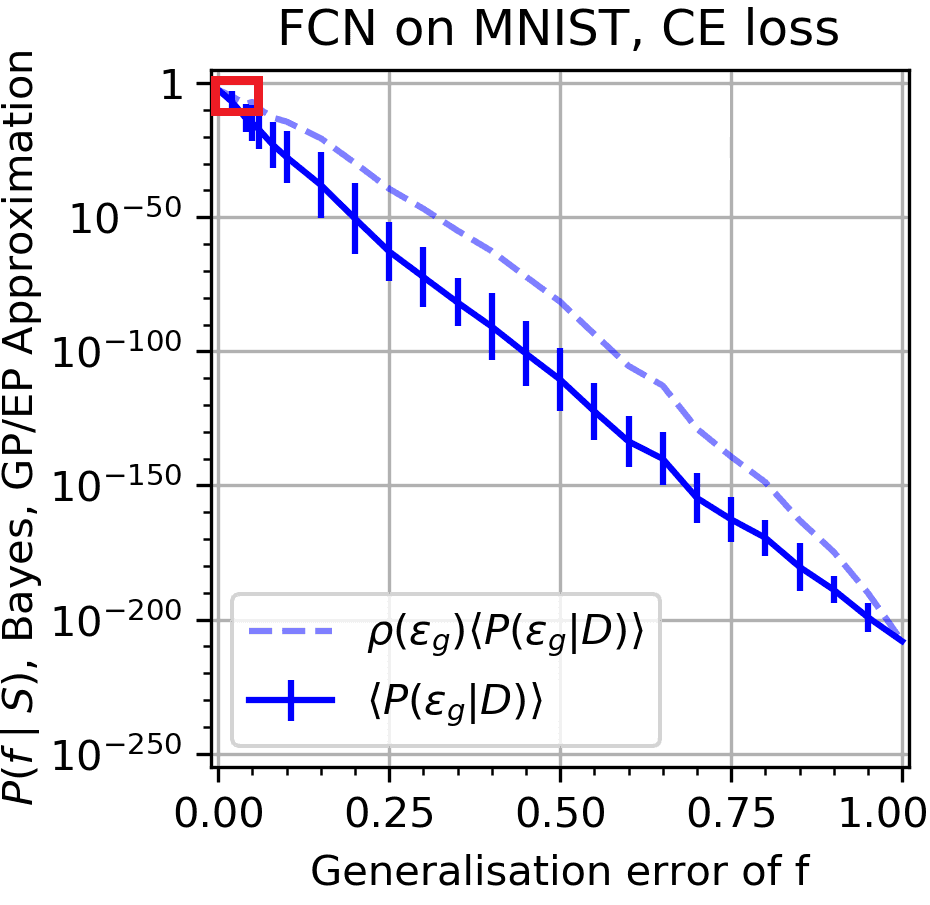} };
 \caption{ $\pb \textrm{\,v.s.\,} \epsilon_G$\\$\quad$}
 \label{fig:summary_fig_exp2_mnist}
\end{subfigure}
~~
\begin{subfigure}[b]{0.31\textwidth}
 \tikz[remember picture] \node[anchor=south, inner sep=0] (replica2) {\includegraphics[width=\textwidth]{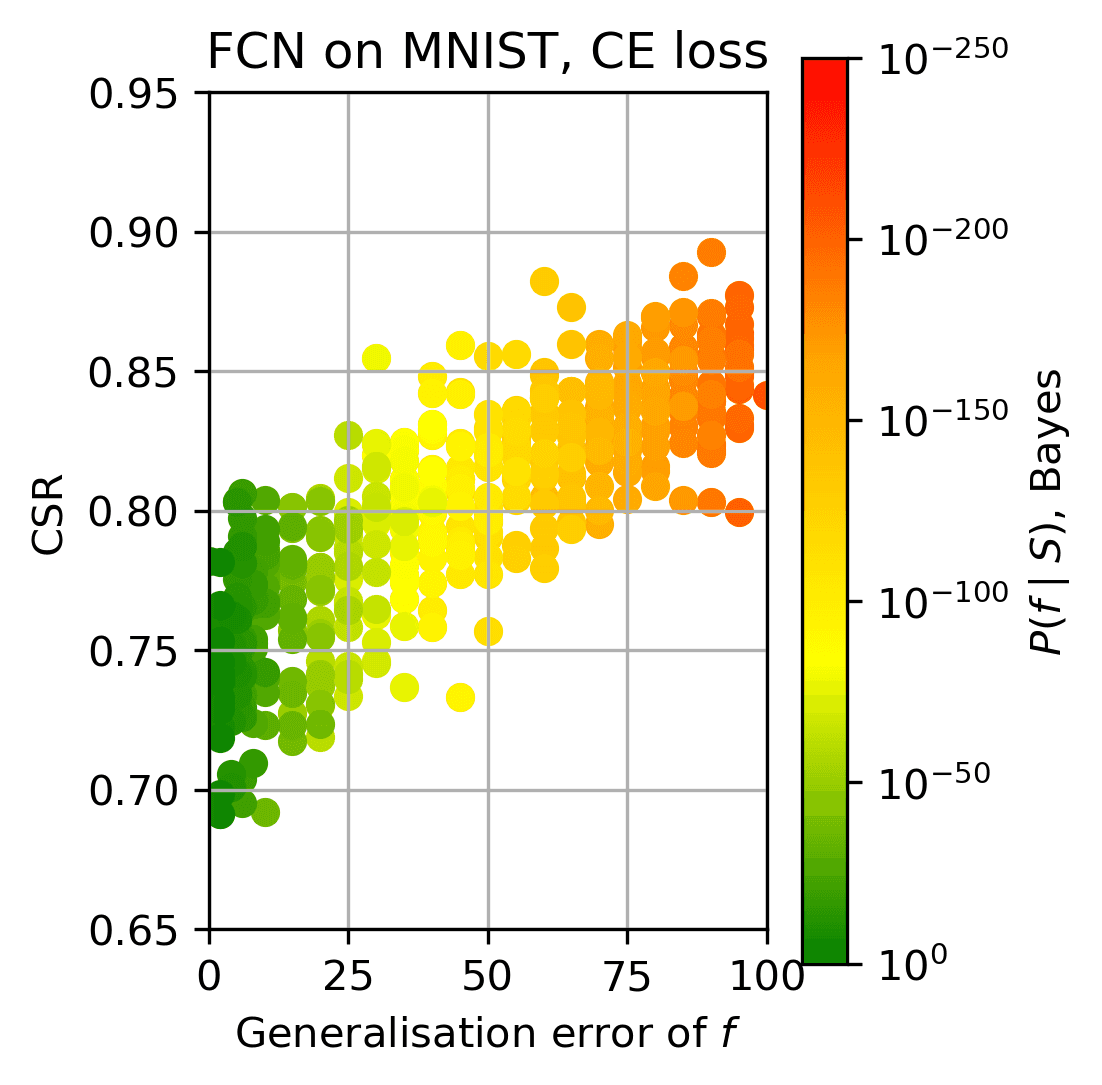}};
 \caption{CSR complexity v.s.\ $\epsilon_G$\\$\quad$}
 \label{fig:summary_fig_exp1_cifar}
\end{subfigure}
\begin{tikzpicture}[overlay,remember picture]
\draw[red, dashed] ([shift={(+4.6cm, 1.97cm)}]model.west)--([shift={(-3.7cm, 1.9cm)}]replica.east);
\draw[red, dashed] ([shift={(+4.6cm, -1.65cm)}]model.west)--([shift={(-3.7cm, 1.7cm)}]replica.east);
\end{tikzpicture}

\begin{subfigure}[b]{0.31\textwidth}
 \includegraphics[width=\textwidth]{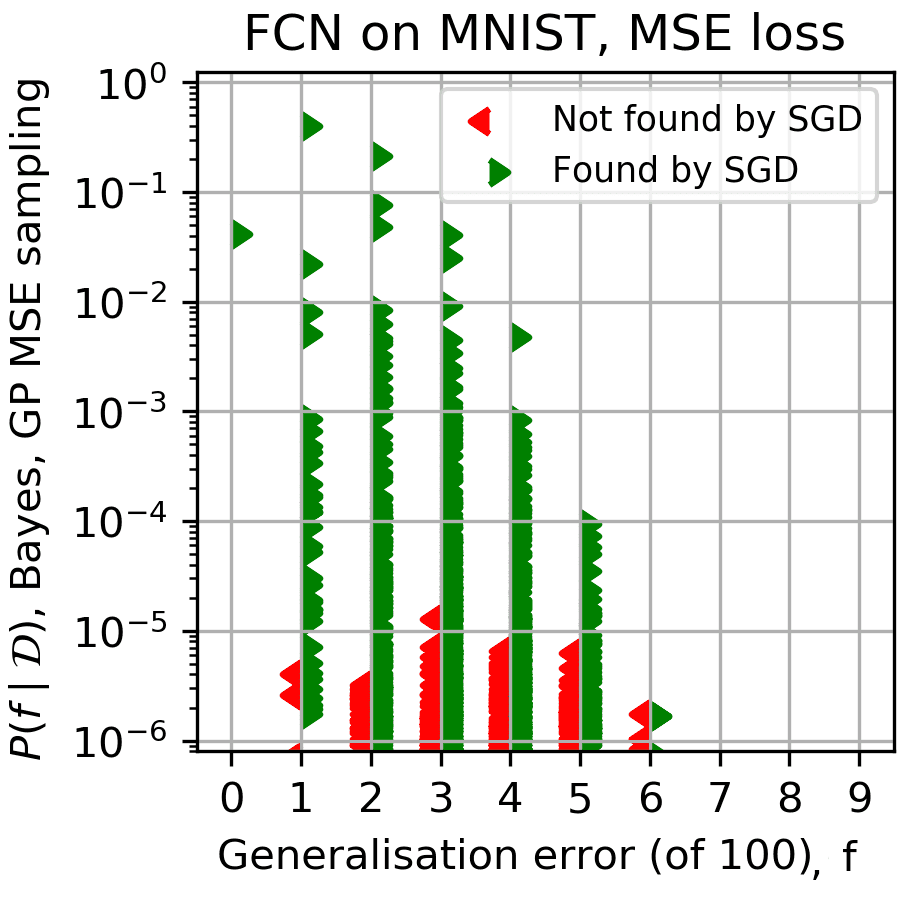} 
 \caption{$f$ found by NNGP in (a). }
 \label{fig:summary_fig_exp3_mnist}
\end{subfigure}
\begin{subfigure}[b]{0.31\textwidth}
 \includegraphics[width=\textwidth]{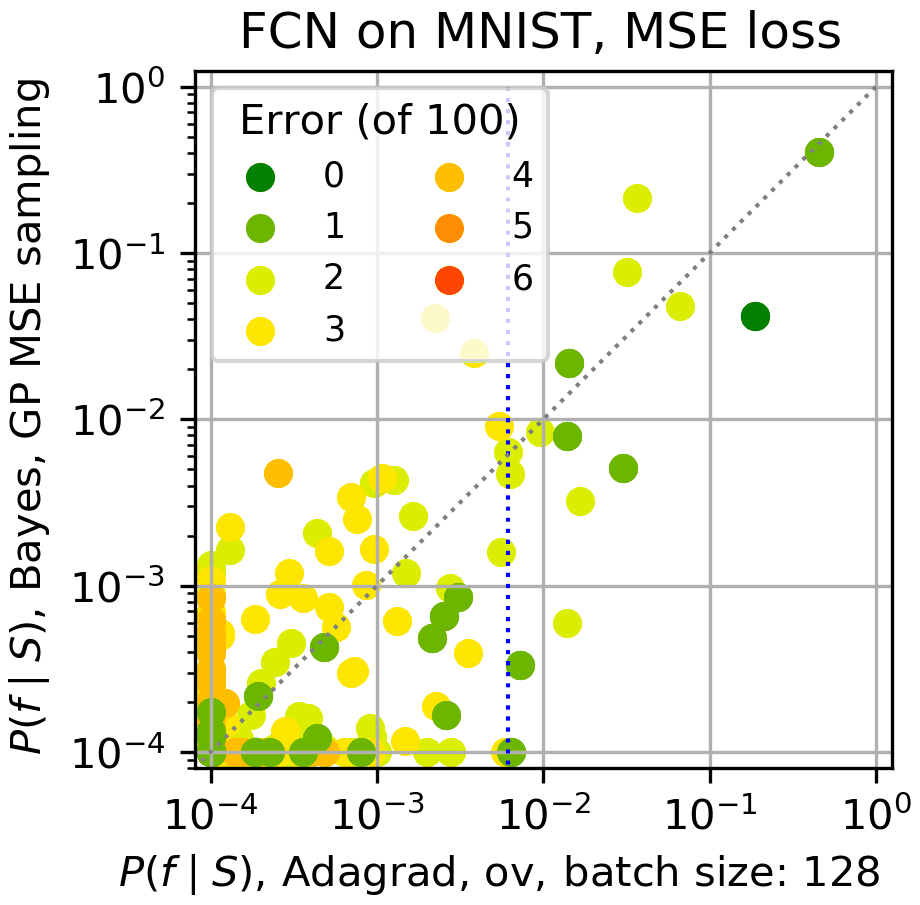}
 \caption{$\pb \textrm{\,v.s.} P_{Adagrad}(f|S)$}
 \label{fig:sum_fig_batch_1}
\end{subfigure}
~~
\begin{subfigure}[b]{0.31\textwidth}
\includegraphics[width=\textwidth]{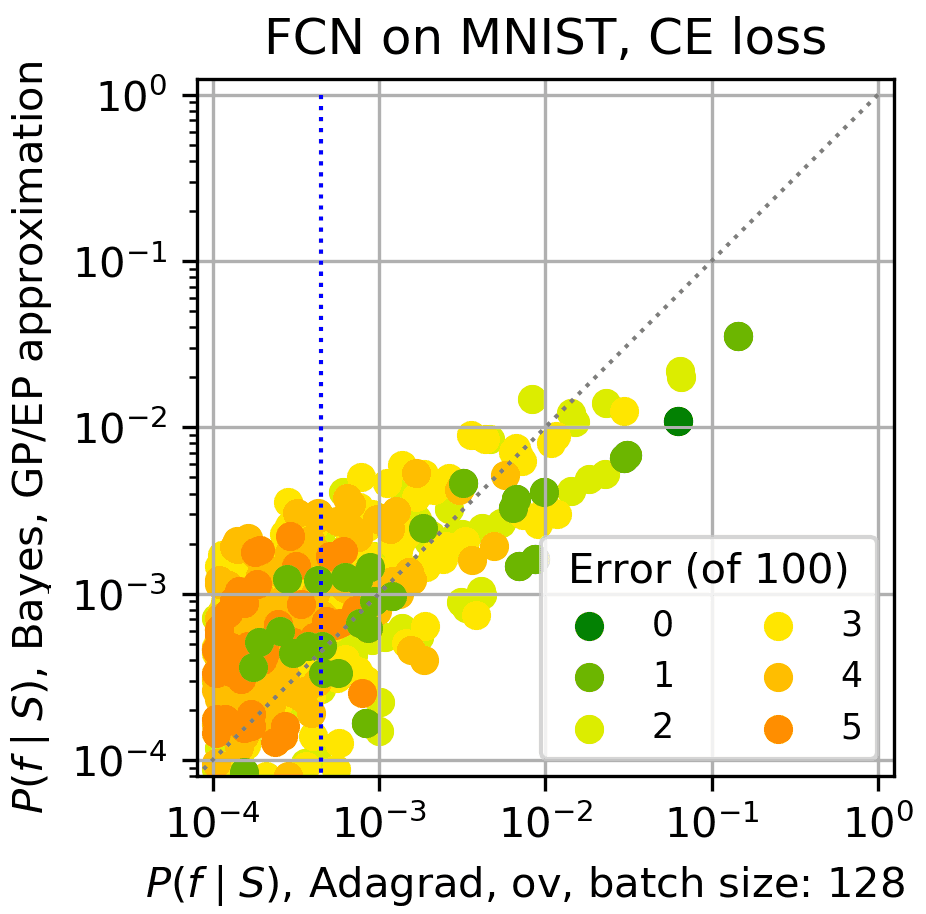}
 \caption{$\pb \textrm{\,v.s.} P_{Adagrad}(f|S)$}
 \label{fig:optvsopt_adagrad_o_main}
\end{subfigure}

\caption{ \textbf{Comparing the Bayesian prediction $\bf P_{B}(f|S)$ to $\bf P_{OPT}(f|\bf S)$ for SGD and Adagrad, for an FCN on MNIST}
[We use training/test set size of 10,000$/$100; For (a,e,f), the vertical dotted blue lines are drawn at the highest value of $\popt$ such that the sum of $\popt$ for all functions above the line is $>90\%$ ($90\%$ probability boundary); dashed grey line denotes $\pb=\popt$.] \\ 
(a) $\pb \textrm{\,v.s. } \psgd$ for MSE loss; Both $\pb$ and $\psgd$ were sampled $n=10^6$ times. The color shows the number of errors in the test set. The GP has average error $\eg_{GP}=1.61\%$, while SGD has average error $\eg=1.88\%$.\\
(b) $\pb$ (with CE loss) v.s.\ $\epsilon_G$ for the full range of possible errors on $E$. We use the methods from \cref{pb_vs_error_computation} with 20 random functions sampled per value of error. The solid blue line shows $\langle\log(\pb)\rangle_{\epsilon_G}$, where the average is over the functions for a fixed $\epsilon_G$; error bars are $2$ standard deviations.
The dashed blue line shows the weighted $\rho(\epsilon_G) \langle \pb \rangle_{\epsilon_G}$, where $\rho(\epsilon_G)$ is the number of functions with error $\epsilon_G$. The small red box and dashed red lines illustrate the range of probability and error found in (a). \\ 
(c) CSR complexity versus generalisation error for the same functions as in fig (b). Color represents $\pb$, computed as in (b). \\
(d) Functions from (a) found by the sample of $\pb$, versus error. 913 functions of the functions are also found by SGD, taking up 97.70\% of the probability for $\psgd$, and 99.96\% for $\pb$.\\
(e) $\pb$ v.s.\ $P_{Adagrad}(f|S)$ for MSE loss; $P_{Adagrad}(f|S)$ was sampled $n=10^5$ times (while the GP sample was the same as in (a)). Adagrad was overtrained until 64 epochs had passed with zero error. The average error is $\eg = 1.53\%$. \\
(f) is as (e) but with CE loss, so that the EP approximation was used for $\pb$, making the estimate of $\pb$ slightly less accurate. $\eg = 2.63\%$.
}
\label{fig:main}
\end{figure}
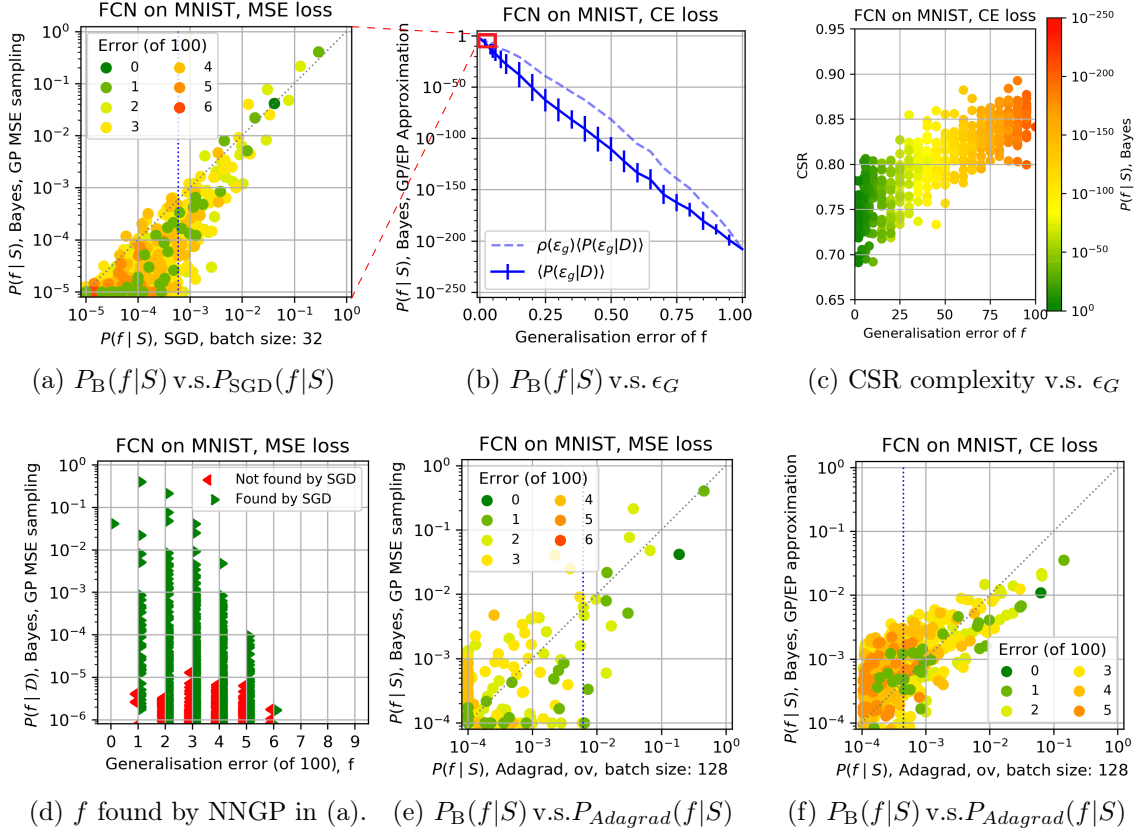

In \cref{fig:main} we present a series of results for a standard DNN setup: an FCN (2 hidden layers, each $1024$ node wide with ReLU activations), trained on (binarised) MNIST to zero training error with a training set size of $|S|=10,000$ and a test set size of $|E|=100$. Note that even for this small test set, there are $2^{100} \approx 1.3 \times 10^{30}$ functions with zero error on $S$, all of which an overparametrized DNN could express~\citep{zhang2016understanding}\footnote{We also find in \cref{fig:CSR_prior} and \cref{fig:CSR_posterior} that our 2-layer FCN is capable of expressing functions on MNIST with the full range of training and generalisation errors }.
We chose standard values for batch size, learning rate, etc., if given by the default values in Keras 2.3.0 (e.g.\ batch size of $32$ and learning rate of $0.01$ for SGD). Our experiments in \cref{sec:batch_opt} and the appendices will show that our results are robust to the choice of these hyperparameters. 

\noindent\textbf{\cref{fig:summary_fig_exp1_mnist_mse}} compares the value of $\pb$ and $\psgd$ for the highest probability functions of each distribution, for MSE loss. Each data point in the plot corresponds to a unique function (a unique classification of images in the test set E). The functions are obtained by sampling both $\pb$ and $\psgd$ and taking the union of the set of functions obtained. $\pb$ and $\psgd$ were estimated as frequencies from the corresponding sample as explained in \cref{popt_computation,pb_computation}. If a function does not appear in one of the samples, we set its frequency to take the minimum value so that it would appear on top of one of the axes.  For example, a function that appears in the SGD runs, but not in the sampling for $\pb$, will appear on x-axis at the value obtained for $\psgd$.  Here we used MSE loss rather than the more popular (and typically more computationally efficient) CE loss because for MSE, the posterior can be sampled from without further approximations, while for CE loss, the expectation propagation (EP) approximation needs to be used making $\pb$ less accurate (see \cref{appendix:GP_explanation} for further details). 

\cref{fig:summary_fig_exp1_mnist_mse} also demonstrates that $\psgd$ and $\pb$ are remarkably closely correlated for MSE loss, and that a remarkably small number of functions account for most of the probability mass for both $\pb$ and $\psgd$. To appreciate how remarkably tight this agreement is, consider the full scale of probabilities for functions $f$ that achieve zero error on the MNIST training set. The average $\pb$ of all these functions is $2^{-100} \approx 10^{-30}$. Therefore the functions in \cref{fig:summary_fig_exp1_mnist_mse} have probabilities that are many orders of magnitude higher than average. At the same time,  $\pb$ and $\psgd$ for these functions typically agree within less than one order of magnitude. Another way of quantifying the agreement is that 90\% of the cumulative probability weight from both $\psgd$ and $\pb$ for the test set $E$ in \cref{fig:summary_fig_exp1_mnist_mse} is made up from the contributions of only a few tens of functions with zero training error out of $\approx 10^{30}$ such possible functions (see vertical doted line in \cref{fig:summary_fig_exp1_mnist_mse}). Moreover, these particular functions are the same for both $\pb$ and $\psgd$.  The agreement between the two methods is remarkable.  Overall, the observations in \cref{fig:summary_fig_exp1_mnist_mse} suggest that the main inductive bias of this DNN is present prior to training.

\noindent\textbf{\cref{fig:summary_fig_exp2_mnist}} plots the mean probability for obtaining a generalisation error of $\epsilon_G$ in the training set $E$, which is estimated as $ \rho(\epsilon_G) \langle P_B(f|S) \rangle_{\epsilon_G}$ where $\rho(\epsilon_G) =|E|!/((|E|-\epsilon_G|E|)!(\epsilon_G|E|)!)$ denotes the number of functions with $\epsilon_G|E|$ errors on $E$, and $\langle P_B(f|S) \rangle_{\epsilon_G}$ denotes the expected value of $\pb$, where the expectation is with respect to uniformly sampling from the set of functions with fixed $\epsilon_G$. As explained in \cref{pb_vs_error_computation}, we estimate the average $\langle \cdot \rangle_{\epsilon_G}$ by sampling, and we estimated $\pb$ for each $f$ in the sample using the EP approximation.

\cref{fig:summary_fig_exp2_mnist} can be interpreted as showing that the inductive bias encoded in $\pb$ gives good generalisation. More precisely, we find that $\pb$ is exponentially biased towards functions with low generalisation error.   To illustrate how strong the bias is, we can look at $\rho(\epsilon_G)$. Over $50\%$ of functions are in the range of $\epsilon_G = 50 \pm 3$ errors, while only $10^{-23}\%$ have $\epsilon_G \leq 3$. Therefore for $\pb$ to overcome the `entropic' factor $\rho(\epsilon_G)$ and show the behaviour in \cref{fig:summary_fig_exp2_mnist}, it must in average give a probability many orders of magnitude higher to low error functions than to high error functions. In \cref{subapp:exp3}, we also observed that the probability $p_i$ of misclassifying an image in the test set varies a lot between images, and that these probabilities are to first order independent. As a corollary, in \cref{fig:experiment_2_figures} we show for $\pb$ and $\psgd$ that the probabilities of multiple images being misclassified can be accurately estimated from the products of the probabilities $p_i$ for misclassifying individual images. Thus this system appears to behave like a Poisson-Binomial distribution with independent and non-identically distributed random $p_i$, which most likely also explains why $ \langle \log P_B(f|S)\rangle_{\epsilon_G}$ scales nearly linearly with $\epsilon_G$.

Although we cannot measure $\psgd$ for the high generalisation error functions, the agreement in \cref{fig:summary_fig_exp1_mnist_mse} (and elsewhere in this paper) implies that $\psgd$ must also be on average orders of magnitude lower for high error functions than low error functions. However, at the moment we can only conjecture that $\psgd$ follows the same exponential behaviour as $\pb$ over the whole range of $\epsilon_G$. Finally, in \cref{subapp:exp3}, we make some further remarks and caveats about this experiment, and other similar experiments.



\noindent\textbf{\cref{fig:summary_fig_exp1_cifar}} shows the correlation between the complexity of the functions obtained to create \cref{fig:summary_fig_exp2_mnist}, and their generalisation error, as well as their $\pb$ (from EP approximation) represented in their color. The complexity measure we used is the critical sample ratio (CSR) complexity~\citep{krueger2017deep} computed on the inputs in $E$, which measures what fraction of inputs are near the decision boundary (see \cref{sec:CSR}).

\cref{fig:summary_fig_exp1_cifar} also shows that there is a inverse correlation between the generalisation of a function and its CSR complexity, as well as between $\pb$ and CSR. In \cref{sec:PB}, we showed that $\pb$ is proportional to the prior probability of a function $P(f)$ for functions that have zero error on the training set $S$. We can thus understand the inverse correlation between $\pb$ and CSR in the light of previous \textit{simplicity bias} results showing that the prior $P(f)$ of Bayesian DNNs is exponentially biased towards functions with low Kolmogorov complexity (simple functions)~\citep{valle2018deep,mingard2019neural}. In~\citep{valle2018deep}, it was further shown for an FCN on a subsample of MNIST that $P(f)$ correlated remarkably well with CSR\footnote{Furthermore in \citep{valle2018deep}, it was shown that this was not an exclusive property of CSR and that any measure that could approximate Kolmogorov complexity seems to also correlate well with $P(f)$.}, and our results are in agreement with that finding. The results in this figure extend those of \cref{fig:summary_fig_exp2_mnist} to show that $\pb$ is biased both towards low error and simple functions, and that simple functions are the ones that tend to have good generalisation on MNIST.  



\noindent\textbf{\cref{fig:summary_fig_exp3_mnist}} shows the correlation between $\pb$ and $\epsilon_G$ for functions used for \cref{fig:summary_fig_exp1_mnist_mse}.   We note that, as can also be observed in  \cref{fig:summary_fig_exp1_mnist_mse}, values of $\pb$ are high for low error functions, and high error functions have relatively lower values of $\pb$.    This figure also uses colour to show which functions were not found in the sampling of $\psgd$. It shows clearly that SGD finds all the high $\pb$ functions.



\noindent\textbf{\cref{fig:sum_fig_batch_1}} shows the same type of experiment as in \cref{fig:summary_fig_exp2_mnist}, but using a different SGD-based optimiser, Adagrad~\citep{duchi2011adaptive} with overtraining (where training was halted after 64 epochs had passed with 100\% training accuracy). We see that it exhibits similar correlation between $\pb$ and $\popt$ to vanilla SGD (and very similar agreement was observed without overtraining). We will see throughout the paper remarkably good correlations between $\pb$ and $\popt$ holds for a large range of optimisers and hyperaparameters

\noindent\textbf{\cref{fig:optvsopt_adagrad_o_main}} shows the same type of experiment as in \cref{fig:summary_fig_exp1_mnist_mse}, but using CE loss, the Adagrad optimiser, and overtraining (also to 64 epochs). See \cref{fig:optvsopt_adagrad} for the equivalent plot but without overtraining. As we are using CE loss (see \cref{pb_computation} and \cref{pb_comparing}), we sample functions from $\popt$, and then use the EP to estimate $\pb$ for the functions obtained. We find similar results to \cref{fig:sum_fig_batch_1}, where we used MSE loss (and direct sampling for $\pb$). The errors introduced by the EP approximation may explain why the correlation does not follow the x=y line as closely as it does for the MSE calculations.
Nevertheless, the correlation between $\pb$ and $\padagrad$ is strong, providing evidence that our results for an FCN on MNIST are not an artefact of the exact optimiser or loss function used.


\begin{figure}[H]
\centering

\begin{subfigure}[b]{0.31\textwidth}
	\includegraphics[width=\textwidth]{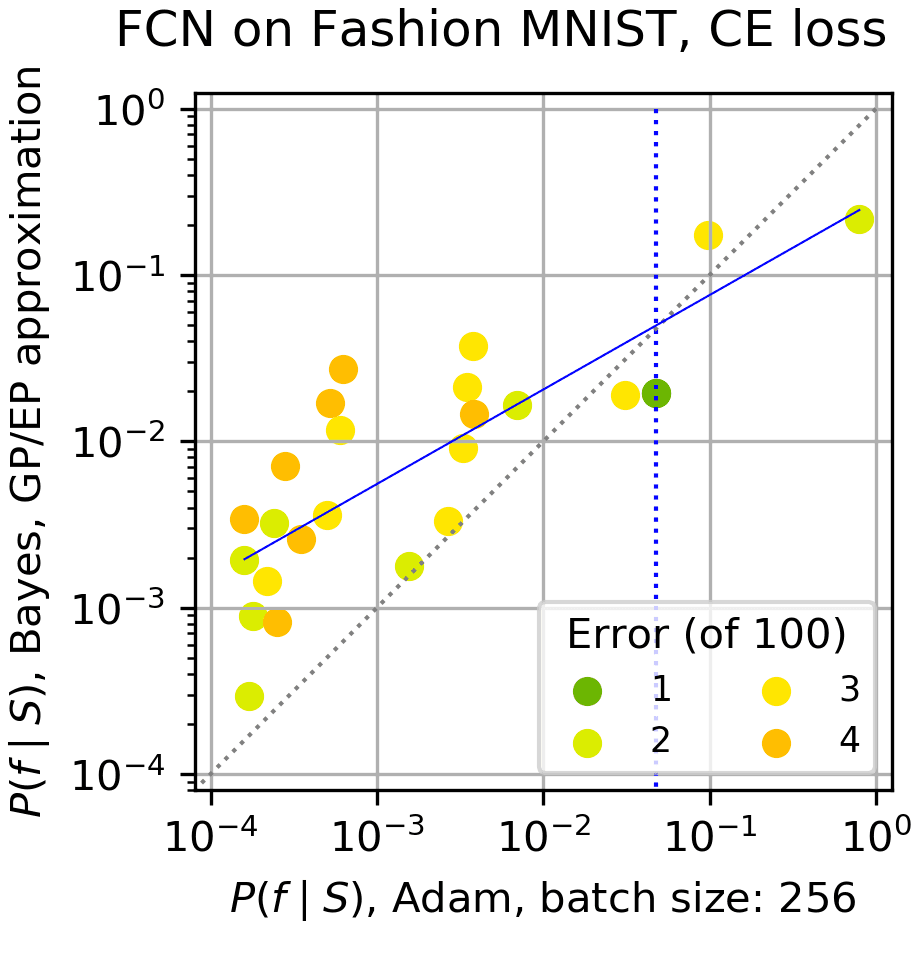}
 \caption{$\pb \textrm{\,v.s.} \padam$ for FCN}
 \label{fig:fashionMNIST_FCN}
\end{subfigure}
\begin{subfigure}[b]{0.31\textwidth}
	\includegraphics[width=\textwidth]{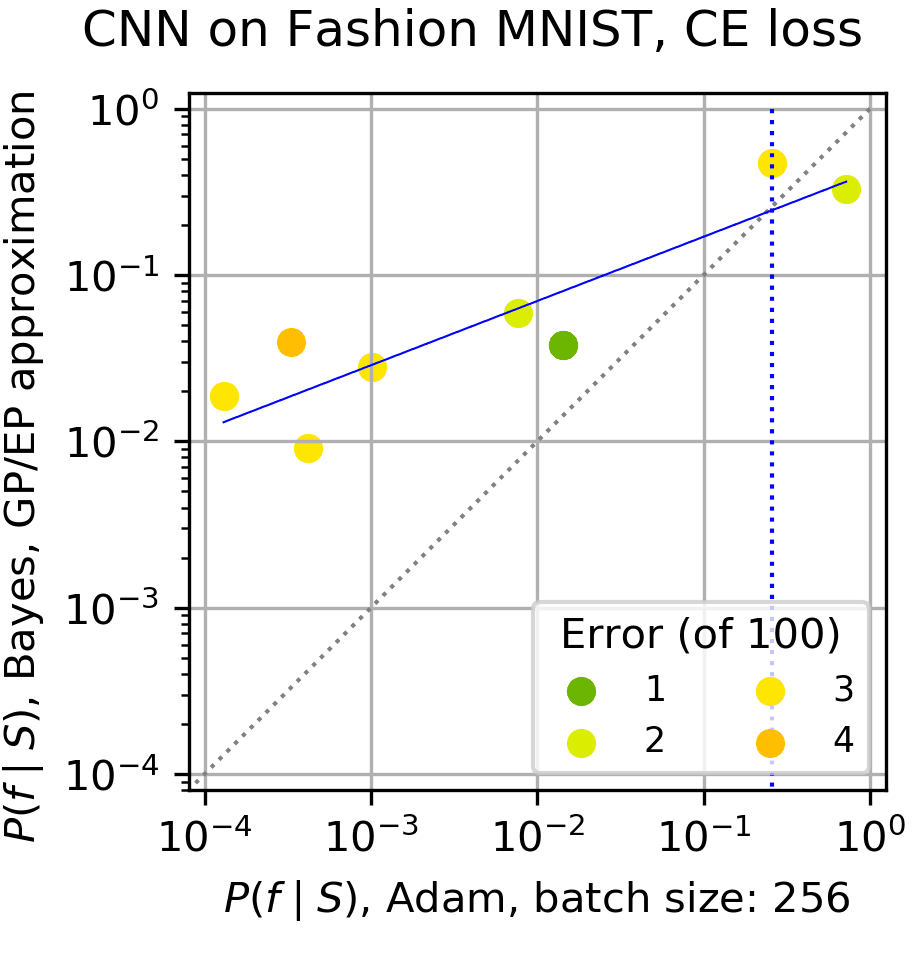}
 \caption{$\pb \textrm{\,v.s.} \padam$ for CNN w/o pooling}
 \label{fig:fashionMNIST_C}
\end{subfigure}
\begin{subfigure}[b]{0.31\textwidth}
 \includegraphics[width=\textwidth]{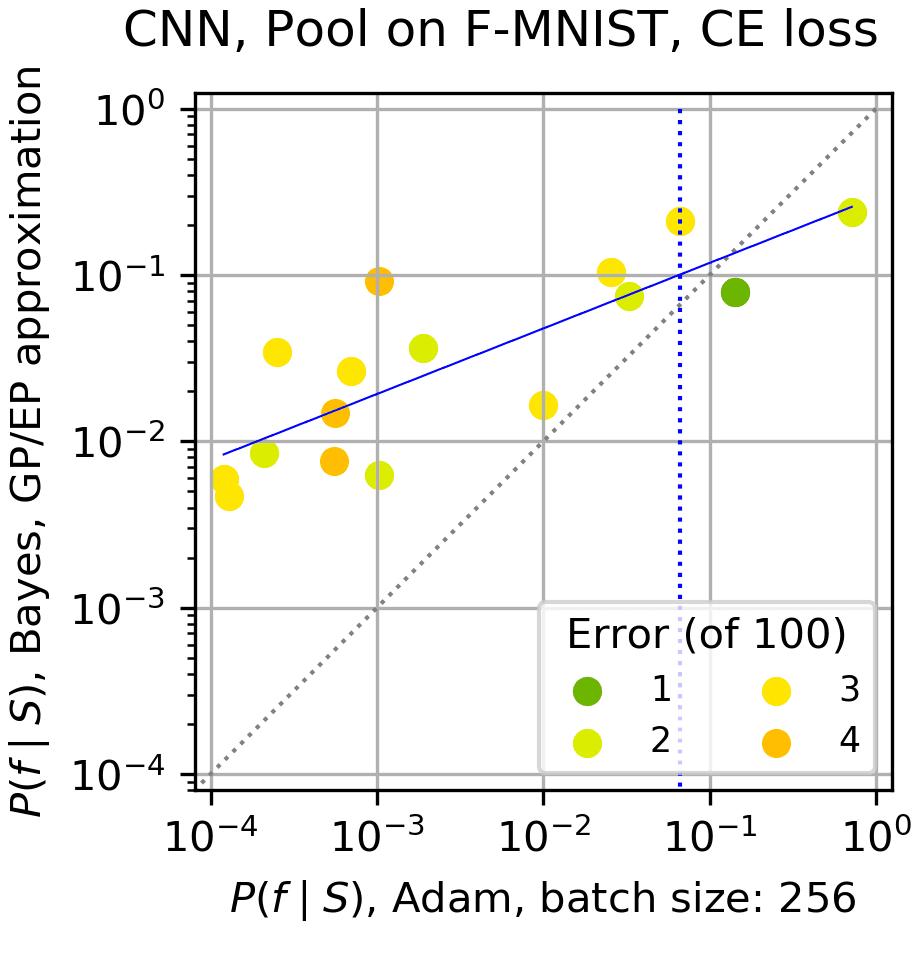}
 \caption{$\pb \textrm{\,v.s.} \padam$ for CNN w/ pooling}
 \label{fig:fashionMNIST_CP}
\end{subfigure}

\caption{{\bf Comparing $\bf P_{B}(f|S)$ to $\bf P_{Adam}(f|\bf S)$ for CNNs and the FCN on Fashion-MNIST} [We use a training/test set size of 10,000$/$100; vertical dotted blue lines denote $90 \%$ probability boundary; dashed grey line is $\pb=\popt$.]
\textbf{(a)} FCN on Fashion-MNIST; $\eg=2.11\%$ for Adam with CE loss. 
\textbf{(b)} Vanilla CNN on Fashion-MNIST; $\eg=2.25\%$ for Adam with CE loss. 
\textbf{(c)} CNN with max-pooling on Fashion-MNIST; $\eg=1.96\%$ for Adam with CE loss. 
Note that when max-pooling is added, the probability of the lowest-error function increases notably for both $\padam$ and $\pb$. There is a strong correlation between $\pb$ and $\psgd$ in all three plots. See \cref{fig:fashionMNIST_appendix} for related results, including $\pb$ vs $\epsilon_G$, a CNN with batch normalisation, and a CNN with MSE loss.}\label{fig:fashionMNIST}
\end{figure}

\subsection{Comparing $\pb$ to $\padam$ for CNNs on Fashion-MNIST}

We next turn to a more complex dataset, namely Fashion-MNIST~\citep{xiao2017fashion} which consists of images of clothing, as well as a more complex network architecture, the CNN \citep{lecun1999object} which was designed in part to have a better inductive bias for images. See \cref{sec:datasets} and \cref{sec:architetures} for details on dataset and architecture. We can see in \cref{fig:fashionMNIST} a strong correlation between $\pb$ and the probabilities found by the Adam optimiser~\citep{kingma2014adam}, a variant of SGD.
Note that instead of MSE loss we used CE loss because it is more efficient. A downside of this choice is that we need to use the EP approximation for the GP calculations (see \cref{app:GP_exp_0-1}). Although the correlation is strong, it does not follow x=y as closely as we generally find for MSE loss, which is quite possibly an effect of the EP approximation. See \cref{fig:fashionMNIST_appendix} for an example with MSE loss where the correlation does follow x=y more closely. 
Both the FCN and the CNNs exhibit a strong bias towards low error functions on Fashion-MNIST as we can see in \cref{fig:fashionMNIST_FCN_3} and \cref{fig:fashionMNIST_CPBN}.

For an example of how the effects of architecture modifications can be observed in the function probabilities, compare results in \cref{fig:fashionMNIST_C} for the vanilla CNN to those in \cref{fig:fashionMNIST_CP} for a CNN with max-pooling~\citep{he2016deep}, a method 
designed to improve the inductive bias of the CNN. As expected, the generalisation performance of the CNN improves, and an important contributor is the increase in the probability of the highest probability 1-error function in both $\pb$ and $\padam$, directly demonstrating an enhancement of the inductive bias. See \cref{fig:fashionMNIST_appendix} for related results.
This example demonstrates how a function based picture as well as analysis of the Bayesian $\pb$ sheds light on the inductive bias of a DNN. 
Such insights could help with architecture search, or more generally with developing new architectures with improved implicit bias toward desired low error functions.

\subsection{Comparing $\pb$ and $\psgd$ to Neural Tangent Kernel results} 

\begin{figure}[H]
\centering

\begin{subfigure}[b]{0.4\textwidth}
 \includegraphics[width=\textwidth]{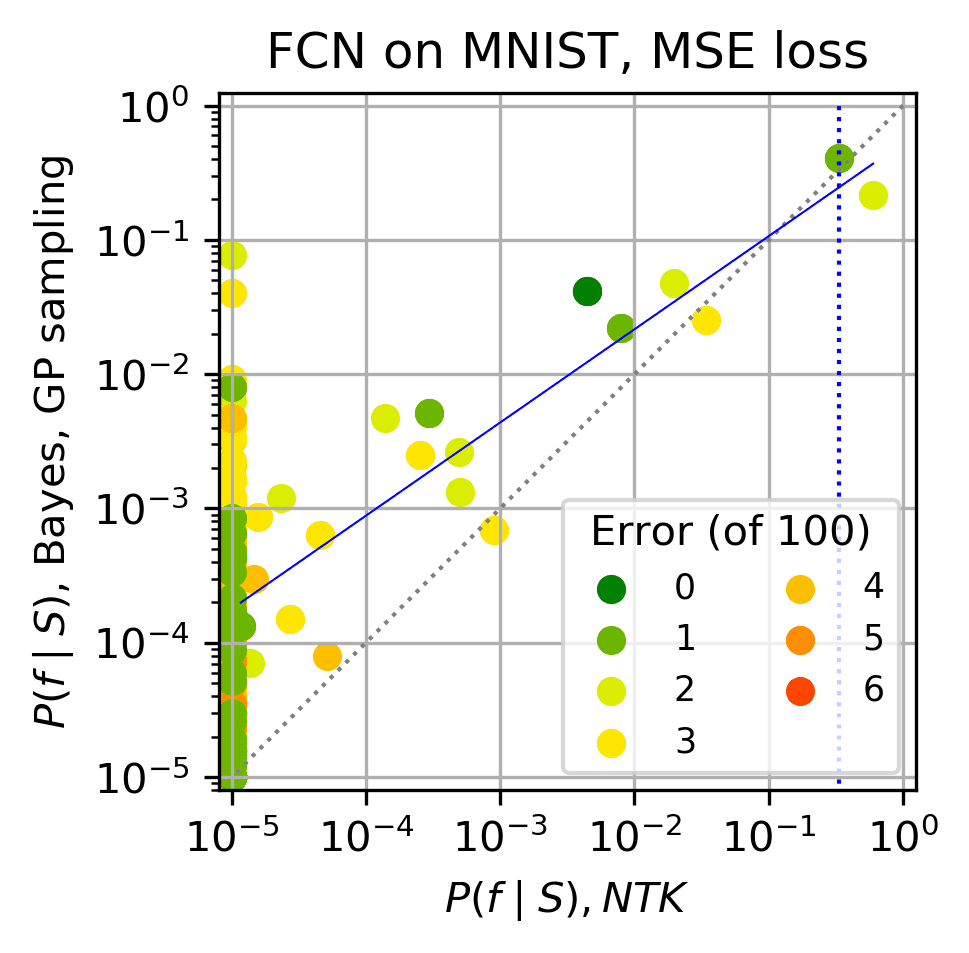}
 \caption{$\pb$ v.s. $\pntk$}
 \label{fig:summary_fig_gpmse_vs_ntk}
\end{subfigure}
~~
\begin{subfigure}[b]{0.4\textwidth}
 \includegraphics[width=\textwidth]{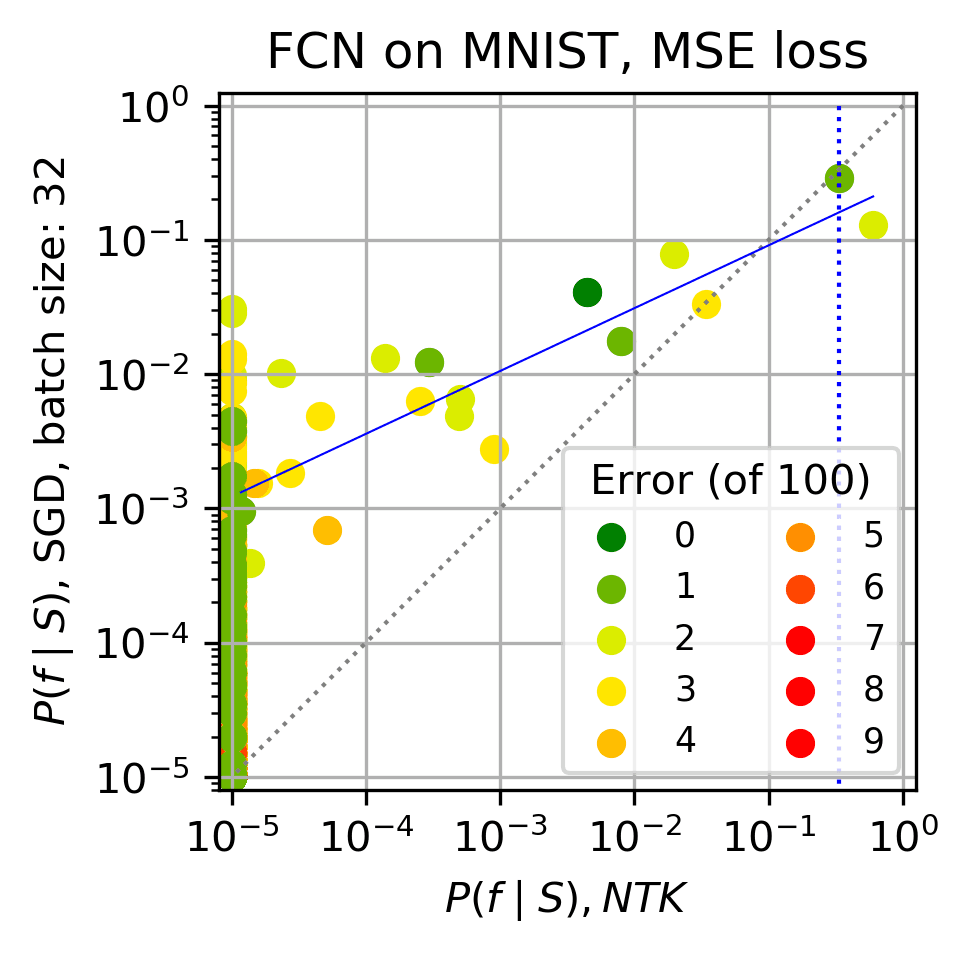}
 \caption{$\psgd$ v.s. $\pntk$}
 \label{fig:summary_fig_gpmse_vs_sgd}
\end{subfigure}

\caption{{\bf Comparing $\bf P_{B}(f|S)$ and $\bf P_{SGD}(f|S)$ to $\bf P_{NTK}(f|\bf S)$ for an FCN on MNIST}. [The functions to the right of the blue dotted lines make up $90\%$ of the total probability. We did $10^7$ samples for NTK and GP, and $10^6$ for SGD]. In (a) we show the correlation between $P_{NTK}(f|\mathcal{D})$ and $\pb$. Weighted by probability, 77.5\% of functions found by sampling from the GP are found by NTK; all functions found by NTK are found by sampling from the GP. In (b), we show the correlation between the $P_{NTK}(f|\mathcal{D})$ and $\psgd$. Weighted by probability, 65.8\% of functions found by SGD are found by NTK; all functions found by NTK are found by SGD.
$\eg=1.69\%$ (NTK), $\eg=1.61\%$ (GP), $\eg=1.88\%$ (SGD).
}
\label{fig:sum_fig_ntk}
\end{figure}

In \cref{fig:sum_fig_ntk} we compare $\pb$ to the output of the neural tangent kernel (NTK)~\citep{jacot2018neural}, which approximates gradient descent in the limit of infinite width and infinitesimal learning rate. The generalisation error of NTK and NNGPs have been shown to be relatively close, and they produce similar functions on simple 1D regression \citep{lee2019wide,novak2019neural}. Here we show that
this similarity also holds for the function probabilities for a more complex classification task. However, we also find the NTK misses many relatively high probability functions that both SGD and the GP find. 
We are currently investigating this surprising behaviour, which may arise from the infinitesimal learning rate. Their low probability may also be exacerbated by the fact that in \cref{fig:sum_fig_ntk} the NTK is very highly biased towards one 2-error function, forcing other functions to have low cumulative probability. Again, this example demonstrates how a function based picture picks up rich details of the behaviour that would be missed when simply comparing generalisation error. 

\subsection{Comparing $\pb$ to $\padam$ for LSTM on IMDb sentiment analysis}
 
\begin{figure}[H]
\centering

\begin{subfigure}[b]{0.31\textwidth}
 \includegraphics[width=\textwidth]{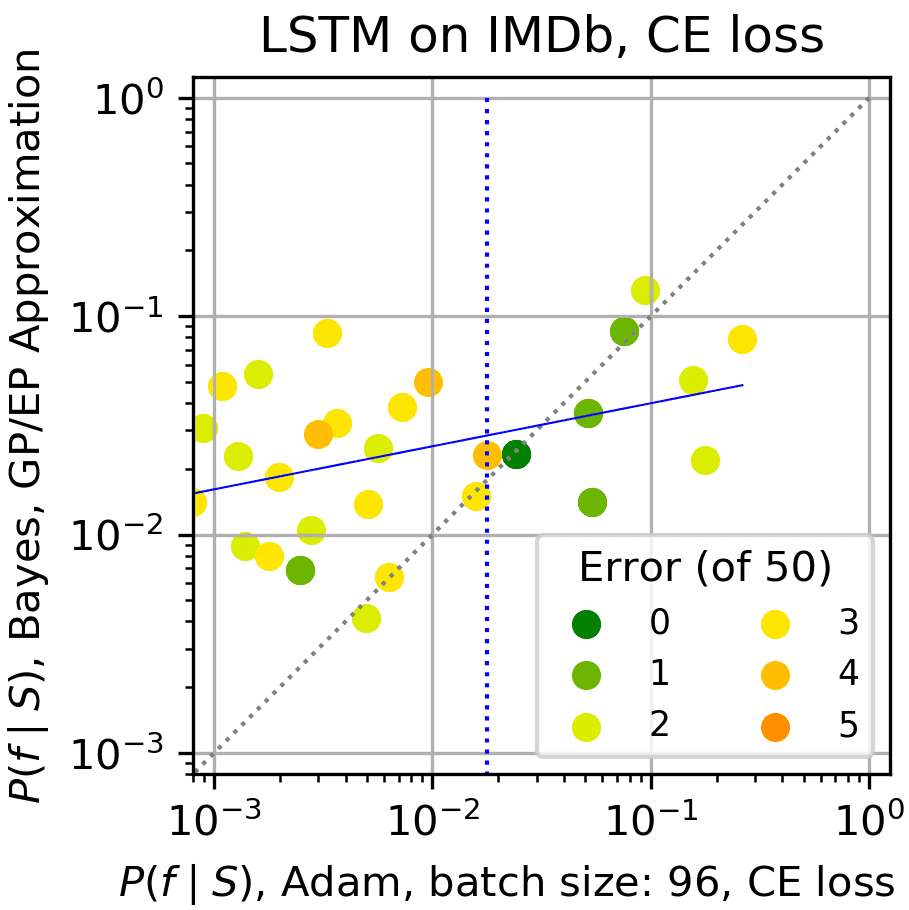}
 \caption{$\pb$ v.s. $\padam$}
 \label{fig:lstm_exp1_main}
\end{subfigure}
~~
\begin{subfigure}[b]{0.31\textwidth}
	\includegraphics[width=\textwidth]{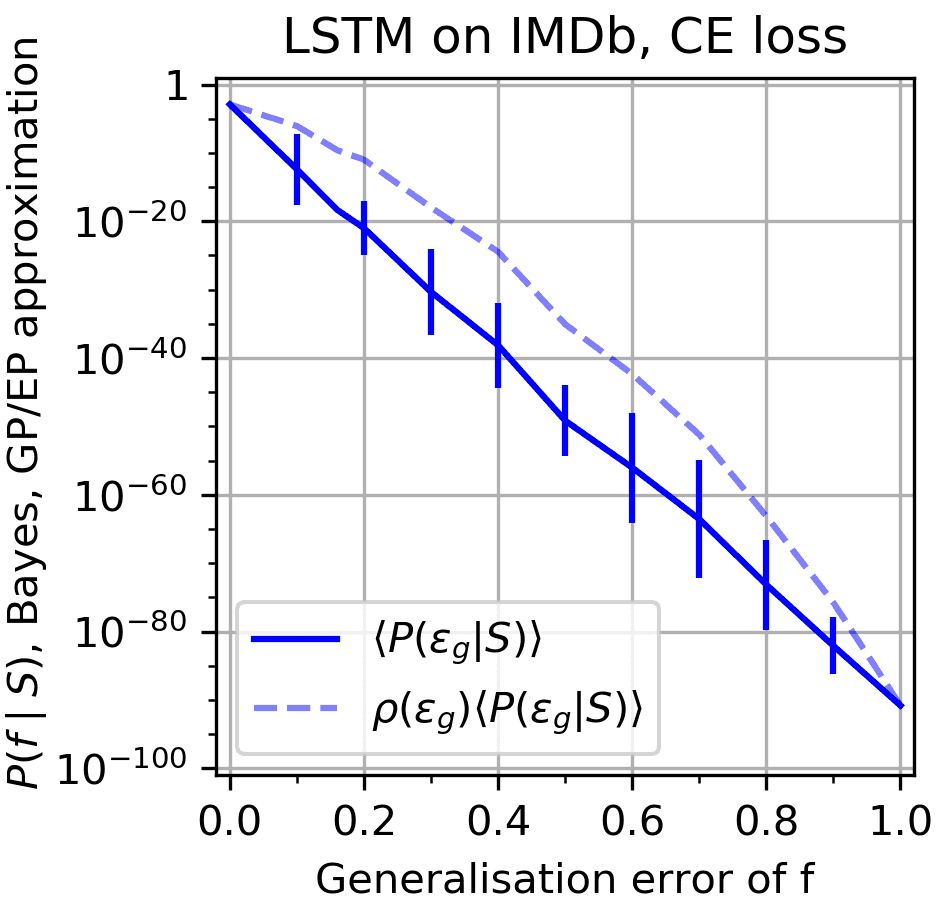}
 \caption{$\pb$ v.s.\ $\eg$}
 \label{fig:lstm_exp2_main}
\end{subfigure}
~~
\begin{subfigure}[b]{0.31\textwidth}
	\includegraphics[width=\textwidth]{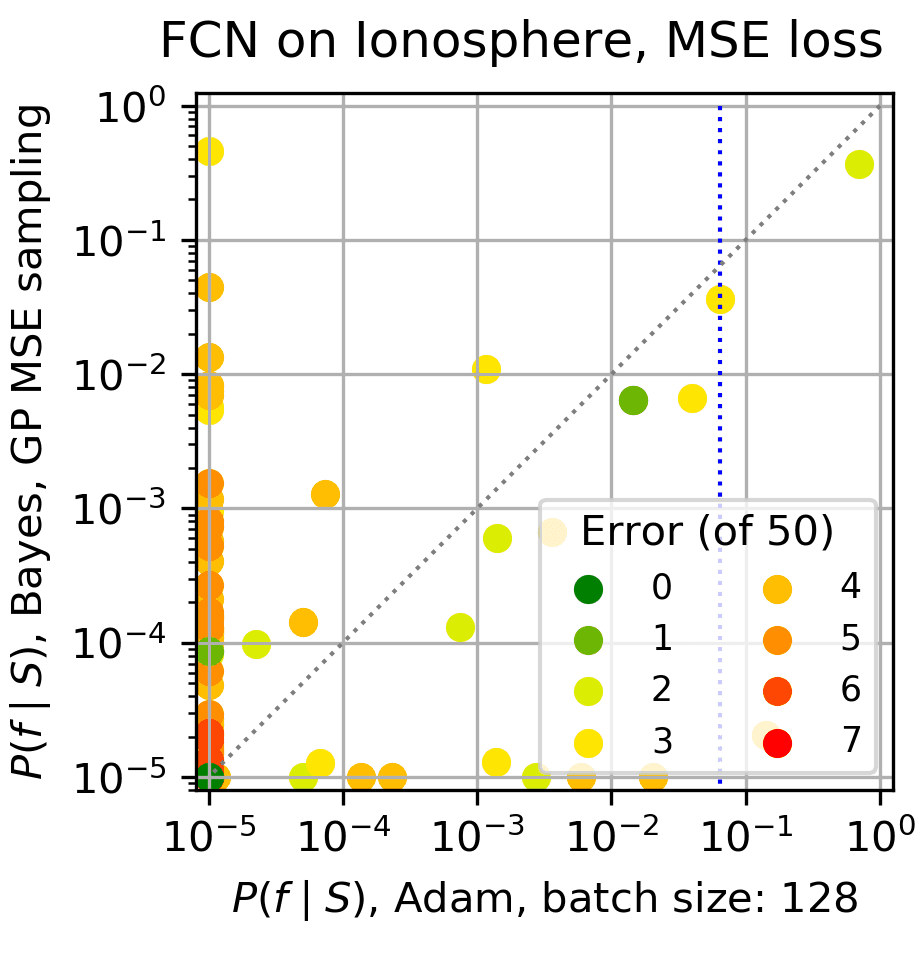}
 \caption{$\pb$ v.s. $\padam$}
 \label{fig:ion_main}
\end{subfigure}

\caption{{\bf Comparing $\bf P_{B}(f|S)$ to $\bf P_{Adam}(f|\bf S)$ for a LSTM on the IMDb movie review dataset, and an FCN on the ionosphere dataset.}
(a) $\pb$ v.s. $\padam$ for LSTM on IMDb dataset, ($\eg=$4.28\%, $10^4$ samples). Because of the computational cost of the problem, we used a training set size of 45000 and a test set of size 50.
(b) $\pb$ v.s.\ $\eg$ for the LSTM on IMDb shows that the functions found by the Adam optimiser are in the small fraction of high $\pb$ probability/low error functions. 
(c) $\pb$ v.s. $\padam$ for an FCN with 3 hidden layers of width 256 on the Ionosphere dataset. Training set size is 301 and the test set size is 50. ($\eg = 4.59 \% $ for Adam, $\eg=5.41\%$ for the GP). See \cref{fig:lstm,fig:nonimage} for further results for these systems.
}\label{fig:lstm_main}
\end{figure}

We test a more complex DNN with a LSTM layer \citep{hochreiter1997long}, applied to a problem of sentiment analysis on the IMDb movie database. We used a smaller test set $|E|=50$ and a larger training set $|S|=$45,000 to ensure that generalisation was good enough to ensure that functions are found with sufficient frequency to be able to extract probabilities.
As can be seen in \cref{fig:lstm_exp1_main} we again observe a reasonable correlation between the functions found by Bayesian sampling, and those found by the optimiser. \cref{fig:lstm_exp2_main} also shows that, as observed for other datasets, this system is highly biased towards low error functions. We show some further experiments with the LSTM in \cref{fig:lstm} in \cref{appendix:further_results}, including an experiment with MSE loss to avoid the EP approximation. 
 
\subsection{Comparing $\pb$ to $\padam$ for FCN on Ionosphere dataset}

As another non-image classification example, we use the small non-image Ionosphere dataset (with a training set of size 301), using an FCN with 3 hidden layers of width 256. As can be seen in \cref{fig:ion_main}, for MSE loss we find a fairly good correlation. Further details and an example with CE loss
can be found in \cref{fig:nonimage}. 

\subsection{Effects of training set size}

We performed experiments comparing $\pb$ and $\popt$ for different training set sizes for the FCN on MINST.  We observe that increasing the amount of training data from $|S|=1000$ to $|S|=20000$ increases the bias towards low error functions.  This increase has the following effects:  1) An increase in the value of $\pb$ and $\psgd$ for functions with low $\eg$ by several orders of magnitude, 2) an increase by several orders of magnitude of $\pb$ and $\popt$ for the mode functions (the ones with highest probability), 3) A decrease in the number of functions that cumulatively take up $90\%$ of the observed probability weight, and 4) a significant increase in the tightness of correlation between $\pb$ and $\popt$. See \cref{app:training_set_size} in \cref{app:training_set_size} for detailed results and plots.

\subsection{Results for other test sets} 

For the experiments shown in this section, sampling efficiency considerations means that we have limited ourselves to a relatively small test sets ($|E|\leq 100$). In \cref{fig:test_set_dependence}, we have checked that other test sets also show close agreement between $\pb$ and $\psgd$. For larger $|E|$, $\popt$ quickly becomes impossible to directly measure empirically -- doubling the test set roughly means squaring the number of samples to obtain qualitatively similar results, as the values for $\pb$ decrease exponentially with test  set size. However, if we assume that the images are approximately independently distributed throughout the larger test set, as \cref{app:notes_on_exp} suggests, then we can estimate the highest probabilities from products of $\pb$ or $\psgd$ on the smaller sets.

\section{The effect of hyperparameter changes and optimisers on $\pb$ and $\popt$ }\label{sec:batch_opt}

In the first section we focussed on the first-order similarity between $\pb$ and $\popt$.  
In this second main results section, we focus on second-order effects that affect $\popt$ differently from $\pb$. These include the effects of hyperparameter settings and optimiser choice.

\subsection{Changing batch size and learning rate }\label{sec:batch_sizes_main_text}

\begin{figure}[H]
\centering

\begin{subfigure}[b]{0.31\textwidth}
 \includegraphics[width=\textwidth]{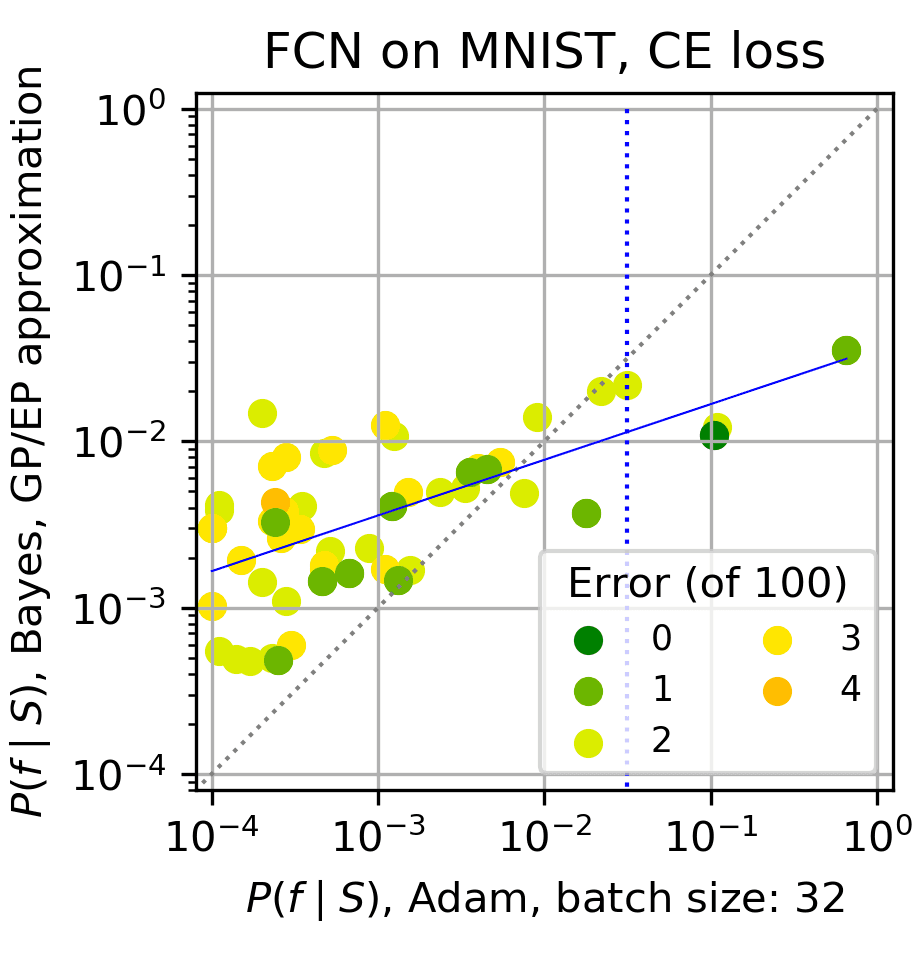}
 \caption{Adam, batch size=$32$} 
 \label{fig:batch:32}
\end{subfigure}
~~
\begin{subfigure}[b]{0.31\textwidth}
 \includegraphics[width=\textwidth]{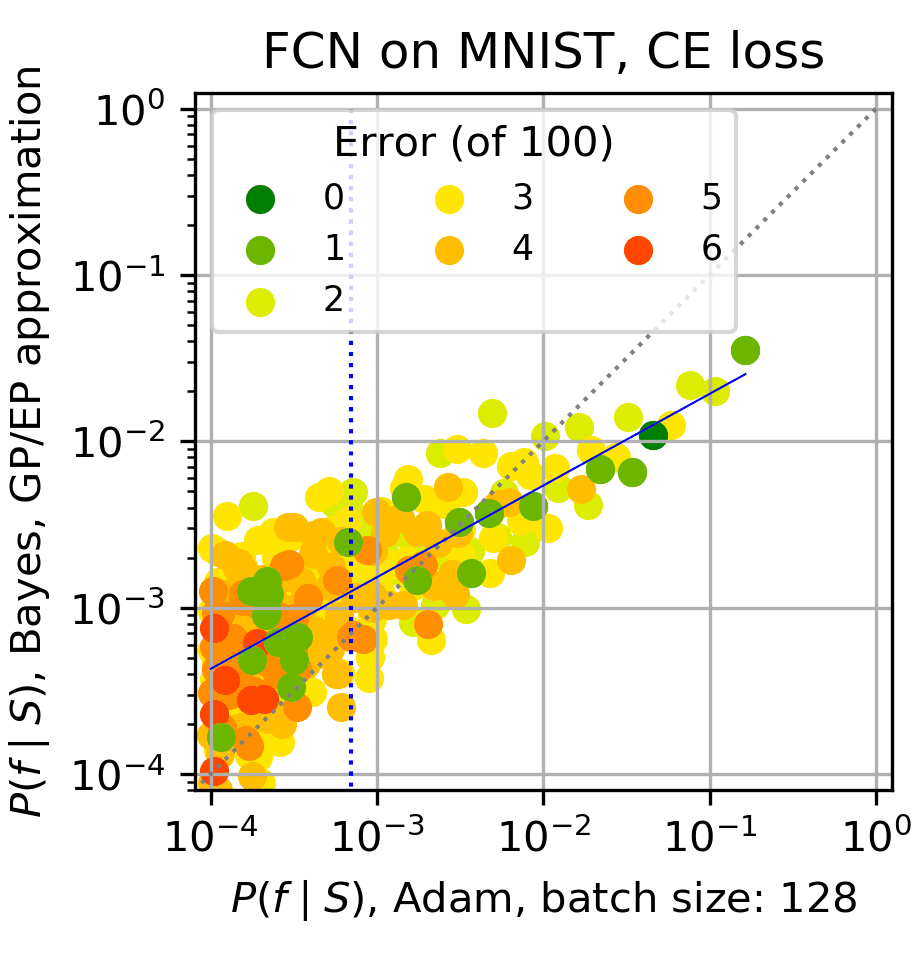}
 \caption{As (a) but batch size=$128$}
 \label{fig:batch:128}
\end{subfigure}
~~
\begin{subfigure}[b]{0.31\textwidth}
 \includegraphics[width=\textwidth]{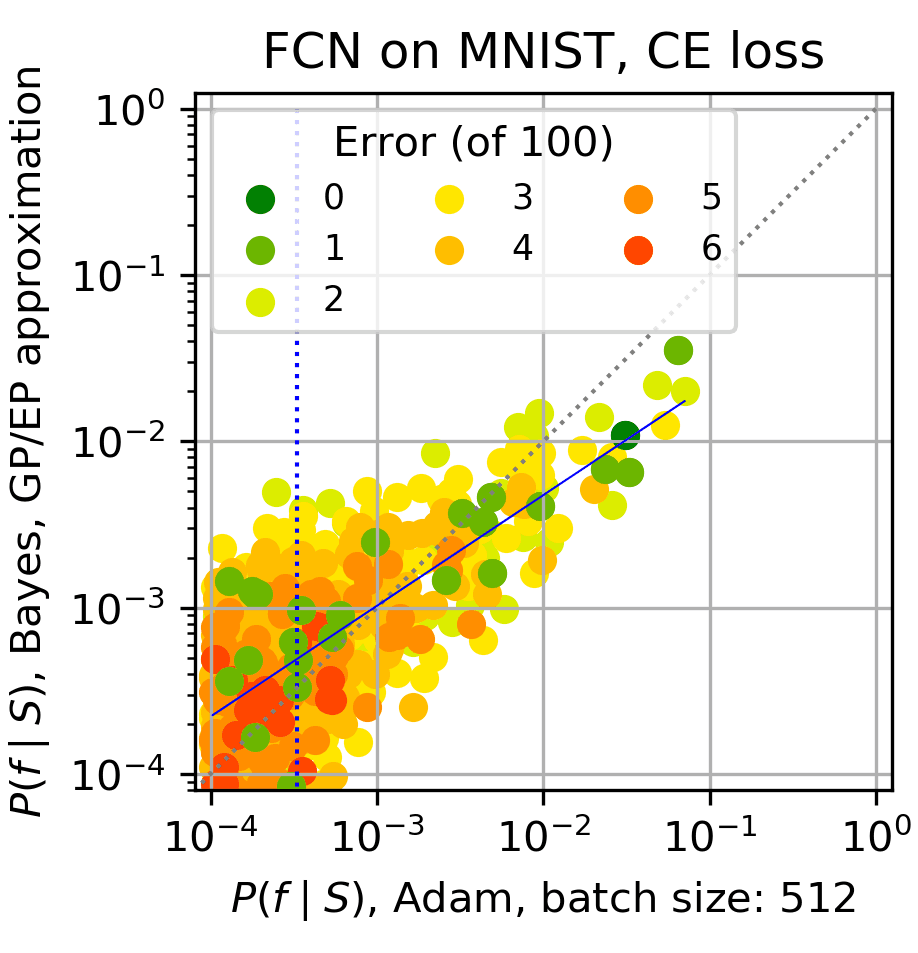}
 \caption{As (a) but batch size=$512$}
 \label{fig:batch:512}
\end{subfigure}

\begin{subfigure}[b]{0.31\textwidth}
 \includegraphics[width=\textwidth]{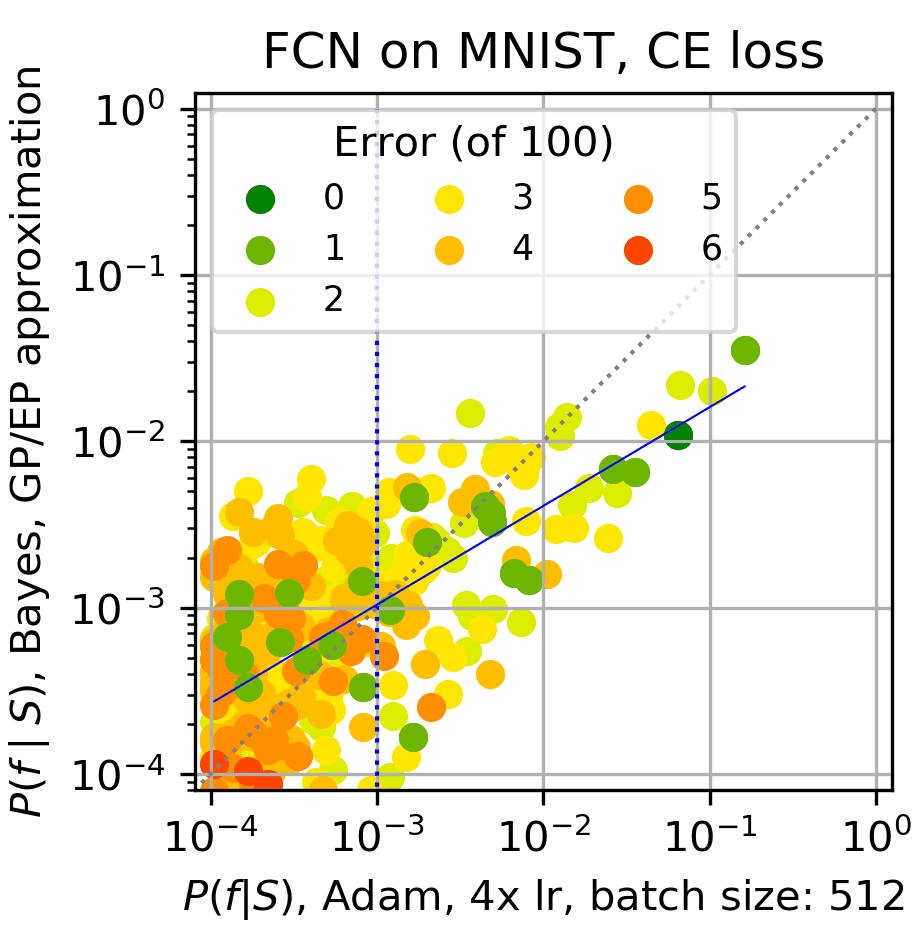}
 \caption{As (c) but 4x learning rate}
 \label{fig:batch:512x4}
\end{subfigure}
~~
\begin{subfigure}[b]{0.31\textwidth}
 \includegraphics[width=\textwidth]{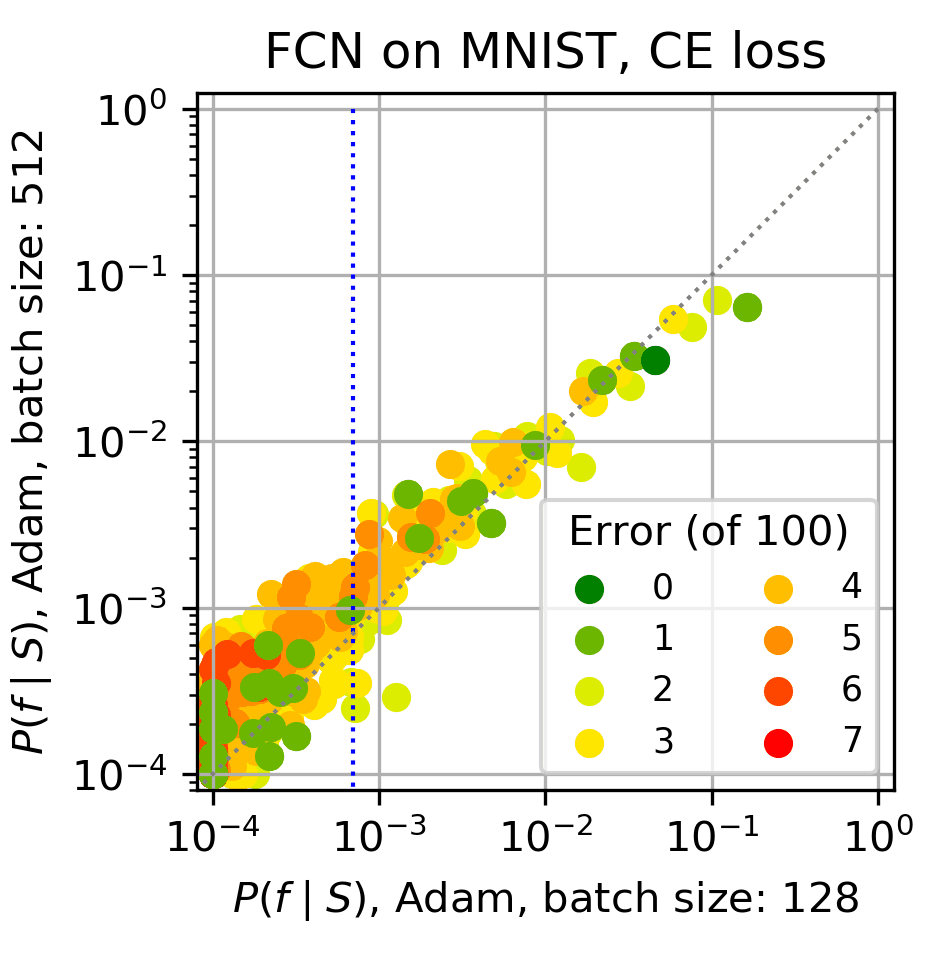}
 \caption{Batch size 128 v.s.\ 512}
 \label{fig:batch:128vs512}
\end{subfigure}
~~
\begin{subfigure}[b]{0.31\textwidth}
 \includegraphics[width=\textwidth]{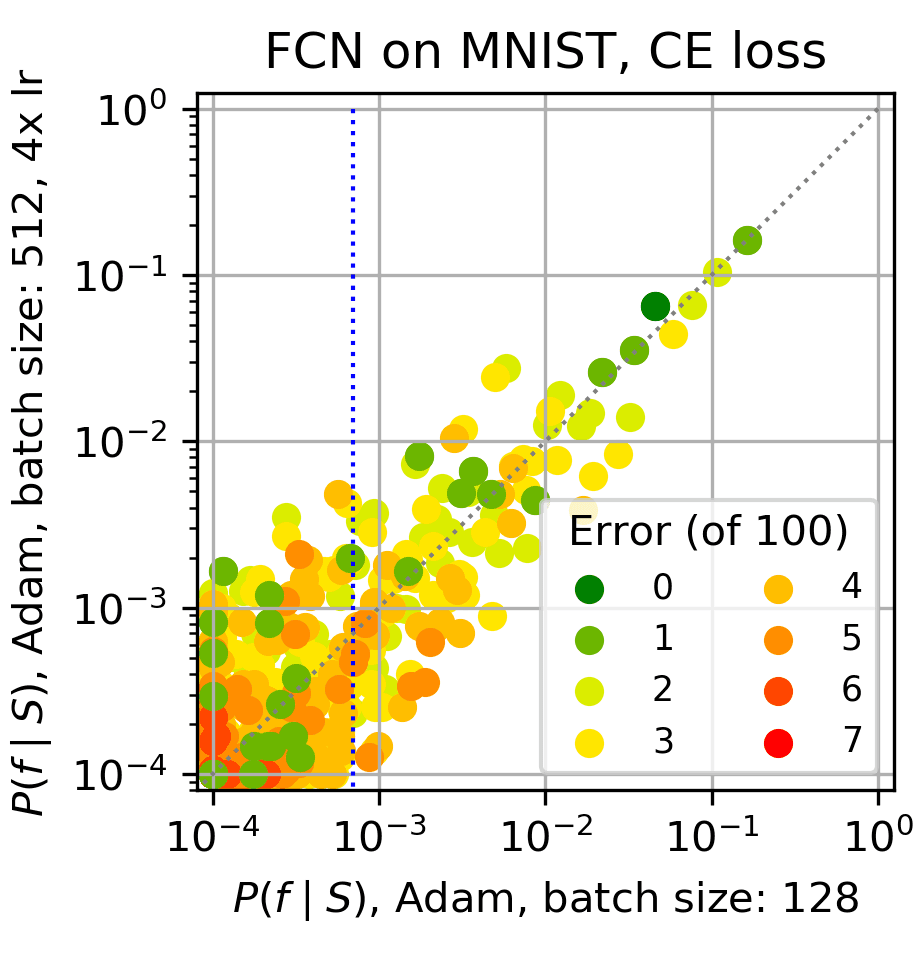}
 \caption{Batch size 128 v.s.\ 512 4x lr}
 \label{fig:batch:128vs512flr}
\end{subfigure}

\caption{\textbf{Effects of changing batch size and learning rate on $\pb$ and $\padam$ for FCN on MNIST with CE loss} [We use training/test set size 10,000$/$100. Vertical dotted blue lines denote $90 \%$ probability boundary; dashed grey line is $x=y$.] \\
 (a) Batch size = 32, \,\, $\eg=1.13\%$.\\
 (b) Batch size= 128, \, $\eg=2.20\%$. \\
 (c) Batch size = 512, $\eg=2.67\%$.\\ 
 (d) Batch size =512 and faster learning rate (4x the others), $\eg=2.14\%$. \\
 (e) Direct comparison of $\padam$ for batch size 128 and 512. \\
 (f) Direct comparison of $\padam$ for batch size 128 and 512 with a $4\times$ faster learning rate. \\
The $\padam$ probabilities for the dominant functions in (d) and (b) are remarkably similar, as can be seen by comparing (e) and (f). It is these higher probability functions that explain the similarity in $\eg$ for batch size 128 and batch size 512 with a faster learning rate. 
See \cref{fig:batchvsbatchmse} for related batch size results for MSE loss.
}
\label{fig:sum_fig_batch}
\end{figure}

In a well-known study, \citep{keskar2016large} showed that, for a fixed learning rate, using smaller batch sizes could lead to better generalisation. In \cref{fig:sum_fig_batch} (a)-(c) we observe this same effect but reflected in the more finely grained spectrum of function probabilities. For batch size 512, we also reproduce in \cref{fig:batch:512x4} the effect observed in \citep{goyal2017accurate,hoffer2017train,smith2017don}, that speeding up the learning rate for a fixed batch size can mimic the improvement in $\eg$ for smaller batches. Interestingly, as can be seen by comparing \cref{fig:batch:512x4,fig:batch:128vs512,fig:batch:128vs512flr}, the overall correlation of the function probability spectrum appears tighter for the 128 and 512 batch size with the same learning rates, even though the generalisation errors are different. However, if the learning rate is increased $4 \times$ for the the 512 batch size system, then there is a closer correlation with batch size 128 for the higher probability functions. It is these latter functions that dominate the average for $\eg$ and so the closer correlation for those functions, rather than the less good correlation for low probability functions, explains the better agreement seen in generalisation error for the two systems. 

Finally, in \cref{fig:batchvsbatchmse} of \cref{appendix:further_results},  we vary batch size for MSE, finding different trends to CE loss. For MSE, increasing batch size leads to better generalisation due to second order effects where $\psgd$ preferentially converges on a few key higher probability/lower error functions. The batch size can be correlated with the noise spectrum of the underlying Langevin equation that describes SGD~\citep{bottou2018optimization,jastrzebski2018finding,zhang2018energy}. What our function based results demonstrate is that the behaviour of the optimiser on the loss-landscape is affected in subtle ways by the form of the loss function, as well as the amount noise, and possibly also by correlations in the noise. 

\subsection{Changing optimisers}

We trained the FCN on MNIST with different optimisers (Adam, Adagrad, RMSprop, Adadelta), and found that to first order $\pb$ correlated well with $\popt$ for all four optimisers. We also observed some second order effects, including that the distribution of $\padam$ and $\padagrad$ were very similar to one another, as were $\prmsprop$ and $\padadelta$, but there was noticeable variation between the two groups. We find that $\padam$ with batch size of 32 is very similar to $\prmsprop$ with a batch size of 128. The effect of optimiser choice, batch size, learning rate, and other hyperameters is complex, and the parameter space is large.  Analysing optimisers in function-space could be a way to better understand the interaction of these choices with the loss landscape, and understanding the effects of hyperparameter tuning.
See \cref{sec:optimisers} for further detail and the plots.

\section{Heuristic arguments for the correlation between $\bf \pb$ and $\bf \psgd$}\label{sec:heuristic_arguments}

\begin{figure}
\centering
\begin{subfigure}[b]{0.31\textwidth}
 \includegraphics[width=\textwidth]{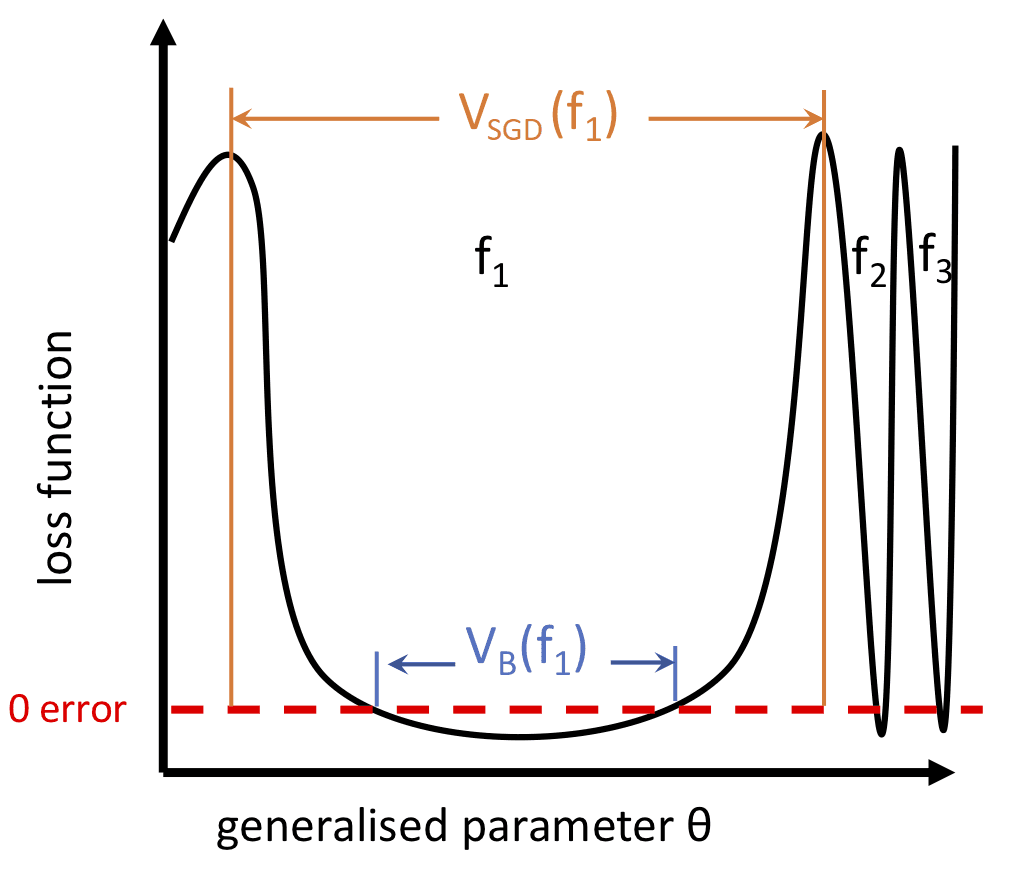}
 \caption{Schematic loss landscape}
 \label{fig:Fig-schematic}
\end{subfigure}
~~
\begin{subfigure}[b]{0.31\textwidth}
 \includegraphics[width=\textwidth]{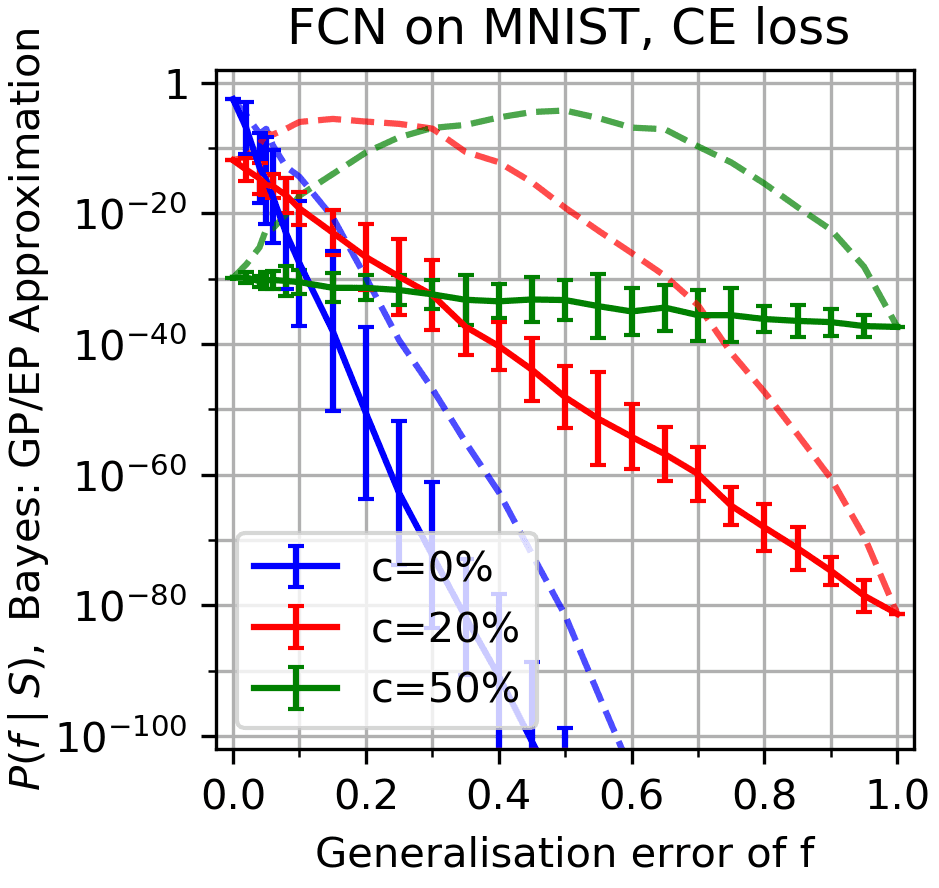} 
 \caption{Corrupted data, CE loss} 
 \label{fig:sf3:fake_multi}
\end{subfigure}
~~
\begin{subfigure}[b]{0.31\textwidth}
 \includegraphics[width=\textwidth]
 {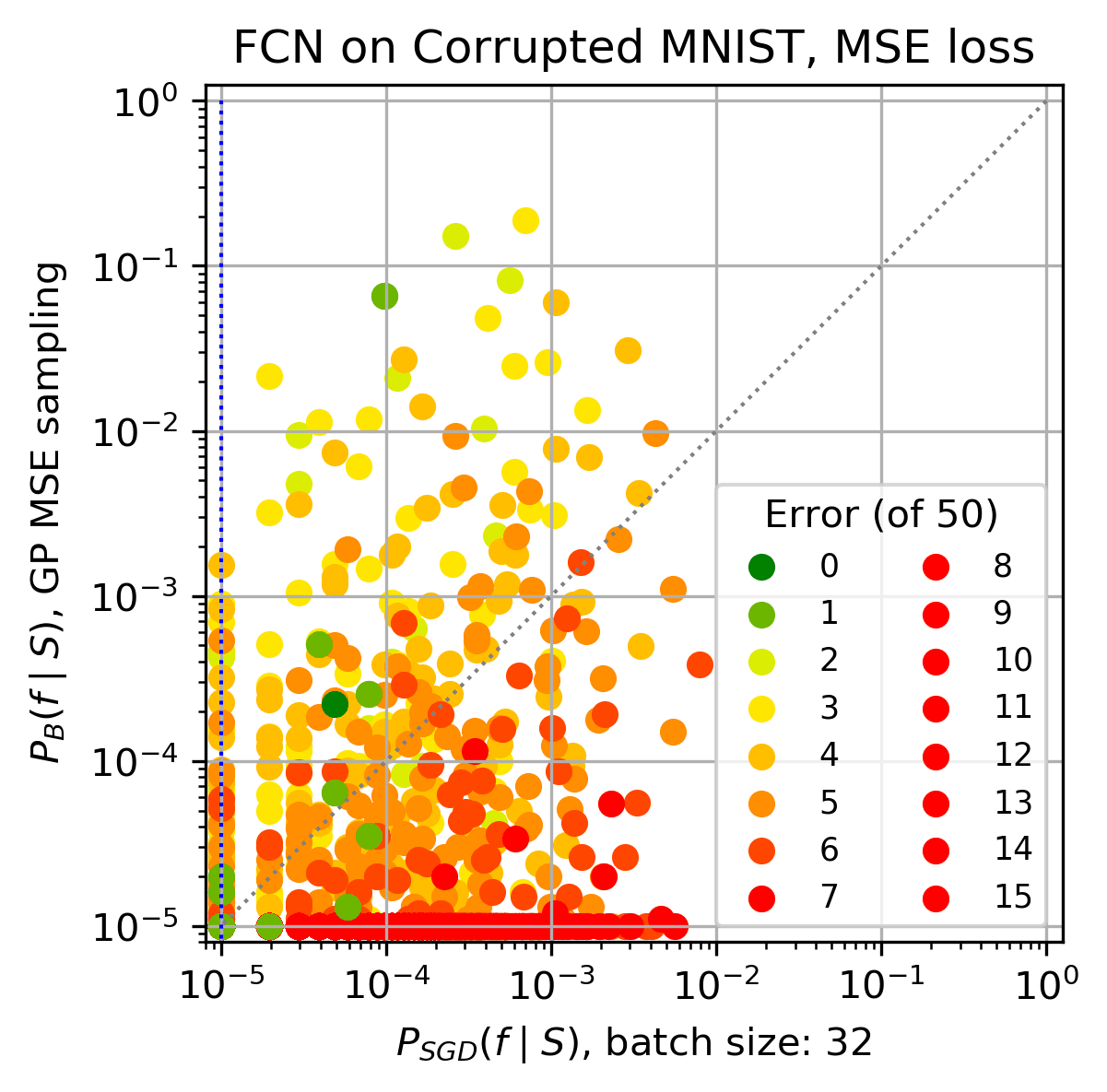}
 \caption{$\pb$ v.s.\ $\psgd$}
 \label{fig:sum_fig_corupted}
\end{subfigure}

\caption{\textbf{Schematic landscape and effects of randomising training labels. } (a) Cartoon of a biased loss-landscape. The three functions $f_1$, $f_2$ and $f_3$ all reach zero classification error (dashed red line), but due to bias in the parameter-function map, the ``basin size'' $V_B(f_1) \gg V_B(f_2),V_B(f_3)$, which typically implies that for the ``basins of attraction'' $V_{SGD}(f_1) \gg V_{SGD}(f_2),V_{SGD}(f_3)$. $\pb$ is proportional to $V_{B}(f)$, and $\psgd$ is proportional to $V_{SGD}(f)$. 
(b) $\pb$ (solid) and $\rho(\epsilon_G) \pb$ (dashed) v.s.\ $\epsilon_G$, for test set of size 100 and CE loss (as in \cref{fig:summary_fig_exp2_mnist}) but including label corruption $c$. 
(b) $\psgd$ v.s.\ $\pb$ on MNIST with a 2-layer 1024 node wide FCN with MSE loss, test set size 50, and 20\% of the training labels randomised ($\eg_{SGD}=13.4\%$ and $\eg_{GP}=5.80\%$). Here functions with frequency $< 10$ are also shown on the plot. The correlation is much less pronounced than for the unrandomised case shown in \cref{fig:summary_fig_exp1_mnist_mse}. Dots on the axes denote functions found by just one of the two methods.
Let $F$ be the set of functions found by both the optimiser and under GP sampling. Then $\sum_{f\in F}\pb=99.3\%$, and $\sum_{f\in F}\psgd=24.3\%$. In other words, while the Adam optimiser finds almost all functions with high $\pb$, it also finds many functions with low $\pb$. The much weaker bias under label corruption observed in (b) likely explains the weaker correlation between the Bayesian results and that of the optimiser found here. 
}\label{fig:random}
\end{figure}

At first sight it may seem rather surprising that SGD, which follows gradients down a complex loss-landscape, should converge on a function $f$ with a probability anything like the Bayesian posterior $\pb$ that upon random sampling of parameters, a DNN generates functions $f$ conditioned on $S$. Indeed, in the general case of an arbitrary learner we don't expect this correspondence to hold. However, as shown e.g.\ in Fig 1, $\pb$ is orders of magnitude larger for functions with small generalisation error than it is for functions with poor generalisation. As explained in \cref{subapp:simplicity_bias,rel:simplicity_bias_generalisation}, such an exponential bias towards low complexity/low error functions can be expected on fairly general grounds~\citep{valle2018deep,mingard2019neural,dingle2018input,dingle2020generic}. If our null expectation is of a large variation in the prior probabilities, then the good correlation can be heuristically justified by a landscape picture~\citep{wales2003energy}, where $\pb$ is interpreted as the ``basin volume'' $V_B(f)$ (with measure $p_{\textrm{par}}(\theta)$) of function $f$), while $\psgd$ is interpreted as the ``basin of attraction'' $V_{SGD}(f)$, which is loosely defined as a measure of the set of initial parameters $\theta_i$ for which the optimiser converges to $f$ with high probability (this concept also found in related form in the dynamical systems literature~\citep{strogatz2018nonlinear}). If $V_B(f)$ varies over many orders of magnitude, then it seems reasonable to expect that $V_{SGD}(f)$ should correlate with $V_B(f)$, as illustrated schematically in~\cref{fig:Fig-schematic}. Such general intuitions about landscapes are widely held~\citep{wales2003energy,massen2007power,ballard2017energy}, and have also been put forward for the particular landscapes of deep learning; see in particular \citet{wu2017towards} who also argue that functions with good generalisation have larger basins of attraction.

Another source of intuition follows form a well trodden path linking basic concepts from statistical mechanics to optimisation and learning theory. For example, simple gradient descent (GD) with a small amount of white noise can be described by an over-damped Langevin equation~\citep{welling2011bayesian,smith2018bayesian,naveh2020predicting} that converges (under some light further conditions) to the Boltzmann distribution
The Boltzmann distribution can, in turn, be interpreted as being equivalent to a Bayesian posterior $P_B(f|S) \propto e^{S(f)-\beta E(f)}$~\citep{mackay2003information} where $S(f)$ is configurational ``entropy'' that counts the number of states that generate $f$ and encodes the prior, and $E(f)$ represents the energy, encoding the log likelihood or loss function. For SGD the equivalent coarse-grained differential equation reduces to Langevin equation with anisotropic noise~\citep{smith2018bayesian,zhang2018energy} and doesn't exactly converge to the Bayesian posterior~\citep{stephan2017stochastic,brosse2018promises}. Nevertheless, it has been conjectured that with small step size, SGD may approximate the Bayesian posterior \citep{naveh2020predicting,cohen2019learning}, as we empirically find in our experiments. These connections are rich and worth exploring further in this context. Nevertheless, some caution is needed with these analogies to statistical mechanics because they depend on assumptions which may only to hold on prohibitively long time-scales.

A better analogy may be to the ``arrival of the frequent'' phenomenon in evolutionary dynamics~\citep{schaper2014arrival}, which, like the ``basin of attraction'' arguments, does not require steady state. Instead it predicts which structures are likely to be \emph{reached first} by an evolutionary process. For RNA secondary structures, for example, it predicts that a stochastic evolutionary process will reach structures with a probability that to first order is proportional to the likelihood that uniform random sampling of genotypes produces the structure. Indeed, this phenomenon -- where the probability upon random sampling predicts the outcomes of a complex search process -- can be observed in naturally occurring RNA~\citep{dingle2015structure}, the result of evolutionary dynamics. This type of non-equilibrium analysis may be more relevant for the way we train most of the DNNs in this paper, since we stop the first time 0 training error is reached. The analogy between these evolutionary results with what we observe for SGD is intriguing, but needs further exploration.

To illustrate the effect of the amount of bias in the posterior, we randomise labels for MNIST and calculate the $\pb$. As we can see in~\cref{fig:sf3:fake_multi}, this results in a less strongly biased posterior. The mean log-probability $\langle \log(\pb) \rangle$ v.s.\ $\epsilon_G$ curve becomes less steep with increasing corruption 
For a relatively small fraction of low error functions to dominate, as they do for zero corruptions in~\cref{fig:summary_fig_exp1_mnist_mse}, the bias must be strong enough here to overcome the ``entropic'' factor $\rho(\epsilon_G)$. For the $20\%$ and $50 \%$ corruption this is clearly not the case, and a huge number of functions with larger error will dominate $\pb$ and $\psgd$. As can be seen in \cref{fig:sum_fig_corupted}, one effect of weaker bias is that the correlation between the optimiser and the Bayesian sampling is much less strong. This behaviour is consistent with the heuristic arguments above, which should only work if the differences in basin volumes are large enough to overcome the myriad other factors that can affect $\popt$. 

\section{Related work on inductive bias on neural networks}\label{app:related_work}

In this section we summarise some key aspects of the literature related to why DNNs exhibit good generalisation while overparameterised, expanding on some briefer remarks in \cref{sec1:intro}.

\subsection{The link between inductive bias and generalisation}\label{rel:inductive_bias_generalisation}

Much of the work on inductive biases in stochastic gradient descent (SGD) is framed as a discussion about generalisation. The two concepts are of course intimately related. Before discussing related work on inductive bias DNNs, it may be helpful to distinguish two different questions about generalisation: 

\begin{description}

\item[\textbf{1)}] \textbf{Question of over-parameterised generalisation}:  Why do DNNs generalise at all in the overparameterised regime, where classical learning theory doesn't guarantee generalisation?

\item[{\bf 2)}] \textbf{Question of fine-tuned generalisation}:  
Given that vanilla DNNs already generalise reasonably well, how can architecture choice and hyperparameter tuning further improve generalisation? 
\end{description}

\noindent The first question arises because among the functions that an overparameterised DNN can express, the number that can fit a training data set $S$, but generalise poorly, is typically many orders of magnitude larger than the number that achieve good generalisation. From classical learning theory we would therefore expect extremely poor generalisation. However, in practice it is often found that many DNN architectures, as long as they are expressive enough to fit the data, generalise sufficiently well to imply a significant inductive bias towards a small fraction of functions that generalise well.

This question is also related to the conundrum of why DNNs avoid the ``curse of dimensionality'', which relates to the poor generalisation that certain highly expressive non-parametric models have in high dimensions \citep{donoho2000high}. \citet{valle2018deep} argue that the curse of dimensionality is linked to a prior which is not sufficiently biased and that DNNs may avoid this problem by virtue of the strong bias in the prior.

The second question arises from two common experiences in DNN research. Firstly, changes in architecture can lead to important improvements in generalisation. For example, a CNN with max-pooling typically performs better than a vanilla FCN on image data. Secondly, hyperparameter tuning within a fixed architecture can lead to further improvements of generalisation. 
While these methods of improving generalisation are important in practice, the starting point is normally  a DNN that already has enough inductive bias to raise question 1) above. 
It is therefore important not to conflate the study of question 2) -- as vital as this may be to successful practical  implementations --- with the more general question of why DNNs generalise in the first place. 

\subsection{Related work on implicit bias in optimiser-trained networks}
\label{rel:optimiser_trained}

As mentioned in the introduction, there is an extensive literature on inductive biases in SGD. Much of this literature is empirical: improvements are observed when using particular tuned hyperparameters with variants of SGD. One of the most common rationalisation is in terms of ``flatness'' which is inspired by early work~\citep{hochreiter1997flat} who predicted that flatter minima would generalise better. Flatness is often measured using some combination of the eigenvalues of the Hessian matrix for a trained DNN. \citep{keskar2016large} showed that DNNs trained with small batch SGD generalise better than identical models trained with large batch SGD (by up to $5\%$), and also found a correlation between small batch size and minima that are less ``sharp'' (using not the eigenvalues of the Hessian but a more computationally tractable sensitivity measure). 
While these results are genuinely interesting, they are mainly relevant to issues raised by question 2 above. For example in~\citep{keskar2016large} the authors explicitly point out that their results are not about ``overfitting'' (e.g.\ question 1 above). 

The effects of changing hyperparameters can be subtle. For example, another series of recent papers~\citep{goyal2017accurate,hoffer2017train,smith2017don} suggest that better generalisation with small batch SGD may be caused by the fact that the number of optimisation steps per epoch decreases when the batch size increases. These studies showed that a similar improvement in generalisation performance to that found by reducing batch size can be created by increasing the learning rate, or by overtraining (i.e.\ by continuing to train after $100 \%$ accuracy has been reached). In particular, in \citep{hoffer2017train} it was argued that overtraining does not generally negatively impact generalisation, as naive expectations based on overfitting might suggest. These results also challenge some theoretical studies that suggested that SGD may control the capacity of the models by limiting the number of parameter updates \citep{brutzkus2017sgd}. 

In another interesting paper, \citet{zhang2018energy} derive a Langevin type equation for both SGD. And argue that in contrast to GD, the noise is anisotropic, and that this may explain why SGD is more likely to find ``flatter minima''. Similarly, \citet{jastrzebski2018finding} argue that isotropic SGD-induced noise also helps push the optimiser away from sharper minima. An important caveat to the work on sharpness can be found in the work of Dinh {\em et al.\ }\citep{dinh2017sharp} who use the non-negative homogeneity of the ReLU activation function to show that for a number of the measures used in the papers cited above, the ``flatness'' can be made arbitrarily large (or sharp) without changing the function (and therefore the generalisation performance) that the DNN expresses. This result suggests that care must be used when interpreting local measures of flatness. Finally in this vein, generalisation has also been linked to related concepts including low frequency \citep{rahaman2018spectral}, and to sensitivity to changes in the inputs \citep{arpit2017closer-arxiv,novak2018sensitivity}. 

There is much more literature on SGD induced inductive bias, but the upshot is that while fine-tuning optimiser hyperparameters can be very important for improving generalisation, and by implication, the inductive bias of a DNN, a complete understanding remains elusive. 
Moreover, where improvements are found, these tend to be in the class of answers to question 2) above. 
An important example of a paper on flatness that does explicitly address question 1 above is \citep{wu2017towards}, who show that generalisation trends for data with different levels of corruption correlates with the log of the product of the top 50 eigenvalues of the Hessian both for SGD and for GD trained networks. By heuristically linking their local flatness measure to the global basin volume, they make a very similar argument to the one we flesh out in more detail here, namely that the basin of attraction volume of ``good'' solutions is much larger than that of ``bad'' solutions that do not generalise well. 

Significant theoretical effort has been spent on extracting properties of a trained neural network that could be used to explain generalisation. By implication, these investigations should also help illuminate the nature of the implicit bias of trained networks. For example, investigators have attempted to use sensitivity to perturbations (whether in inputs or weights) to explain the generalisation performance either using a PAC-Bayesian analysis \citep{bartlett2017spectrally,dziugaite2017computing,neyshabur2018a}, or a compression approach \citep{arora2018stronger,zhou2018nonvacuous}. In contrast to the work described above that studies the specific effect of hyperparameter tuning on SGD, much of the work listed in this paragraph is directly applicable to question 1.
A very comprehensive review of this line of work empirically finds that the PAC-Bayesian sensitivity approaches seem the most promising \citep{jiang2019fantastic}, but no clear answer to the question 1 has emerged. 
 
The more theoretical side of the study of SGD has also seen recent progress. For example, \citep{soudry2018implicit} showed that SGD finds the max-margin solution in unregularised logistic regression, whilst it was shown in \citep{brutzkus2017sgd} that overparameterised DNNs trained with SGD avoid over-fitting on linearly separable data. More recently, \citep{NIPS2019_8847} proved agnostic generalisation bounds for SGD-trained DNNs (up to three layers), which impose less restrictive assumptions (on the data, architecture, and optimiser) than previous works. Such theoretical analyses may be a potentially fruitful source of new ideas to explain generalisation.

Another interesting direction is to investigate properties of the loss-landscape itself. Several studies have shown interesting parallels between the loss landscape of DNNs and the energy landscape of spin glasses \citep{choromanska2015loss,baity2019comparing,becker2020geometry}. While such insights may help explain why SGD works so well as an optimiser in these high dimensional spaces, it is at present less clear how these studies help explain question 1) above. 

A completely different theme builds on the concept of an information bottleneck \citep{tishby2015deep,shwartz2017opening} which suggest that generalisation arises from information compression in deeper layers, aided by SGD. However, recent work \citep{saxe2019information} suggests that the compression is strongly affected by activation functions used, suggesting again that this approach is not general enough to capture the implicit bias needed to answer question 1. We note that the debate about this theme is ongoing. 

Finally, it is important to note that simple vanilla gradient descent (GD), when it can be made to converge, does not differ that much (on the scale of question 1 above) from SGD and its variants in generalisation performance~\citep{keskar2016large,wu2017towards,zhang2018energy,choi2019empirical}. Therefore if training with an optimiser itself generates the inductive bias needed to answer question 1, that bias must already largely be present in simple GD.

\subsection{Related work on implicit bias in random neural networks}
\label{subapp:simplicity_bias}

We briefly review work inspired by a powerful result from algorithmic information theory (AIT) called the coding theorem~\citep{li2008introduction}. First derived by Levin~\citep{levin1974laws}, and building on concepts pioneered by Solomonoff~\citep{solomonoff1964formal}, it is closely related to more recent bound applicable to a wider range of input-output maps~\citep{dingle2018input,dingle2020generic}. This bound predicts (under certain fairly general conditions that the maps must fulfil) that upon randomly sampling the parameters of an input-output map $M$, the probability $P(f)$ of obtaining output $f$ can be bounded as
\begin{align}\label{eqn:levin}
 P(f)\leq 2^{-K(f|M)+\mathcal{O}(1)} \approx 2^{-a \tilde{K}(f) + b}
\end{align}
where $K(f)$ is the Kolmogorov complexity of $f$, the $\mathcal{O}(1)$ terms do not depend on the outputs (at least asymptotically), $\tilde{K}(f)$ is a suitable approximation to $K(f)$ and $a$ and $b$ are parameters that depend on the map, but not on $f$. The computable bound was empirically shown to work remarkably well for a wide range of input-output maps from across science and engineering \citep{dingle2018input}, giving confidence that it should be widely applicable, at least for maps that satisfy the conditions needed for it to apply. In addition, a statistical lower-bound can be derived that predicts that most of the probability weight will lie relatively close to the bound~\citep{dingle2020generic}. 

The application of this bound to DNNs was first shown in~\citep{valle2018deep}. We note that the input-output map of interest is not the map from inputs to DNN outputs, but rather 
the map from the network parameters to the function $f$ it produces on inputs $\mathcal{X}$ which was described in \cref{def:PFM}. The prediction of \cref{eqn:levin} for a DNN with parameters sampled randomly (from, for example, truncated i.i.d.\ Gaussians) is that, if the parameter-function map is sufficiently biased, then the probability of the DNN producing a function $f$ on input data $x_{i=0}^n$ drops exponentially with increasing complexity of the function $f$. Note that technically we should write $f$ as $f|\mathcal{X}$ to indicate the dependence of the function modelled by the DNN on the inputs $\mathcal{X}$. We also note that the AIT bound of \cref{eqn:levin} on its own does not force a map to be biased. It still holds for a uniform distribution. But if the map is biased, then it will be biased according to \cref{eqn:levin}.

In \citep{valle2018deep} it was shown empirically that this very general prediction of~\cref{eqn:levin} holds for the $P(f)$ of a number of different DNNs. This testing was achieved both via direct sampling of the parameters of a small DNN on Boolean inputs and with NNGP calculations for more complex systems. In a complementary approach~\citep{mingard2019neural} some exact results were proven for simplified networks, that are also consistent with the bound of~\cref{eqn:levin}. In particular, they proved that for a perceptron with no bias term, upon randomly sampling the parameters (with a distribution satisfying certain weak assumptions), any value of class-imbalance was equally likely. There are many fewer functions with high class imbalance (low ``entropy'') than low class imbalance. Low entropy implies low $K(f)$ (but not the other way around). Thus, these results imply a bias of $P(f)$ towards certain simple functions. They also proved that for infinite-width ReLU DNNs, this bias becomes monotonically stronger as the number of layers grows. A different direction was pursued in~\citep{de2018random}, who showed that, upon randomly sampling the parameters of a ReLU DNN acting on Boolean inputs, the functions obtained had an average sensitivity to inputs which is much lower than if randomly sampling functions. Functions with low input sensitivity are also simple, thus proving another manifestation of simplicity bias present in these systems.

On the other hand, in a recent paper~\citep{yang2019fine}, it was shown that for DNNs with activation functions such as \textit{Erf} and \textit{Tanh}, the bias starts to disappear as the system enters the ``chaotic regime'', which happens for weight variances above a certain threshold, as the depth grows \citep{poole2016exponential} (note that ReLU networks don't have such a chaotic regime). While these hyperparameters are not typically used for DNNs, they do show that there exist regimes where there is no simplicity bias. Note that the AIT coding theorem bound~\cref{eqn:levin} still holds, but $P(f)$ is simply approaching a uniform distribution, and the bound becomes loose for small complexity. These results are also interesting because, if the bias becomes weaker, then it may also be the case that the correlation between $\pb$ and $\psgd$ starts to disappear, an effect we are currently investigating. 

\subsection{Related work comparing optimiser-trained and Bayesian neural networks}
\label{rel:opt_vs_bayes}

Another set of investigations studying random neural networks use important recent extensions of Neal's seminal proof~\citep{neal1994priors,neal2012bayesian} -- that a single-layer DNN with random i.i.d.\ weights is equivalent to a Gaussian process (GP)~\citep{mackay1998introduction} in the infinite width limit -- to multiple layers and architectures~\citep{lee2017deep,matthews2018gaussian,novak2018bayesian,garriga-alonso2018deep,yang2019tensor}. These studies have used this correspondence to effectively perform a very good approximation to exact Bayesian inference in DNNs. When they have compared them to SGD-trained DNNs~\citep{lee2017deep,matthews2018gaussian,novak2018bayesian}, the results have generally shown a close agreement between the generalisation performance of optimiser-trained DNNs and their corresponding Bayesian neural network Gaussian process (NNGP).

In this context another significant development is the introduction of the neural tangent kernel (NTK) \citep{jacot2018neural} which approximates the dynamics of an infinite width DNN with parameters that are trained by gradient descent in the limit of an infinitesimal learning rate. Recent comparisons to NNGPs show relatively similar performance of the NTK, see for example~\citep{arora2019exact,lee2019wide,novak2019neural}. While there are small performance differences, the overall agreement between NNGPs and the NTK or optimiser trained DNNs is close enough to suggest that the primary source of inductive bias needed for question~1 above is already present in the untrained network, and is essentially maintained under training dynamics. 

 
The linearisation of DNNs offered by NTK can also be used to prove that, in this regime, GD samples from the Bayesian posterior in a sample-then-optimise fashion. For linear regression models, \citet{matthewssample} showed that solutions after training GD with a Gaussian initialisation correspond to exact posterior samples. This idea is also related to Deep Ensembles which has been proposed to be ``approximately Bayesian'' in \citet{wilson2020bayesian}.

In this context, further indirect evidence comes from \citet{valle2018deep} who used a simple PAC-Bayesian bound \citep{mcallester1999pac} that applies to exact Bayesian inference, to predict the generalisation error of SGD-trained DNNs. The bound was shown to provide relatively tight predictions for optimiser-trained DNNs for an FCN and CNNs on MNIST, Fashion-MNIST and CIFAR-10. Moreover, this bound, which takes the Bayesian marginal likelihood as input, reproduced trends such as the increase in the generalisation error upon an increased fraction of randomised labels.

These lines of work serve as independent evidence to suggest that optimiser-trained DNNs behave very similarly to the same DNNs trained with Bayesian inference, and helped inspire the work in this paper, where we directly tackle this question.
These studies also suggest that the infinite-width limit may be enough to answer question~1, as the number of parameters in a DNN typically doesn't have a drastic effect on generalisation (as long as the network is expressive enough to fit the data).

\subsection{Related work on complexity of data, simplicity bias and generalisation}
\label{rel:simplicity_bias_generalisation}

In~\cref{subapp:simplicity_bias}, we discussed work showing that DNNs may have an inductive bias towards simple functions in their parameter-function map. Here, we briefly discuss how this ``simplicity bias'' concept may connect to generalisation. As implied by the no free lunch theorem~\citep{wolpert1994relationship}, a bias towards simplicity does not automatically imply good generalisation.
Instead certain key hypotheses about the data are needed, in particular that it is described by functions that are simple (in a similar sense to the inductive bias). Now the assumption that a more parsimonious hypothesis is more likely to be true has been influential since antiquity and is often articulated by invoking Occam's razor. However, the fundamental justification for this heuristic is disputed, see e.g.\ ~\citep{sober2015ockham} for an overview of the philosophical literature, e.g.~\citep{mackay1992bayesian, blumer1987occam, rasmussen2001occam, domingos1999role} for a set of different perspectives from the machine learning literature, and e.g.~\citep{rathmanner2011philosophical,sterkenburg2016solomonoff} for a spirited discussion of the links between the razor and concepts from AIT (pioneered in particular by Solomonoff). 

Studies which imply that data typically studied with DNNs is somehow ``simple'' include an influential paper \citep{lin2017does} invoking arguments, mainly from statistical mechanics, to argue that deep learning works well because the laws of physics typically select for function classes that are ``mathematically simple'', and so easy to learn. More direct studies have also demonstrated certain types of simplicity. For example, following on previous work in this vein, \citep{spigler2019asymptotic} calculated an effective dimension $d_{eff}\approx15$ for MNIST, which is much lower than the $28^2=784$ dimensional manifold in which the data is embedded. Individual numbers can have effective dimensions that are even lower, ranging from 7 to 13~\citep{hein2005intrinsic}. So the functions that fit MNIST data are much simpler than those that fit random data~\citep{goldt2019modelling}. An implicit bias towards simplicity may therefore improve generalisation for structured data, but it will likely have the opposite effect for more random data.

\section{Discussion}

We argue here that the inductive bias found in DNNs trained by SGD or related optimisers, is, to first order, determined by the parameter-function map of an untrained DNN. While on a log scale we find $\psgd \approx \pb$ there are also measurable second order deviations that are sensitive to hyperparameter tuning and optimiser choice. 

For the conundrum of why DNNs generalise at all in the overparameterised regime, our results strongly suggest that the solution must be found in the properties of $\pb$, and not in further biases introduced by SGD. Arguments that DNN priors are exponentially biased towards simple functions~\citep{valle2018deep,mingard2019neural,de2018random} may help explain the inductive bias of $\pb$, but more work needs to be done to explore the complex interplay between bias in the prior, the data, and generalisation. While they may not explain the fundamental conundrum above, second order deviations from $\pb$ are important in practice for further fine-tuning the generalisation performance.

Our function probability perspective also provides more fine-grained tools for the analysis of DNNs than simply comparing the average test error. This picture can facilitate the investigation of hyperparameter changes, or potentially also the study of techniques such as batch normalisation or dropout. It could assist in the design of new architectures or optimisers. 

It is not obvious how to determine the uncertainty in a prediction of a DNN model. However, if, as we argue here, SGD behaves like a Bayesian sampler, then this offers additional justification for using Deep Ensembles to measure this uncertainty in the case of DNNs \citep{wilson2020bayesian}. Our results could therefore make it easier to use neural networks in applications where it is important to be able to quantify prediction uncertainty

Most of our examples are for image classification. It would be interesting to study the related problem of using DNNs for regression. Sampling considerations means that it is easier to study $\psgd$ for smaller generalisation errors. It would be interesting to study systems with intrinsically larger $\eg$ within this picture as well. There the biasing effect of the optimiser may be larger. 

Finally, to study the correlation between $\pb$ and $\psgd$, we mainly used a fixed test and training set. While we did examine other test and training sets (see Appendices), this was mainly to confirm that our results were not an artefact of our particular choices. A promising future direction would be a Bayesian approach that includes averaging over training sets.

\bibliography{bib}
\appendix
\include{appendices}

\end{document}

%% file: appendices.tex
\section{Further detail for methods (\cref{sec:methodology})}\label{app:exampleresults}

In this Appendix we  provide further details and explanation of the methodology outlined in \cref{sec:methodology}. In particular, we discuss our  experiments in \cref{app:exampleresults:methodology} and the GP approximation for $\pb$ in \cref{appendix:GP_explanation}.

\subsection{Methodology in detail}\label{app:exampleresults:methodology}

For each experiment performed in the main text, we pick a DNN $\mathcal{N}$ (either FCN, CNN, or an LSTM) and a dataset $\mathcal{D}$ (MNIST, Fashion-MNIST or the IMDb dataset).  We also pick a fixed training set $S \subset \mathcal{D} $, and a fixed test set $E \subset \mathcal{D}$. Training sets are typically of size $10,000$ for FCN and CNN and $45,000$ for the LSTM.  Test sets are typically small, $100$ for the FCN and CNN, and $50$ for the LSTM.

\subsubsection{Using an optimiser to calculate $\bf \popt$}\label{subapp:exp1}
When calculating $\popt$ we first pick an optimiser OPT which is either plain SGD, or one of its derivatives: Adam, Adagrad, RMSprop, or Adadelta. Next we pick a loss-function, either mean-square error (MSE) or cross-entropy (CE). We also need to pick an initial parameter distribution $\tilde{P}_{par}(\theta)$ which we take to be from a truncated i.i.d.\ Gaussian distribution (the distribution from which the DNN $\mathcal{N}$ is randomly initialised, see \cref{eqn:basin_of_attraction}).


\begin{algorithm}
\caption{Calculating $\popt$}
\begin{algorithmic}\label{alg:mainexp}
\STATE \textbf{input:} DNN $\mathcal{N}$, training data $S$, test data $E$, optimiser $OPT$.
\STATE $F \leftarrow \langle \rangle$\qquad\qquad\qquad \COMMENT{the `functions' found during training}
\STATE \textbf{do} $n$ \textbf{times:}
\STATE \quad re-initialise the weights of $\mathcal{N}$ from an i.i.d.\ Gaussian distribution
\STATE \quad train $\mathcal{N}$ on $S$ until it reaches 100 \% training accuracy
\STATE \quad record the classification of $\mathcal{N}$ on $E$ and save it to $F$
\STATE $A \leftarrow \varnothing$\qquad\qquad\qquad \COMMENT{the frequency and `volume' of each `function'}
\FOR{each distinct $f \in F$}
    \STATE let $\rho_f$ be the frequency of $f$ in $F$
    \STATE calculate the probability $\popt=\rho_f/n$ of $f$ in $F$
    \STATE save $\popt$ to $A$
\ENDFOR
\STATE \textbf{return} $A$
\end{algorithmic}
\end{algorithm}

For fully-connected layers we used $\sigma_b=0$ and $\sigma_w=1/\sqrt{w}$ where $w$ is the width of the layer. For convolutional and LSTM layers we used the default initialisation provided by Keras 2.3.0\footnote{CNN: \url{https://keras.io/api/layers/convolution_layers/convolution2d/}\\LSTM: \url{https://keras.io/api/layers/recurrent_layers/lstm/}}. This specifies $\tilde{P}_{par}(\theta)$. As we see in \cref{alg:mainexp}, we then sample $n$ times from  $\tilde{P}_{par}(\theta)$, training each time with $OPT$  until the training error on $S$ is zero, at which point we record the function by what errors it makes on $E$. Note that if for some reason SGD does not converge, we don't count that run in order to have normalised distributions over functions.  We typically run between $n=10^4$ and $n=10^7$ runs (depending on the system).  Once the runs are finished, we compile all the empirical frequencies for the functions that are found. 

\subsubsection{Bayesian sampling for $\bf \pb$} \label{subapp:exp2}

We use the GP approximation to estimate the Bayesian posterior $\pb$.  Here we follow  \citep{valle2018deep} where this technique is explained (see also \cref{appendix:GP_explanation} for more details). We need to define a distribution $P_{par}(\theta)$ (the definition of the prior from which we calculate $\pb$, see \cref{eqn:Pf_vol}).
While there are some subtleties in how the prior distribution $P_{par}(\theta)$ relates to the initialisation distribution $\tilde{P}_{par}(\theta)$, we took a simple approach and defined $P_{par}$ to be the same as the corresponding $\tilde{P}_{par}(\theta)$, except we set $\sigma_b$ to be a small constant, typically $0.1\times \sigma_w$.

To estimate $\pb$, there is a small compromise that must be made here.  For a number of reasons, MSE loss is less popular for the kinds of classification problems we mainly study in this paper.  We also find that it typically takes significantly longer to train using an optimiser so that $\popt$ is more expensive to evaluate for MSE loss on the problems we study.   On the other hand, $\pb$ can be directly sampled $n$ times from the exact posterior (described in~\cref{app:GP_exaplanation:MSE}) using~\cref{alg:mainexp2a}), and so is relatively accurate and simple to evaluate.  

For CE loss, which is more frequently used for classification, and is also typically quicker to train than MSE for SGD and its variants, we need to use a further approximation.   Here we follow \citep{valle2018deep} and use the  expectation-propagation (EP) approximation for $\pb$. We then estimate the posterior log probabilities using the estimations of the log marginal likelihoods $\log P(S)$ (see \cref{app:GP_exp_0-1} for explanation) in ~\cref{alg:mainexp2b}, or we sample from the approximate EP posterior and use~\cref{alg:mainexp2a}. These two methods give very similar answers (\cref{fig:gpep_logp}).

\begin{algorithm}[H]
\caption{Calculating $\pb$ (via sampling)}
\begin{algorithmic}\label{alg:mainexp2a}
\STATE \textbf{input:} DNN $\mathcal{N}$, training data $S$, test data $E$.
\STATE $F \leftarrow \varnothing$ \qquad\qquad\qquad \COMMENT{functions sampled from the GP or GP/EP posterior}
\STATE \textbf{do} $n$ \textbf{times:}
\STATE \quad sample a function $f$ from the GP or GP/EP posterior when conditioning on $S$
\STATE \quad find $f$ on $E$
\STATE \quad save $f$ to $F$
\STATE $\mathfrak{R} \leftarrow \varnothing$ \qquad\qquad\qquad \COMMENT{function probabilities}
\FOR{each distinct $f \in F$}
    \STATE let $\rho_f$ be the frequency of $f$ in $F$
    \STATE calculate $\pb=\rho_f/n$
    \STATE save $\pb$ to $\mathfrak{R}$
\ENDFOR
\STATE \textbf{return} $\mathfrak{R}$
\end{algorithmic}
\end{algorithm}

\begin{algorithm}[H]
\caption{Calculating $\pb$ for specific $f$ via the ratio of likelihoods approximation}
\begin{algorithmic}\label{alg:mainexp2b}
\STATE \textbf{input:} DNN $\mathcal{N}$, training data $S$, test data $E$, optimiser $OPT$, set of functions $F$ (from \cref{alg:mainexp})
\FOR{each distinct $f \in F$}
    \STATE let $\rho_f$ be the frequency of $f$ in $F$
    \STATE use GP or GP/EP approximation to estimate $\pb$ of $f$ using the ratio of likelihoods approximation.
    \STATE save $\pb$ to $A$
\ENDFOR
\STATE \textbf{return} $A$
\end{algorithmic}
\end{algorithm}

\subsubsection{Calculating $\pb$ for functions with a wider range of $\epsilon_G$}
\label{subapp:exp3}

Given a training dataset $S$ and test dataset $E$ we can generate a random sample of different partial functions with varying levels of error on $E$ (by taking the test set classification and corrupting some percentage of labels). We can then use the GP/EP approximation to estimate $\pb$. We typically sample 20 examples for each number of errors. The averages are taken on the logs of the probabilities, and error bars on plots are 2$\sigma$, where $\sigma$ is the standard deviation.  Note that the vast majority of functions have such small probabilities, that it is not feasible to estimate their $\popt$ (nor use \cref{alg:mainexp2a})

While the experiments  will be informative for how the space is biased, it does not guarantee that all high-probability functions will be found. Since these functions affect generalisation the most, we rely on the results from~\cref{alg:mainexp2a} to check that there are no high-probability functions that are missed.

\begin{algorithm}[H]
\caption{Calculating $\pb$ for larger range of $\epsilon_G$}
\begin{algorithmic}\label{alg:mainexp3}
\STATE \textbf{input:} DNN $\mathcal{N}$, training data $S$, test data $E$.
\FOR{$\epsilon \in \{0.0, 0.5, \dots 1.0\}$}
    \STATE $V_\epsilon \leftarrow \langle \rangle$
    \STATE generate classification $c$ with error $\epsilon$ on $E$ (by randomly choosing $|E|\times\epsilon$ distinct labels in the correct function (restricted to the test set) to switch to incorrect).
    \STATE use GP/EP and \enquote{ratio of likelihoods} approximation to estimate the $\pb$ of $c$
    \STATE save the relative volume $\pb$ of $f$ to $V_\epsilon$
\ENDFOR
\STATE \textbf{return} $V_{0.0} \dots V_{1.0}$
\end{algorithmic}
\end{algorithm}

We here make a few more remarks for interpreting the results in experiments where we generate functions for a wide range of errors, such as  those in  \cref{fig:summary_fig_exp2_mnist}.
Firstly, as shown in \cref{fig:exp2_histogram2} in \cref{app:notes_on_exp}, there can be a wide variation in the probabilities $p_i$ of misclassifying each of the 100 images in this test set. The generalisation error is therefore dominated by a small number of harder to classify images. This means that the probabilities of functions for a fixed $\epsilon_G$ can vary a lot. This explains why we find that even though the highest probability function in \cref{fig:summary_fig_exp1_mnist_mse} is a 1-error function, on average the probability for 1-error functions in \cref{fig:summary_fig_exp2_mnist} is lower than that of the 0-error function. The high variance in $\pb$ within the functions of fixed $\epsilon_G$, as well as the EP approximation also means that the estimates of $\langle P_B(f|S) \rangle_{\epsilon_G}$ may be less accurate. For \cref{fig:summary_fig_exp2_mnist}, $\sum_\epsilon \rho(\epsilon) \langle P_B(f|S) \rangle_{\epsilon_G} \approx 0.1$, which is not far off the correct value of $1$. Keeping in mind that we may be missing some higher probability outliers in the average due to finite sampling, this agreement is encouraging. In short, although the quantitative values may not be fully accurate, the results in \cref{fig:summary_fig_exp2_mnist} are indicative of the strong exponential trend towards low probability with increasing error. 

A third point of clarification is that in \cref{fig:summary_fig_exp2_mnist} we used CE loss rather than MSE loss. We made this choice because we need to estimate very small $\pb$ values, for which we need to use the \enquote{ratio of likelihoods} approximation. While we only described how to use EP for CE loss, we could also use the EP approximation to estimate $\pb$ for the analytical posterior of the GP with MSE loss. However, from some preliminary tests, the EP approximation introduced significant systematic errors and in particular it didn't show better results than the EP/\enquote{ratio of likelihoods} method for CE loss. In \cref{appendix:GP_GPEP_error}, we can also see that the differences between $\pb$ for MSE and CE loss are probably of less than a few orders of magnitude, and would not significantly affect the results in \cref{fig:summary_fig_exp2_mnist}.

\subsubsection{Further notes on methods }\label{subapp:notes_on_experiments}
When $\pb$ and $\popt$ are  obtained by sampling, we  typically sample between $10^5$ and $10^7$ times. To avoid finite sampling effects, we place any functions found with frequencies $<10$ on the axes of our graphs. However, those functions are included in calculations involving generalisation errors, although typically they contribute very little because they are by definition low probability.

We will also regularly provide values for $\sum_{f\in F}\pb$ and $\sum_{f\in F}\popt$ in our figures comparing $\popt$ with $\pb$ (only when $\pb$ is obtained by direct sampling), where $F$ is the set of functions found by both GP sampling and by the optimiser (within a finite number of samples from the GP and optimiser-trained DNNs, typically $10^4-10^6$).
A value of $\sum \pb$ close to $1$ indicates that almost all functions with high $\pb$ are found by the optimiser.  Similarly, a high value of $\popt$ implies that GP sampling finds almost all functions with high $\popt$ found by the optimiser. Note that for the EP approximation needed for CE loss, we directly calculate the $\pb$ for functions found by the optimiser, and so a $\sum \popt$ is not defined.     

Finally, when values for the generalisation error $\eg$ are depicted in the graphs showing $\pb$ v.s.\ $\popt$, $\eg$ refers to the generalisation error of the optimiser. When $\eg$ is presented in graphs with only $\pb$ it refers to the $\eg$ from GP sampling, or alternatively we use the symbol $\eg_{GP}$.







\subsection{Description of Gaussian process calculations}\label{appendix:GP_explanation}

In this section, we provide more details on the Gaussian process (GP) methods used in the paper. We follow the general Bayesian formalism introduced in \cref{sec:preliminary_definitions}.

\subsubsection{GP with Mean Squared-Error (MSE) loss}\label{app:GP_exaplanation:MSE}

For this formulation, we consider the space of functions to be the space of real-valued functions on $\mathcal{X}$. These functions correspond to the real-valued pre-activations of the last layer of the neural network, before a final non-linearity (like softmax or a step function) is applied. In the GP limit of DNNs, these functions have a GP prior (the NNGP). We then use a Gaussian likelihood defined as
\begin{equation}
        P(S|f) = \prod_{i=1}^m \frac{1}{\sqrt{2\pi\sigma^2}}\exp{\left(\frac{1}{2\sigma^2}(f(x_i)-y_i)^2\right)},
\end{equation}
where $\sigma^2$ is the variance.

This likelihood allows us to analytically compute the exact posterior~\citep{rasmussen2004gaussian}. In the experiments in the paper we therefore sampled from this exact posterior, to get values of $f(x)\in \mathbb{R}$ at the test points, which were then thresholded at $0$ to find the predicted class label. We have chosen a small value of the variance $\sigma^2=0.002$, to simulate SGD achieving a small value of the MSE loss.

Note that under the standard assumption that training and test instances come from the same distribution, this algorithm may be considered to be not fully Bayesian in the sense that the training and test labels are treated differently (Gaussian likelihood in training points versus Bernoulli likelihood at test points).  Nevertheless, we believe that the differences are small.

\subsubsection{GP with 0-1 likelihood, EP, and ratio of likelihoods approximation}\label{app:GP_exp_0-1}

While MSE loss has the advantage that the GP is analytically tractable, it has the disadvantage that it is less principled and less commonly used for classification than the much more popular cross-entropy (CE) loss function.   Unfortunately the GP approximation to CE loss is more complex.  Here our formulation uses a Gaussian process prior over a space of real-valued ``latent'' functions $\tilde{f}$ on $\mathcal{X}$. We call these functions ``latent'' because the true space of functions we are interested in is the space of Boolean functions on $\mathcal{X}$, obtained after applying the threshold nonlinearity to the last layer pre-activations. Formally, $f$ is related to $\tilde{f}$ via a Bernoulli distribution with Heaviside linking function, as
\begin{equation*}
    P(f(x)=1) = \begin{cases} 1\textrm{ if }\tilde{f}(x)>0 \\
                            0\textrm{ otherwise }.
            \end{cases}
\end{equation*}
We then use the previously defined 0-1 likelihood (\cref{01likelihood}) applied to the a binary-valued function $f$ (taking values in $\{0,1\}$). We use this likelihood, arguing that it best approximates the behaviour of an optimiser which is trained (using cross-entropy loss) until it \emph{first} reaches $0$ training error, which is the case in all the experiments in the paper (except those labelled as ``overtraining''). Because of this we  informally refer to this method as ``using the CE loss'' throughout the paper.  Unfortunately, this likelihood makes the posterior of the GP analytically intractable. We therefore use a standard approximation technique known as expectation propagation (EP) \citep{rasmussen2004gaussian}, which  approximates the posterior over the latent function as a Gaussian, which we can sample from and then use the Heaviside function to predict the binary labels at the test points. We use this technique to approximate posterior probabilities of the GP with 0-1 likelihood, by sampling.

The marginal likelihood can also be estimated with the EP algorithm~\citep{rasmussen2004gaussian}. This gives the probability of a labelling of a set of points. Remember that in the Bayesian formalism $S$ is essentially defined as the event that the input points $x_i$ in the training set have labels $y_i$. We can similarly identify a function $f$ with the event that the set of input points $x$ in the whole domain $\mathcal{X}$ have labels $f(x)$, which is analogous, and can thus be computed in the same manner as the marginal likelihood! The posterior in \cref{eqn:bayesian_update} for a function $f$ which is compatible with $S$, can then be simply expressed as
\begin{equation*}
    P(f|S)=\frac{P(f)}{P(S)},
\end{equation*}
where both $P(f)$ and $P(S)$ are readily computed using the EP algorithm to approximate marginal likelihood. 
This will be referred to as the \enquote{ratio of likelihoods} approximation in the appendices.
We have found that this method gives very similar results to the estimates using sampling from the approximate posterior obtained from EP (see \cref{fig:gpep_logp}). When we refer to the EP approximation, we imply that we are using the ratio of $P(f)$ and $P(S)$, each determined using the EP approximation, unless stated otherwise. 

For the LSTM experiments, we used a smooth version of the 0-1 likelihood, analogous to using standard cross-entropy loss rather than miss-classification error. This was because the EP approximation was numerically unstable with the 0-1 likelihood for this system. The smooth version is described in \cref{appendix:GP_GPEP_error}, and we empirically found that the two gave very similar estimates of probability.

\subsubsection{Empirical results concerning the GP approximations}\label{appendix:GP_GPEP_error}

\begin{figure}[H]
\centering

\begin{subfigure}[b]{0.31\textwidth}
    \includegraphics[width=\textwidth]{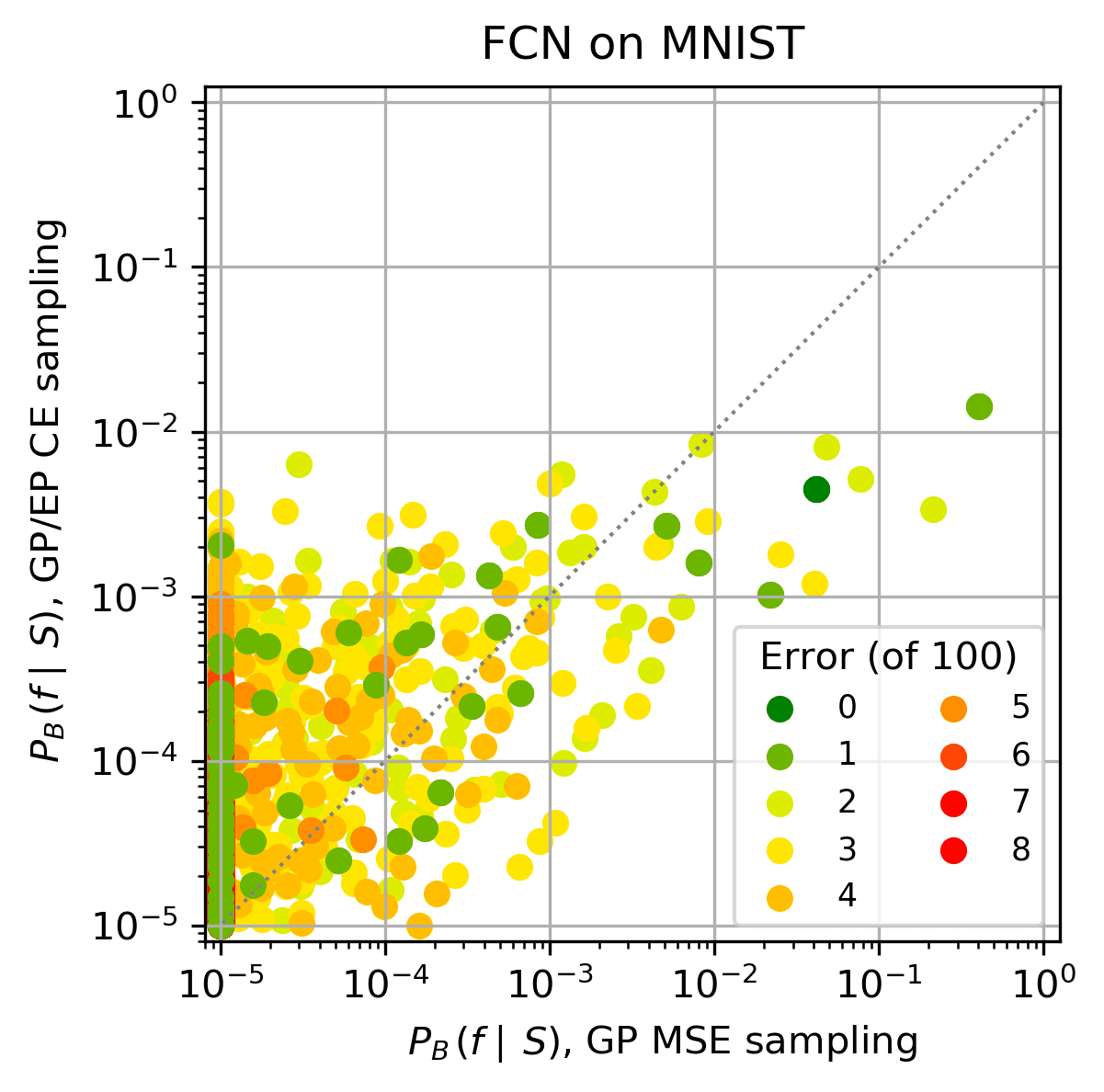}
    \caption{$\pb$(CE/EP) versus $\pb$ (MSE)}
    \label{fig:gpepgp_sample_sample}
\end{subfigure}
~~
\begin{subfigure}[b]{0.31\textwidth}
    \includegraphics[width=\textwidth]{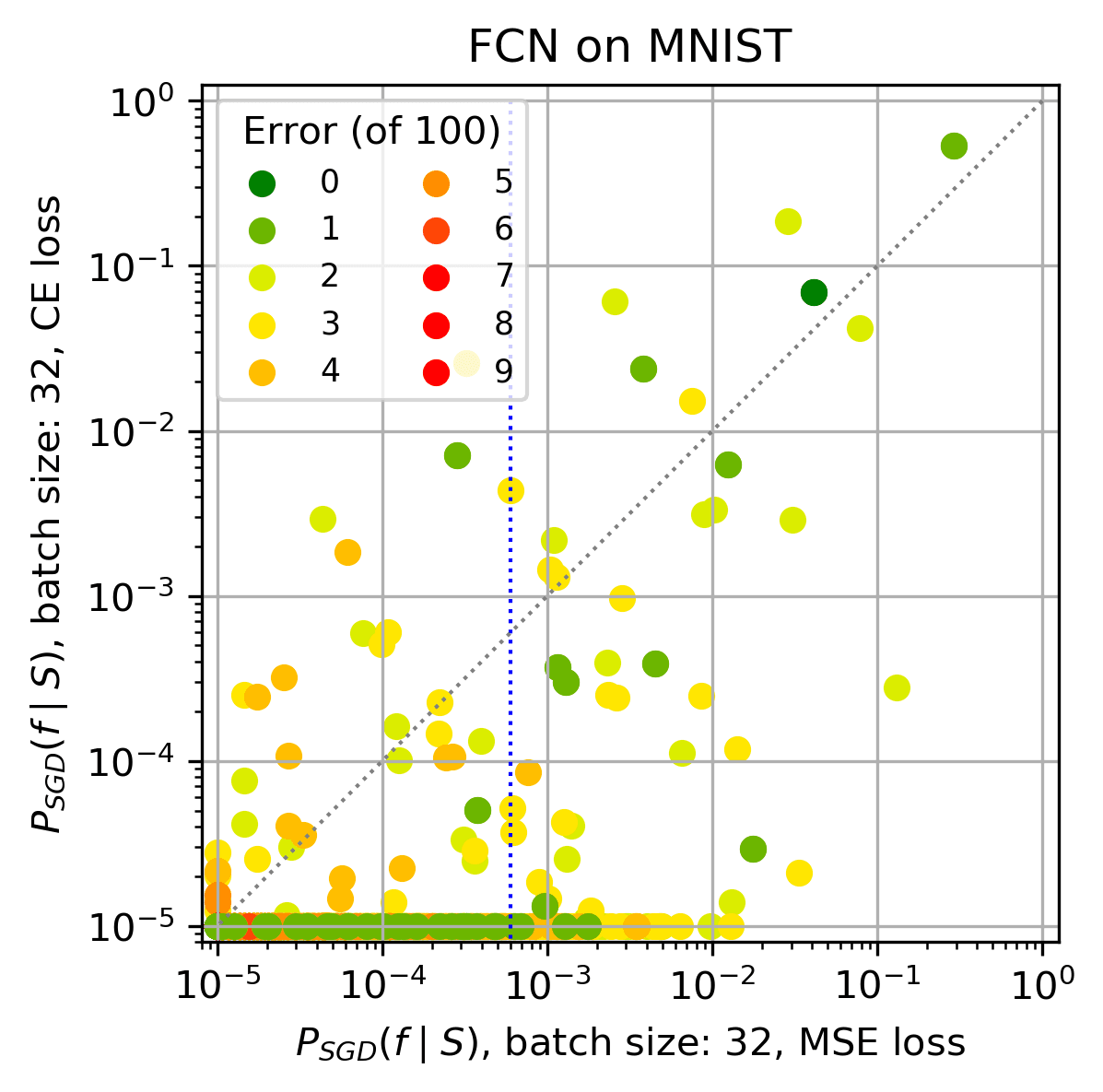}
    \caption{$\psgd$(CE) versus $\psgd$ (MSE)}
    \label{fig:gpepgp_sgd_sgd}
\end{subfigure}
~~
\begin{subfigure}[b]{0.31\textwidth}    
    \includegraphics[width=\textwidth]{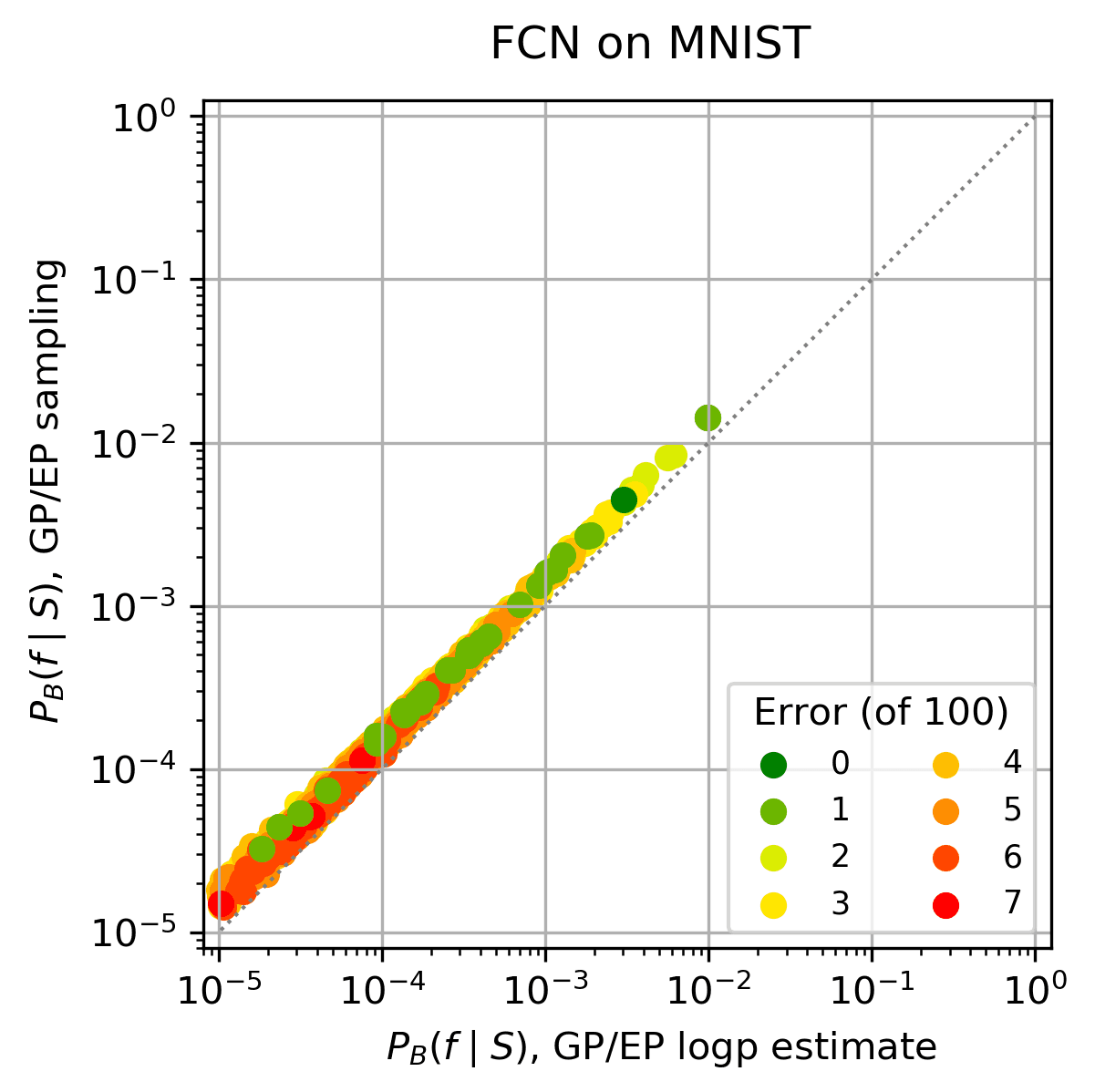}
    \caption{$\pb$(CE/EP) sampled v.s. ratio of likelihoods estimate}
    \label{fig:gpep_logp}
\end{subfigure}

\caption{{\bf Comparing GP approximations with CE and MSE loss for FCN on MNIST} In (a) we compare the behaviour of the GP approximations for $\pb$ with MSE loss to the GP/EP approximation with CE loss. We sampled from the GP MSE posterior, and the GP/EP CE posterior distribution $10^6$ times. It is expected that the two measures should diverge somewhat due to details of the loss function on the training data.  
(b) compares $\psgd$ with MSE loss and $\psgd$ with CE loss. Functions with high $\psgd$ are scattered around $y=x$. This implies that the loss function does not substantially affect $\psgd$ on average.  
Note that in (a) the two methods correlate, but that 1) the GP-EP is systematically lower than the MSE, and 2) that the slope is below $x=y$.  These two trends are, we believe, more general for the GP-EP approximation on CE loss.  
(c) Here  we compare the GP/EP $\log(p)$ approximation with GP/EP sampling. For this figure, we  use only functions found by Adam in $10^6$ samples, and compare probabilities found by the GP/EP $\log(p)$ approximation to those found by GP/EP sampling. We use both methods in this paper, but as is clear from the above figure, there is not much difference between them for functions with high $\pb$.}
\label{fig:gpepgp_summary}
\end{figure}

In this section we compare the behaviour of the GP approximations (and SGD) with different loss functions. We have argued that the GP/EP approximation underestimates probabilities by a power law (that is approximately linear in log-log; see \citep{valle2018deep} for more details).  Our results with the EP approximation are consistent with this expectation.  As detailed in \cref{app:GP_exp_0-1,app:GP_exaplanation:MSE}, there are subtle differences in the way the GP approximation with MSE loss and the GP/EP approximation with CE loss calculate their respective estimates for $\pb$. However, the (latent) function has the same prior in both cases, so we may expect the posterior $\pb$ to correlate. And it is clear from \cref{fig:gpepgp_sample_sample} that they do indeed correlate.  We believe  that the correlation not being centred around $y=x$  is  predominately due to the EP approximation because  apart from scatter, the behaviour of SGD with the two loss functions is centred around $y=x$. In \cref{app:bool_func}, we also compare the EP and MSE approximations with a estimation via direct sampling (and thus with controlled error) of the posterior probabilities for 0-1 likelihood for small Boolean function datasets, where these computations are feasible. We indeed find that EP tends to underestimate posterior probabilities, specially for complex target functions.
Overall what we find is that the EP approximation does reasonably well on relative probabilities, but less well on absolute probabilities. 

To mitigate the effect that the EP approximation underestimates the probabilities, we perform a simple empirical regularisation.   For systems where we find that $\sum \pb \approx 1$ for the MSE approximation, we renormalise the $\pb$ from the EP approximation by a constant factor such that $\sum \pb =1$.   For most systems we study the effect of this regularisation procedure is relatively small on a log scale. This method facilitates the comparison with $\psgd$ because the errors in the absolute values are regularised in the same way for all systems. 
Note also that because we  sample to obtain $\psgd$ its  empirical frequencies automatically sum to 1.    If it were the case that the Bayesian sampling found a significant number of  different high probability functions, then this would be observed in a lack of correlation in the comparison with $\psgd$.    Instead, we find a strong correlation between $\psgd$ and $\pb$ calculated with the EP approximation, suggesting that the functions found are the dominant ones found by both methods, as is explicitly found to be the case for most instances of MSE loss that we studied. 
This regularisation is applied to all experiments (for ease of comparison), unless otherwise specified.

Specifically, for \cref{fig:train_size_main} the renormalisation constant was calculated for Adam without overtraining and batch size 128 (as it had the highest raw value for $\sum\pb$), and the probabilities in the other plots with FCN on MNIST are all adjusted by the multiplicative constant of 3.59, which is modest on the full log scale of the graphs.
For other plots, the probabilities were normalised. The two systems for which this renormalisation has a larger effect are the LSTM and the ionosphere dataset.  While for both systems the MSE sampling looks relatively close to $y=x$, the raw EP approximation has significantly lower probabilities.  The renormalisation factors were $1.15\times10^5$ and 97 respectively.  Smaller MSE experiments verify that there is a strong correlation between $\psgd$ and $\pb$ for these systems.

In \cref{app:GP_exp_0-1}, we described using a Heaviside linking function when using the 0-1 loss function in the EP approximation. To test this, we have also sampled $f$ following a Bernoulli distribution with a Probit linking function
to the latent $\tilde{f}$ (which is analogous to using a cross-entropy loss). To test the differences between the results (which we assume to be small), we tested 100 randomly selected functions (on MNIST) with $\eg$ ranging from 0\% to 100\%. Of these, the average difference between the results as a percentage of the magnitude of the log probabilities was $0.013\%$, and the maximum was $0.58\%$ for the FCN architecture.
We also compare the GP/EP sampling with GP/EP $\log{P(S)}$ approximation in \cref{fig:gpep_logp}. The $\log{P(S)}$ approximation is a further approximation that allows $\log{P(f)}$ to be extracted without requiring sampling. Unless otherwise specified, all results in the main text use the GP/EP $\log{P(S)}$ approximation rather than sampling.


\raggedbottom

\section{Notes on the distribution of MNIST data}\label{app:notes_on_exp}

\begin{figure}
\centering
\begin{subfigure}[b]{0.4\textwidth}
	\includegraphics[width=\textwidth]{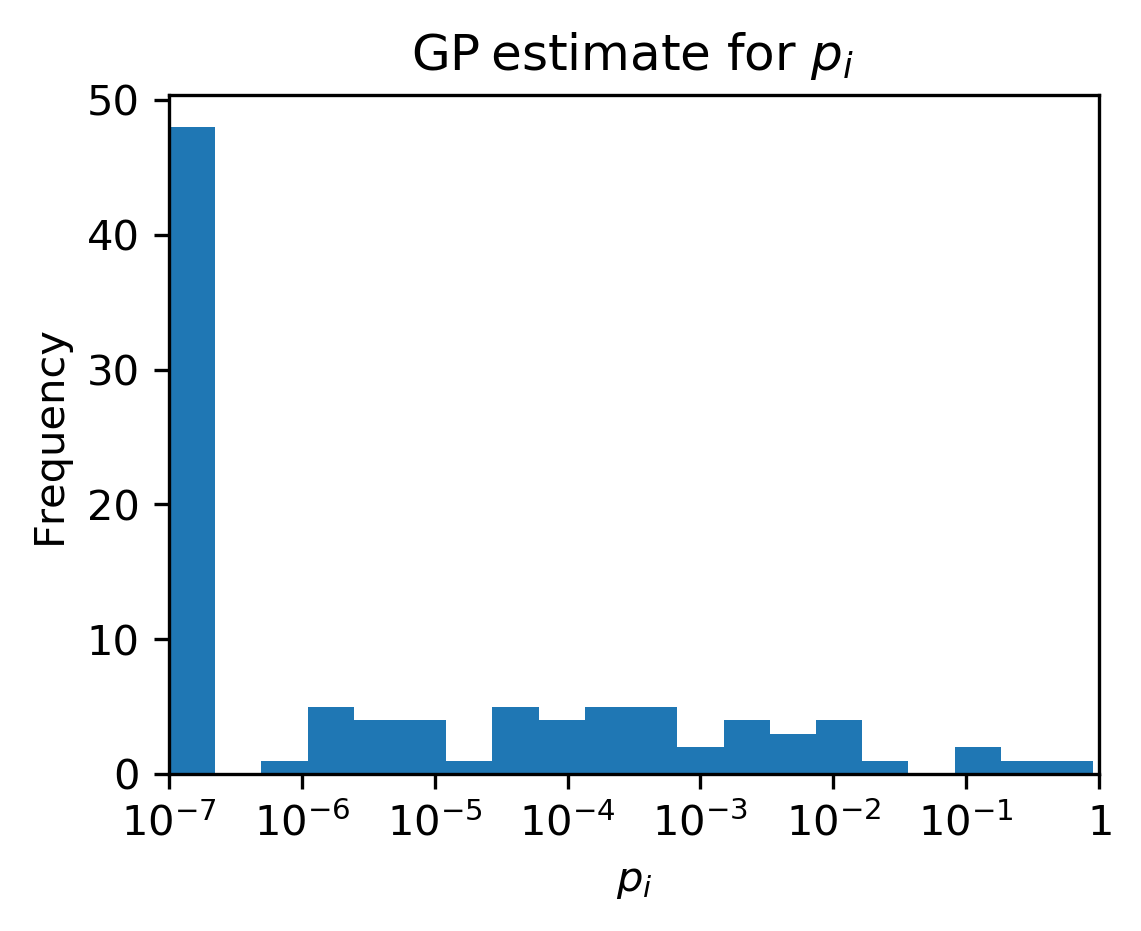}
    \caption{$p_i$ for GP}
    \label{fig:exp2_histogram2}
\end{subfigure}
~~
\begin{subfigure}[b]{0.4\textwidth}
    \includegraphics[width=\textwidth]{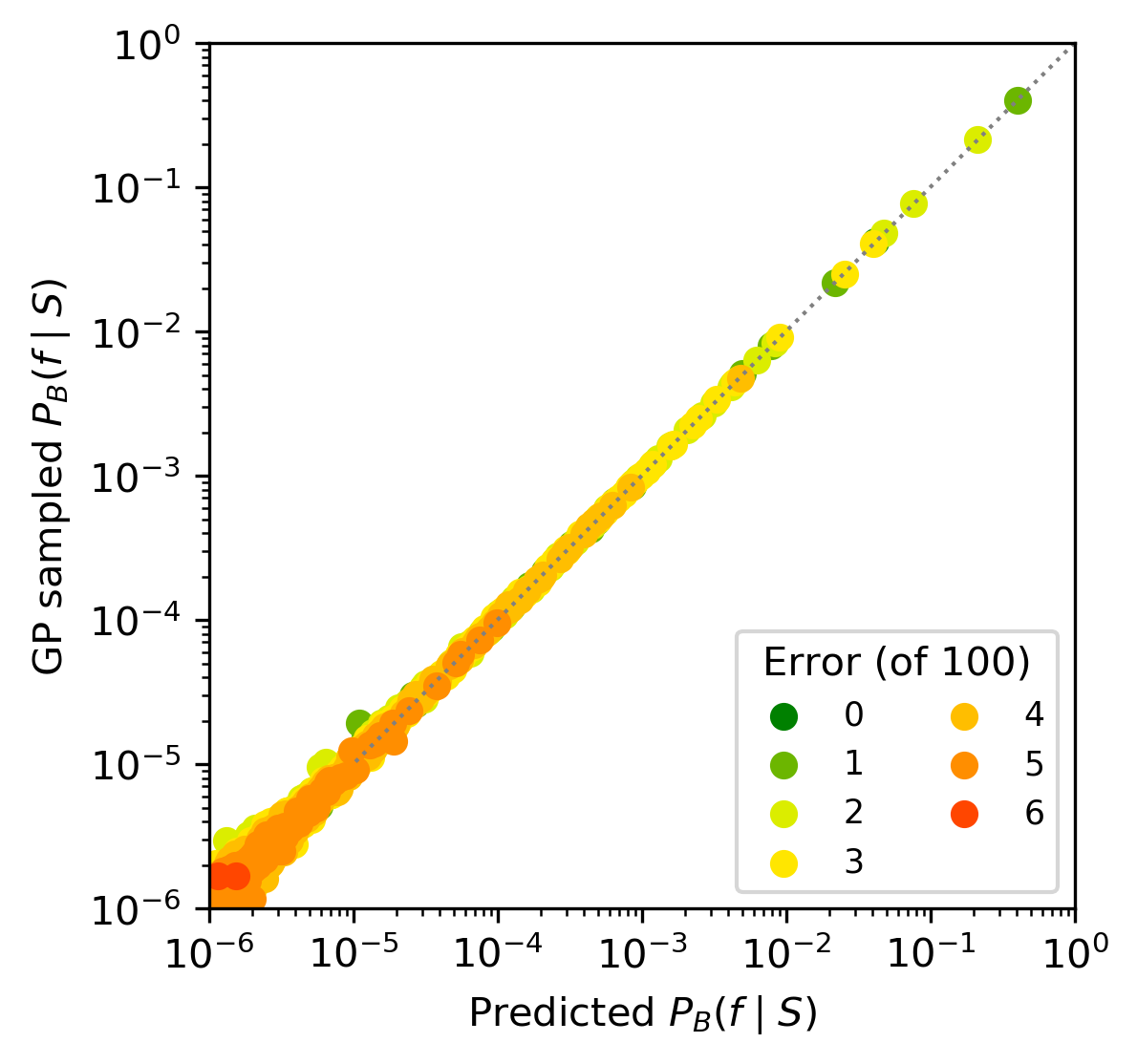}
    \caption{$\pb$ v.s.\ $p_i$ prediction.}
    \label{fig:exp2_corr}
\end{subfigure}

\begin{subfigure}[b]{0.4\textwidth}
	\includegraphics[width=\textwidth]{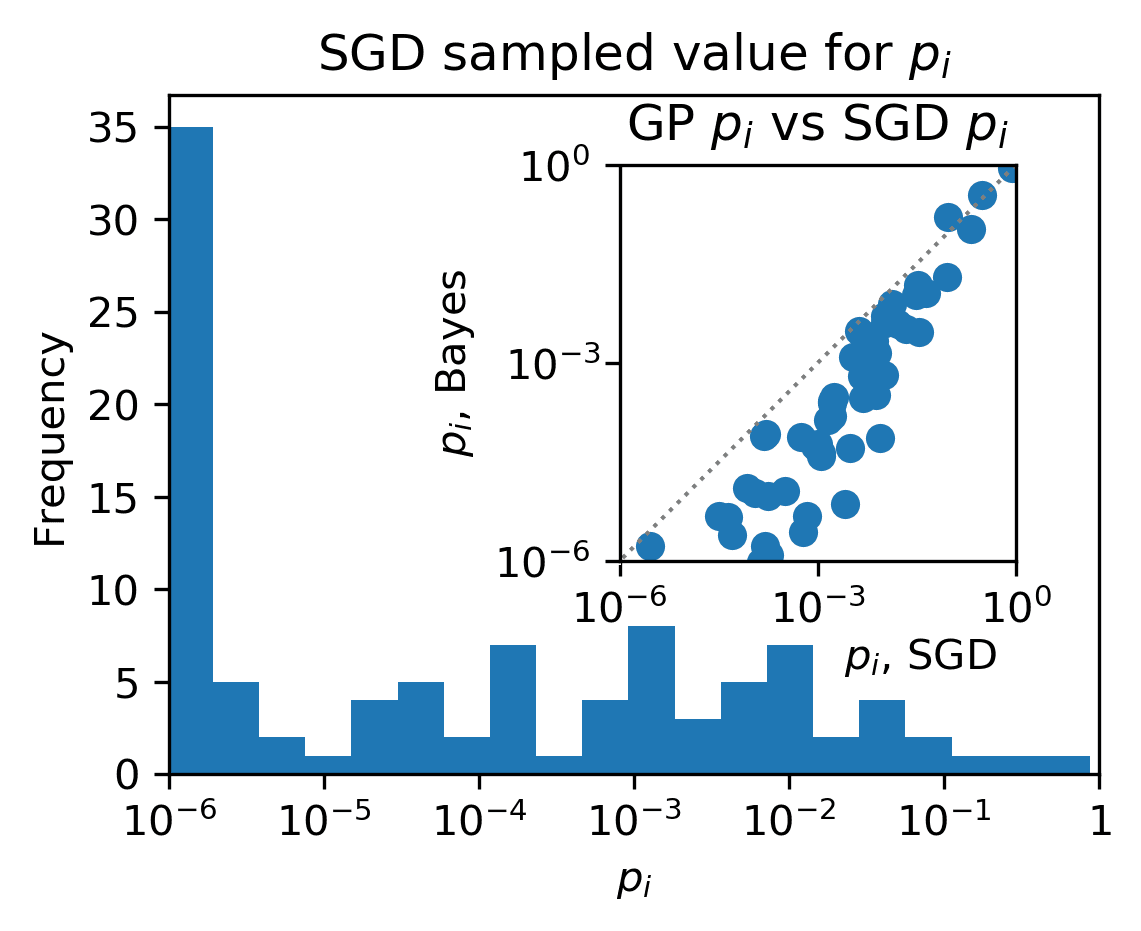}
    \caption{$p_i$ for SGD}
    \label{fig:exp2_histogram_sgd}
\end{subfigure}
~~
\begin{subfigure}[b]{0.4\textwidth}
    \includegraphics[width=\textwidth]{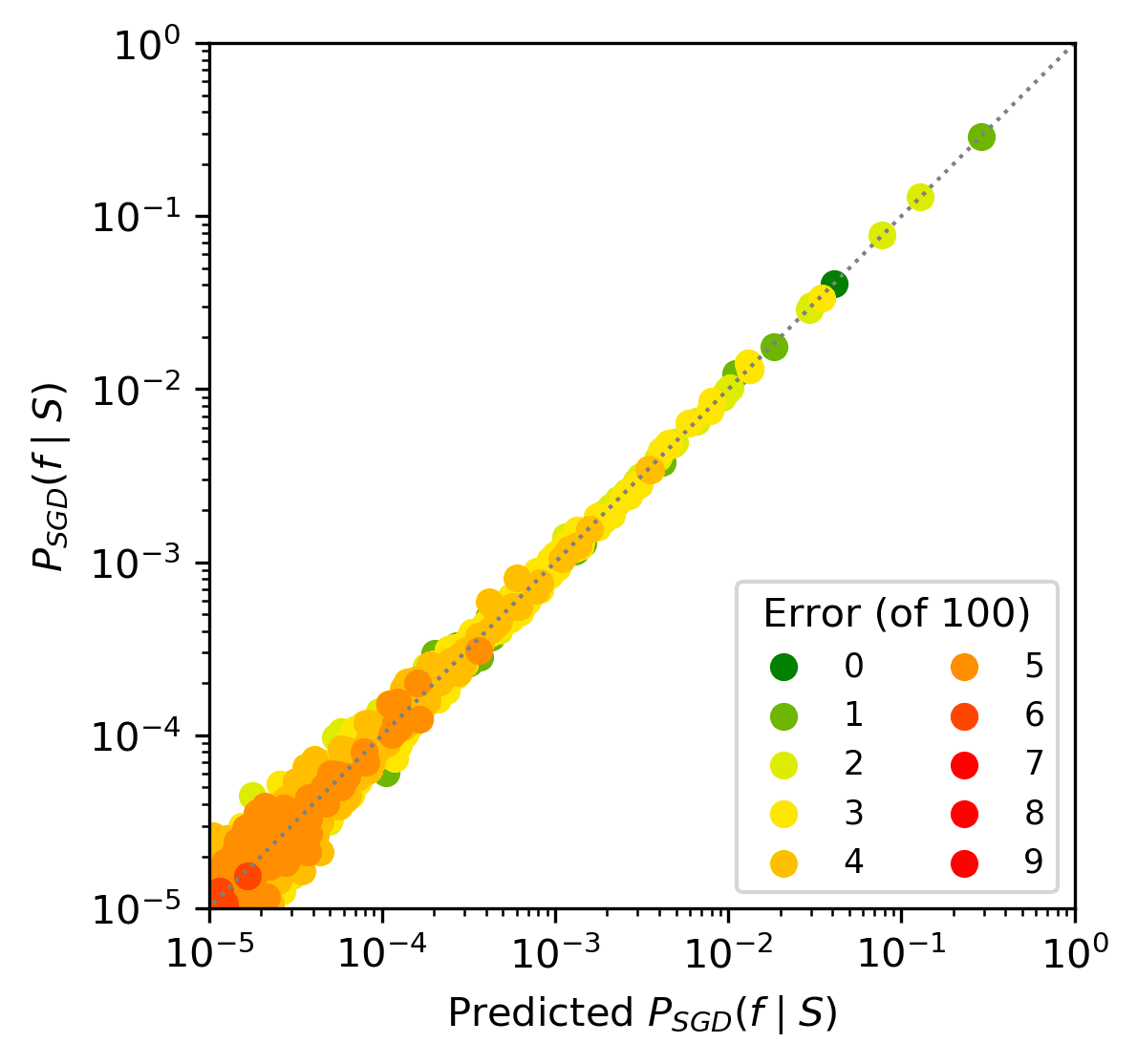}
    \caption{$\pb$ v.s.\ $p_i$ prediction.}
    \label{fig:exp2_corr_sgd}
\end{subfigure}

\caption{(a) shows the NNGP estimate for the values of $p_i$ for an FCN with MSE loss on MNIST, for the training set of size 10000 and test set of size 100 that we use in the main text. Clearly these vary over many orders of magnitude. The sample size is $10^7$, so frequencies were cut off at $10^{-7}$ (so functions in that bin have $p_i\leq 10^{-7}$). 20 bins in total. 
(b) calculates the probability of other error functions using the values of $p_i$  from \cref{fig:summary_fig_exp3_mnist} using the assumption that the images from (a) are independently distributed, and compares this prediction for $\pb$ to the value of $\pb$ obtained by direct GP MSE sampling (see \cref{fig:summary_fig_exp3_mnist} for the data). Each datapoint is for a specific function $f$, and clearly they are close to the $y=x$ line, implying that, at least for these higher probability functions found by direct sampling,  the images in this small test set are classified by the GP in a (close to) independent fashion. Figures 
(c) and (d) are the equivalent of (a) and (b),  but for SGD (see \cref{fig:summary_fig_exp1_mnist_mse} for the data). There were only $10^6$ samples, so (c) and (d) are cut off at one order of magnitude lower than (a) and (b). (c) includes an inset comparing the values for $p_i$ with GP MSE sampling from (a) and $p_i$ with SGD from the main part of (c).  The correlation  is fairly tight  for the highest $p_i$ which dominate the total probability mass, and this correlation helps explain the strong correlation seen for other numbers of errors throughout this paper.
}\label{fig:experiment_2_figures}
\end{figure}

Here, we show that  $p_i$,  the mean probability, for an FCN on MNIST with MSE loss function, that  either SGD or the  NNGP   classifies the $i$'th image in the test set $E$ (used throughout this paper, with $|E|=100$)
incorrectly, varies by many orders of magnitude.  We also show that images are classified  in an approximately independent manner. 

The values of $p_i$ for the MNIST test set which is used in the majority of this paper 
are given in \cref{fig:exp2_histogram2}.  Note that $(1/n)\sum_i p_i=\eg$).  As can be seen, there are a small number of images that have a much higher probability of being miss-classified.   It should be kept in mind that  the exact spectrum of $p_i$ will depend on which images are in the test set.  For example, in \cref{fig:test_set_dependence} we compare two test sets (albeit with CE-loss and the EP approximation). As can be observed, the zero error function has significantly higher probability in the second test set, and the highest 1-error function has lower probability than for the test set we use in the paper.   For test sets of this size such fluctuations are not unexpected.

Next we use the $p_i$ to calculate probabilities for functions with other sets of errors. 
More specifically, consider test sets $E_i$ containing only one image $I_i$ (where $i$ takes values in $\mathbb{N}$). 
Then calculating the probability of a function on $E=\{E_i\}_{i=0}^{i=|E|}$ for a test set of size $|E|$ can be done by multiplying the appropriate probabilities for functions on $E_i$. 
Applying the above method to this data is shown in \cref{fig:exp2_corr}. All the functions shown are very close to the $y=x$ line, indicating that the images are classified in an approximately independent fashion.

In all our examples  we observe a clear linear decay of the mean of  $\log(\pb)$ v.s.\ $\epsilon_G$.  If the $p_i$ were identical, then this would simply be what is expected for a Binomial distribution. For independent but different $p_i$ over images in a finite test set the distribution is called the Poisson Bionomial distribution.   Obviously its values depend on the exact distribution of $p_i$.  However, there exists a Chernoff bound
\begin{align}
p(\epsilon) \leq P[\mathcal{E}>\epsilon]\leq \exp(-\epsilon(\log(\epsilon/\eg)-1)-\eg),
\end{align}
where $p(\epsilon)$ is the pmf, $P[\mathcal{E}>\epsilon]$ is the cmf, and $\eg=(1/n)\sum_i p_i$ is the mean. On a log scale, this means
\begin{align}
    \log_{10}(p(\epsilon))\leq [-\epsilon(\log(\epsilon/\eg)-1)-\eg]\log_{10}(e),
\end{align}
which indicates an exponential like drop-off for $\epsilon\geq \eg$ (similar to what is observed in \cref{fig:summary_fig_exp1_mnist_mse}).
Of course this is an upper bound and the actual distribution can strongly depend on the full spectrum of $p_i$ values.  We are currently exploring these issues in more detail.  

\section{Effects of training set size}\label{app:training_set_size}

\begin{figure}[H]
\centering

\begin{subfigure}[b]{0.31\textwidth}
    \includegraphics[width=\textwidth]{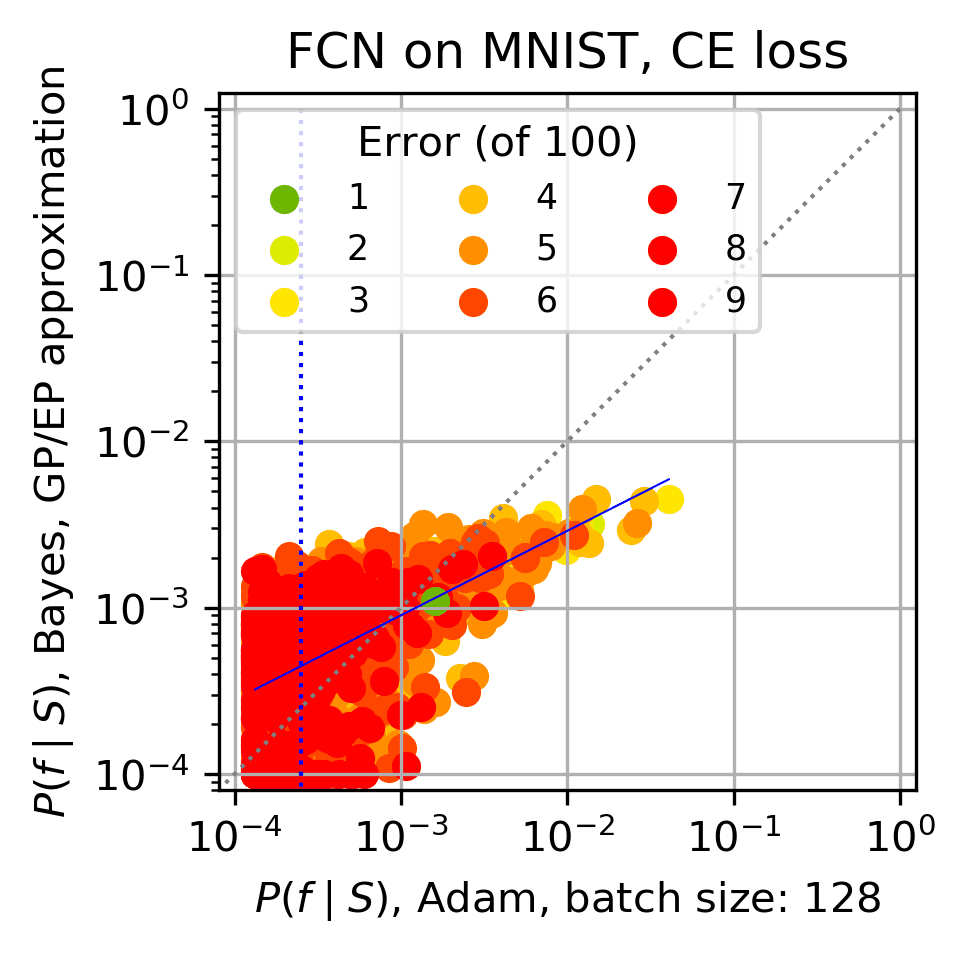}
    \caption{$|S|=1000$}
    \label{fig:mnist:gp_2_1k}
\end{subfigure}
~~
\begin{subfigure}[b]{0.31\textwidth}
    \includegraphics[width=\textwidth]{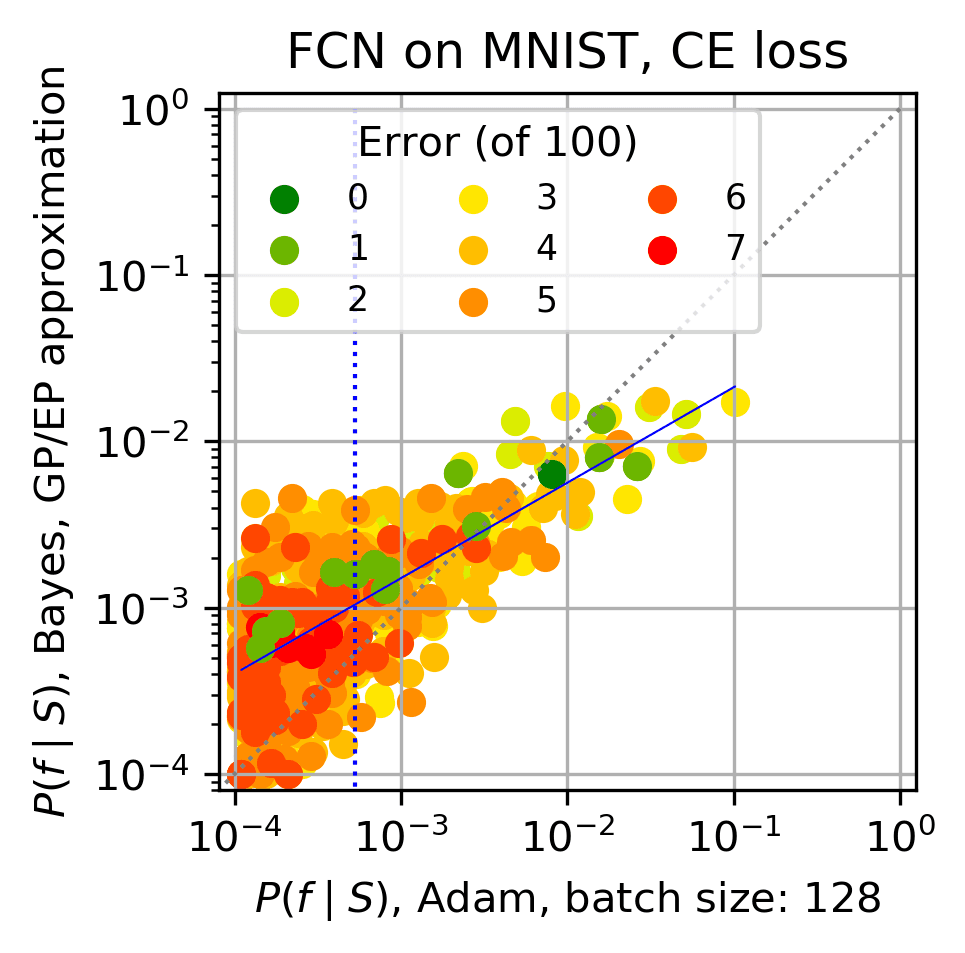}
    \caption{$|S|=5000$}
    \label{fig:mnist:gp_2_5k}
\end{subfigure}
~~
\begin{subfigure}[b]{0.31\textwidth}
    \includegraphics[width=\textwidth]{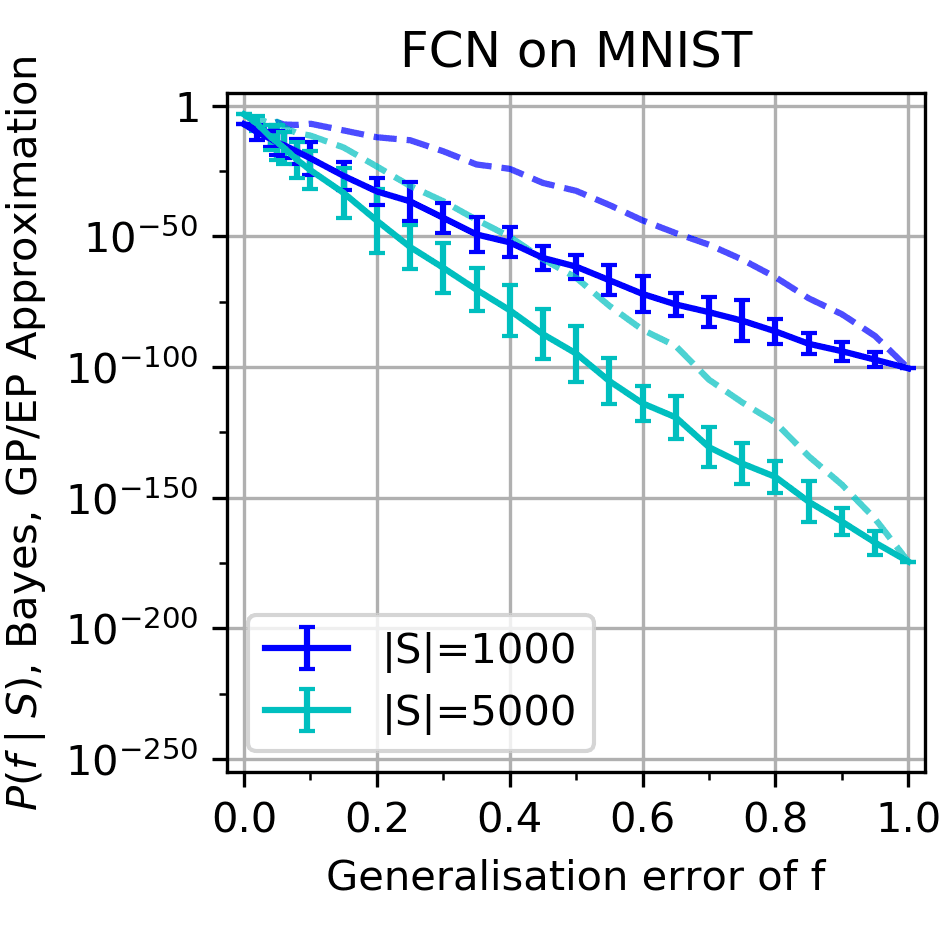}
    \caption{$|S|=1000,5000$}
    \label{fig:mnist:gp_1k5k}
\end{subfigure}

\begin{subfigure}[b]{0.31\textwidth}
    \includegraphics[width=\textwidth]{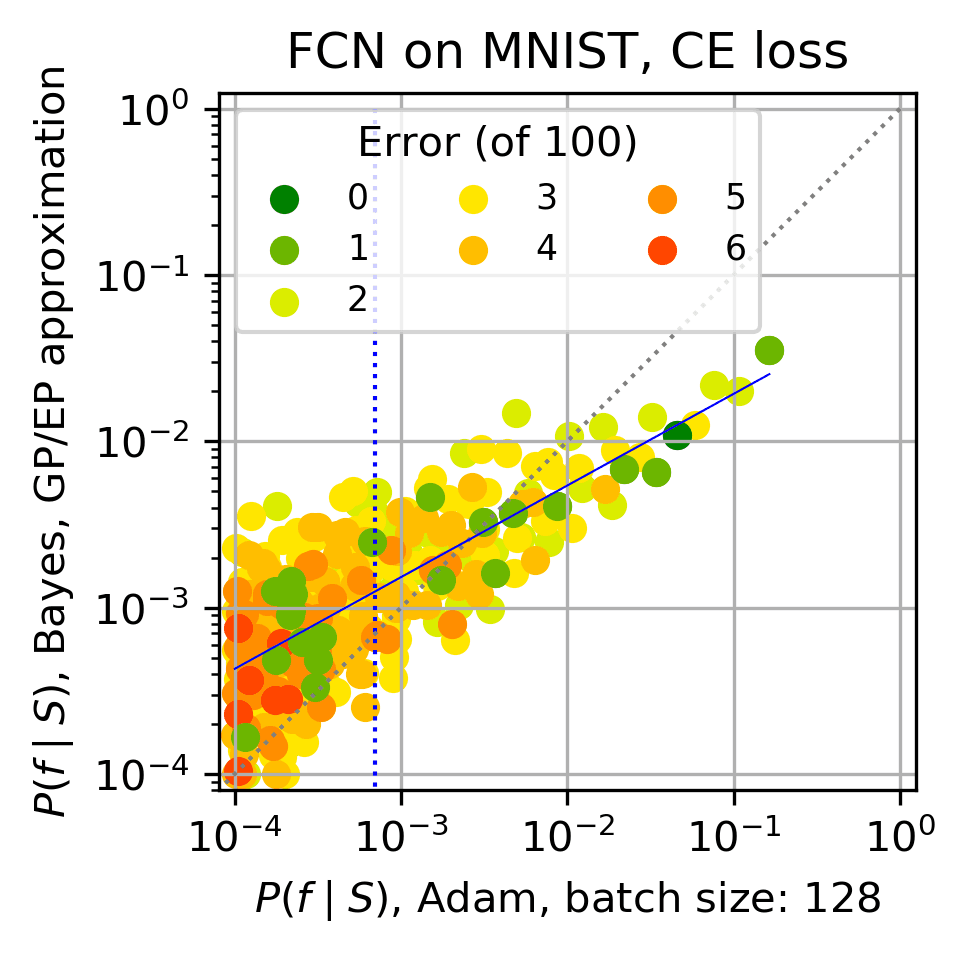}
    \caption{$|S|=10000$}   
    \label{fig:mnist:gp_2_10k}
\end{subfigure}
~~
\begin{subfigure}[b]{0.31\textwidth}
    \includegraphics[width=\textwidth]{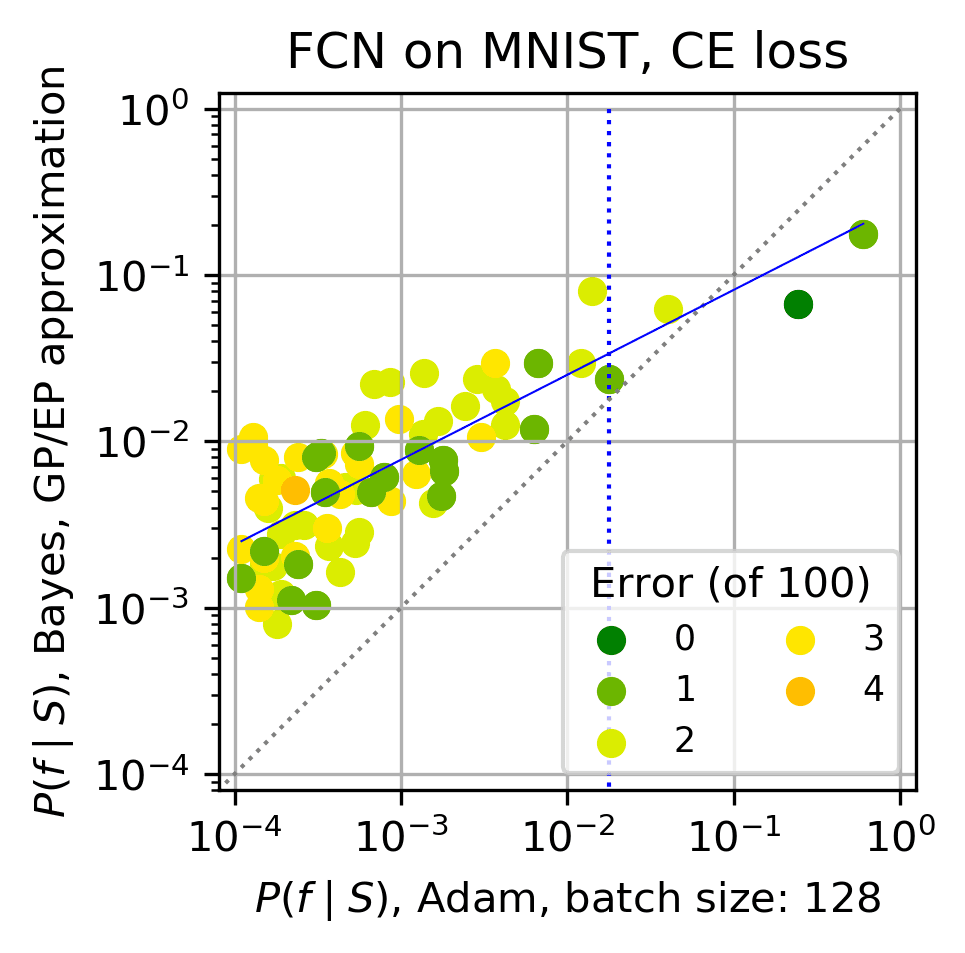}
    \caption{$|S|=20000$}
    \label{fig:mnist:gp_2_20k}
\end{subfigure}
~~
\begin{subfigure}[b]{0.31\textwidth}
    \includegraphics[width=\textwidth]{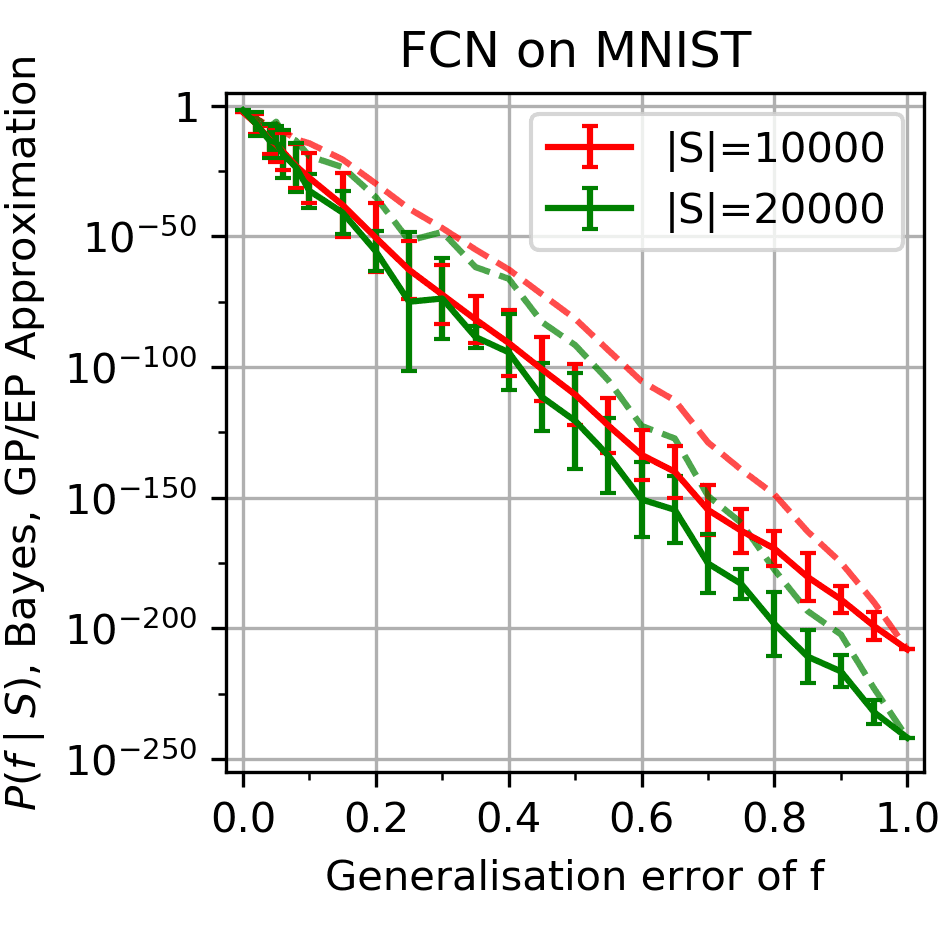}
    \caption{$|S|=10000,20000$}
    \label{fig:mnist:gp_10k20k}
\end{subfigure}

\caption{ {\bf Comparing  $\bf P_{B}(f|S)$ to  $\bf P_{Adam}(f|\bf S)$  for an FCN on MNIST with CE loss for different training set sizes.}  [We use test set size of $|E|=100$;  vertical dotted blue lines denote $90 \%$ probability boundary; solid blue line is a guide to the eye,   dashed grey line is $x=y$.] \\
(a) 1000 training examples. $\eg=$ 6.65\% for Adam. \\
(b) 5000 training examples. $\eg=$ 3.33\% for Adam. \\
(c) $\pb$ v.s.\ $\epsilon_G$ for 1000 and 5000 training examples.\\
(d) 10000 training examples. $\eg=$ 2.20\% for Adam.\\
(e) 20000 training examples. $\eg=$ 0.89\% for Adam.\\
(f) $\pb$ v.s.\ $\epsilon_G$ for 5000 and 10,000 training examples.\\
A trend of increasing bias towards lower error functions with increasing training set size can be clearly observed.  See \cref{fig:train_mse} for related results with MSE loss.}
\label{fig:train_size_main}
\end{figure}

It is also instructive to study the correlation between $\pb$ and $\popt$ for different training set sizes $|S|$.  As can be seen clearly in \cref{fig:train_size_main}, as the training set increases in size, the functions with zero training error are more strongly biased towards low generalisation error, as expected.  \cref{fig:mnist:gp_1k5k,fig:mnist:gp_10k20k}  also illustrate how the stronger bias with increasing $|S|$ means that the entropic factor $\rho(\epsilon_G)$ plays a smaller role. Thus, for larger training set size, but for the same amount of sampling $n$, fewer functions are found, but on average they have higher probability. 

An important question in deep learning is: How does the error reduce with increasing the training set size?  There is intriguing evidence that such ``learning curves'' follow a power law that depends on data complexity, and only weakly on the architecture \citep{hestness2017deep,spigler2019asymptotic,rosenfeld2019constructive,kaplan2020scaling}.  \cref{fig:train_size_main} shows how the spectrum of function probabilities changes with increasing $|S|$.  Investigations based on this more fine-grained picture  may help improve our understanding of learning curves. 

\begin{figure}[H]
\centering

\begin{subfigure}[b]{0.35\textwidth}
    \includegraphics[width=\textwidth]{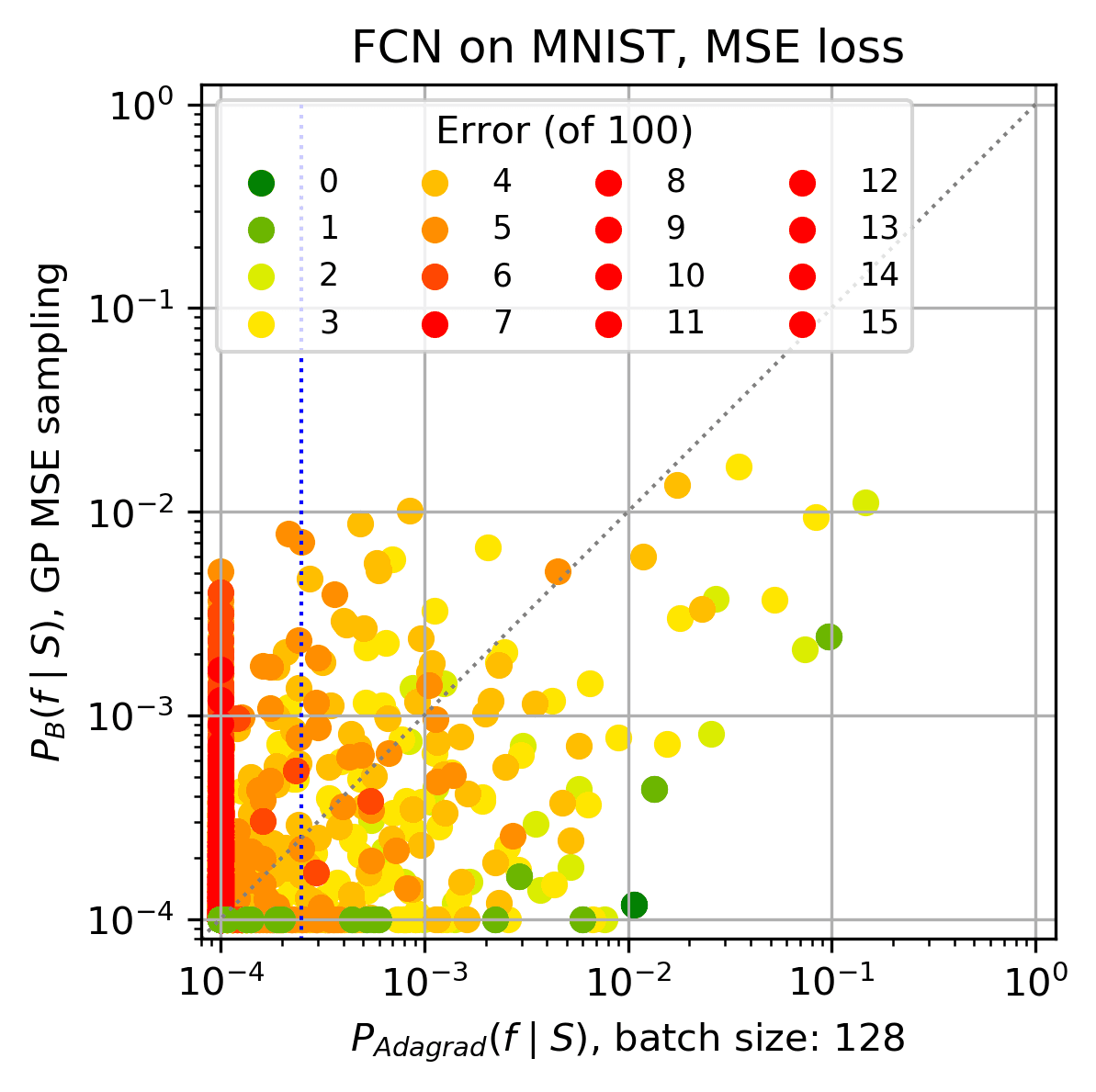}
    \caption{$|S|=1000$. 
    $\eg=2.68\%$}
    \label{fig:mnist:train_1k_mse}
\end{subfigure}
\begin{subfigure}[b]{0.35\textwidth}
    \includegraphics[width=\textwidth]{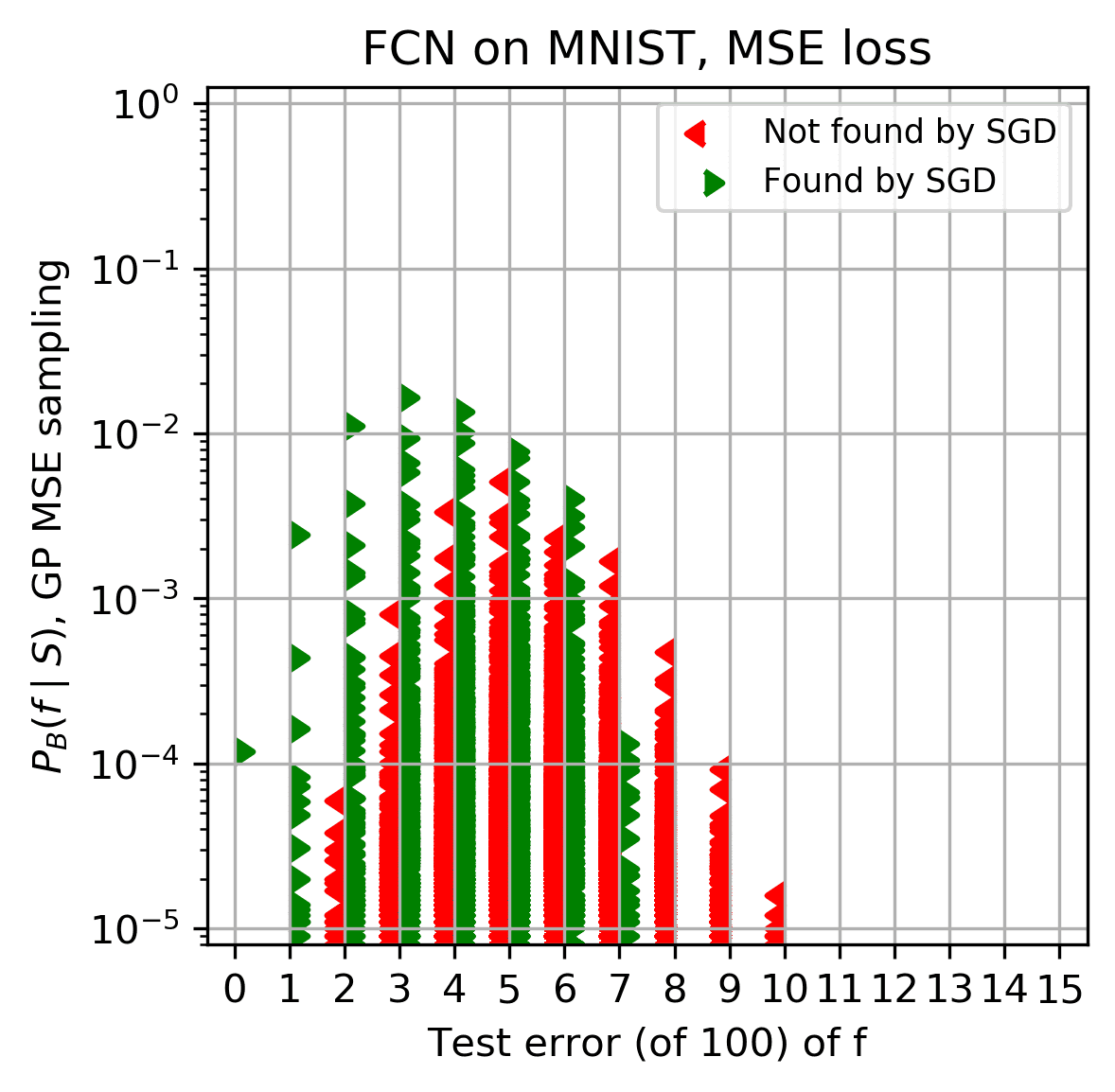}
    \caption{$|S|=1000$. 
    $\eg_{GP}=5.29\%$}
    \label{fig:mnist:train_1k_mse_3}
\end{subfigure}
\begin{subfigure}[b]{0.35\textwidth}
    \includegraphics[width=\textwidth]{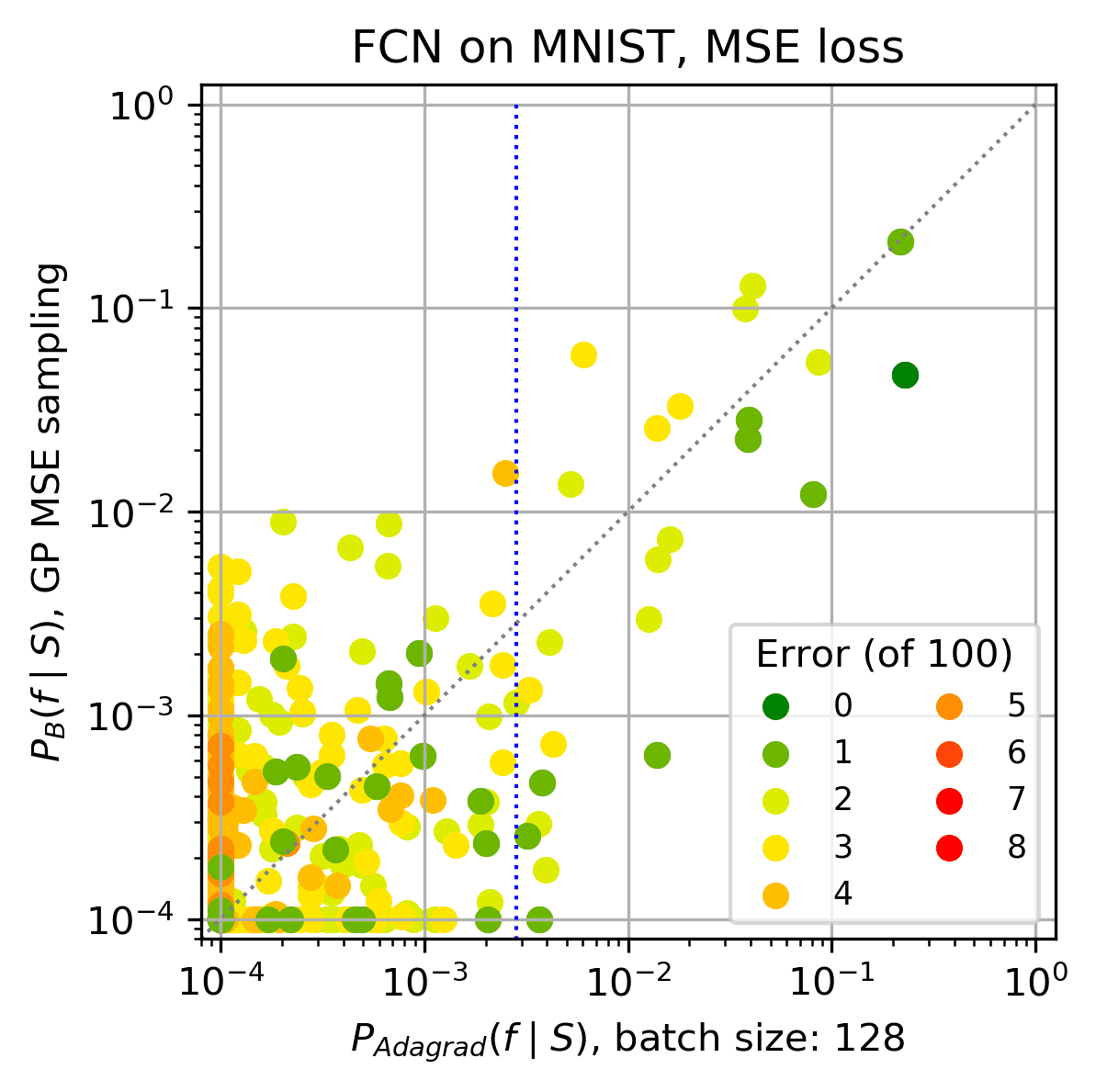}
    \caption{$|S|=5000$. 
    $\eg=1.24\%$}   
    \label{fig:mnist:train_5k_mse}
\end{subfigure}
\begin{subfigure}[b]{0.35\textwidth}
    \includegraphics[width=\textwidth]{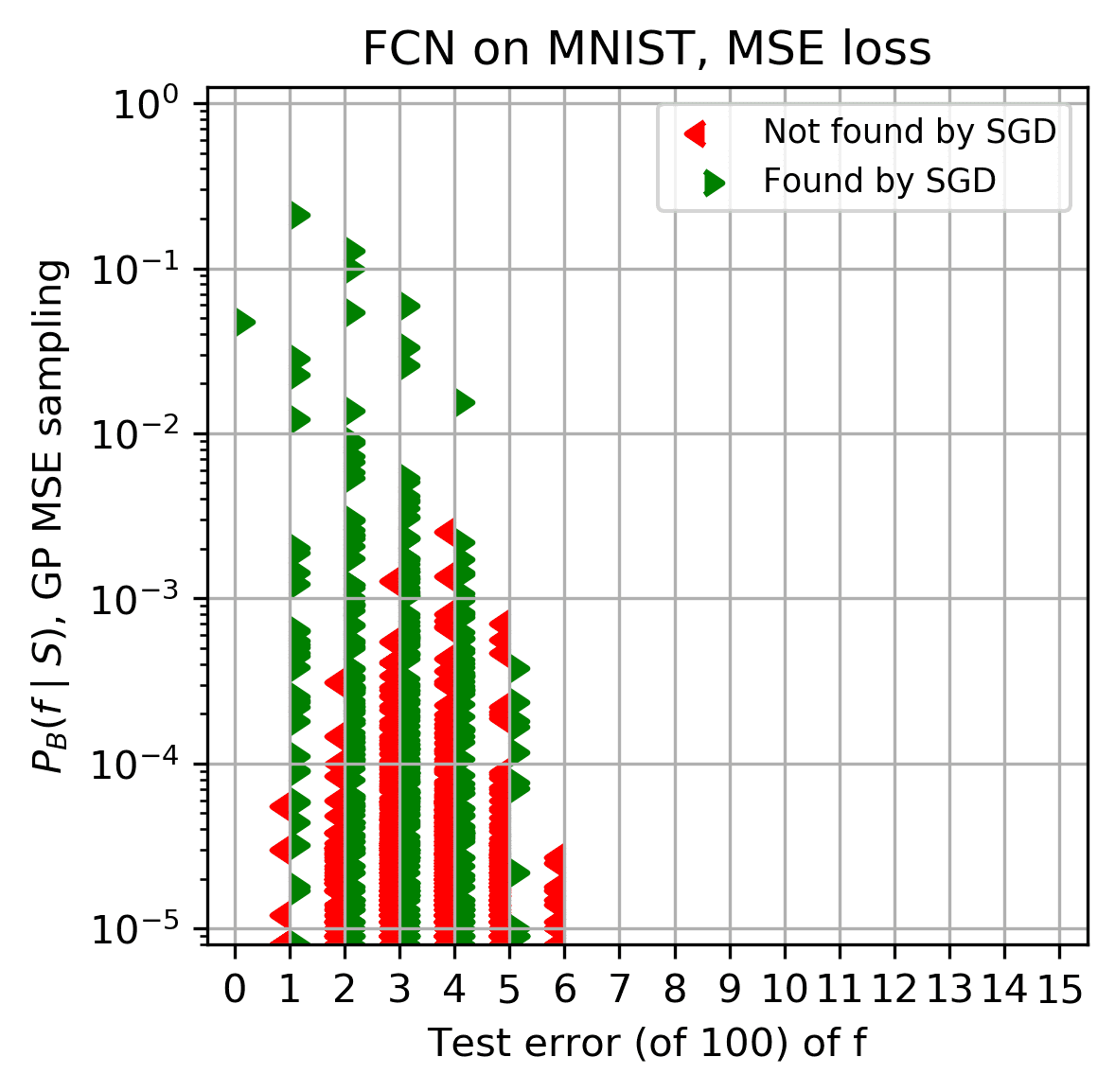}
    \caption{$|S|=5000$. 
    $\eg_{GP}=1.98\%$}
    \label{fig:mnist:train_5k_mse_3}
\end{subfigure}

\begin{subfigure}[b]{0.35\textwidth}
    \includegraphics[width=\textwidth]{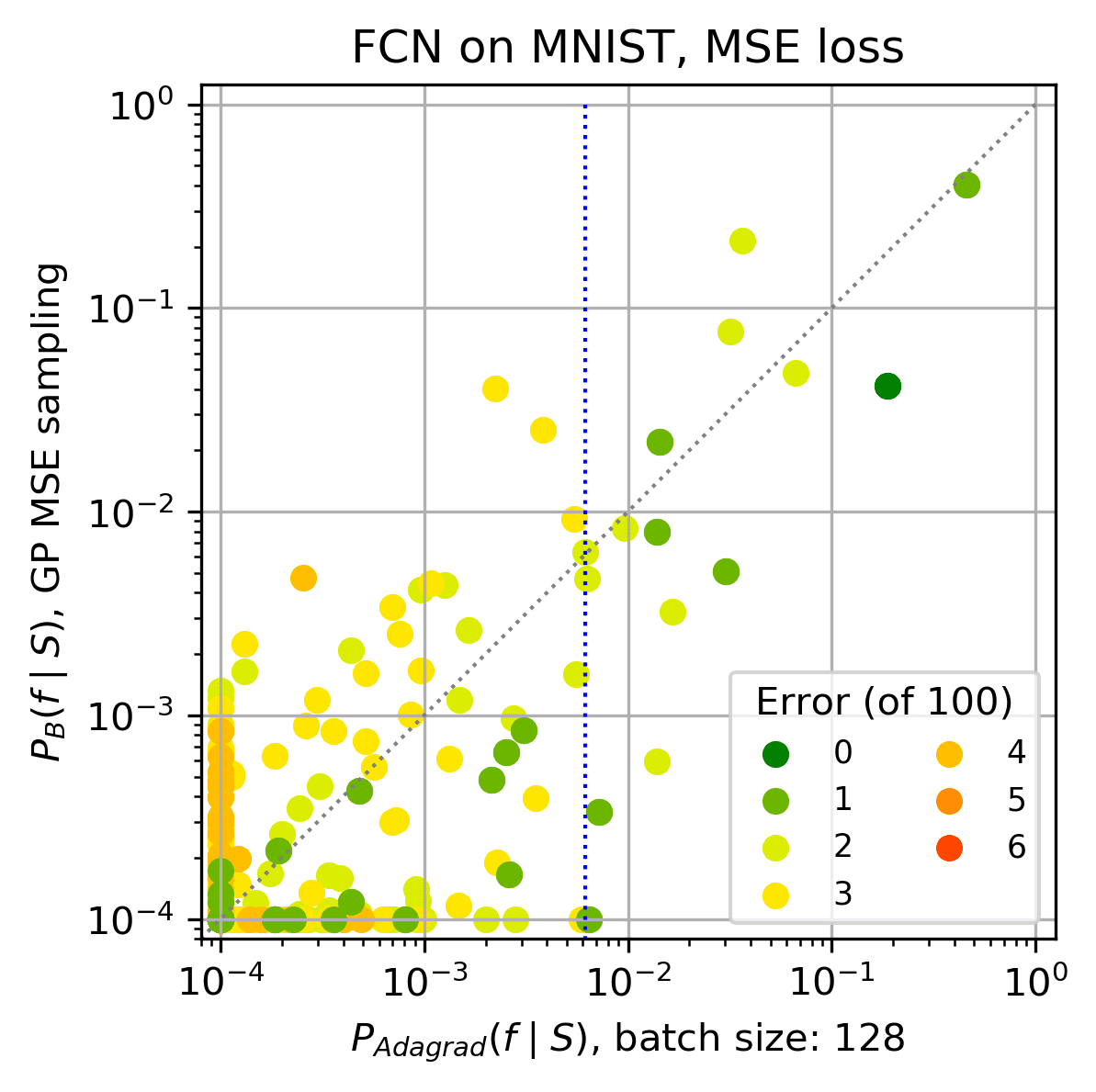}
    \caption{$|S|=10000$. 
    $\eg=1.14\%$}   
    \label{fig:mnist:train_10k_mse}
\end{subfigure}
\begin{subfigure}[b]{0.35\textwidth}
    \includegraphics[width=\textwidth]{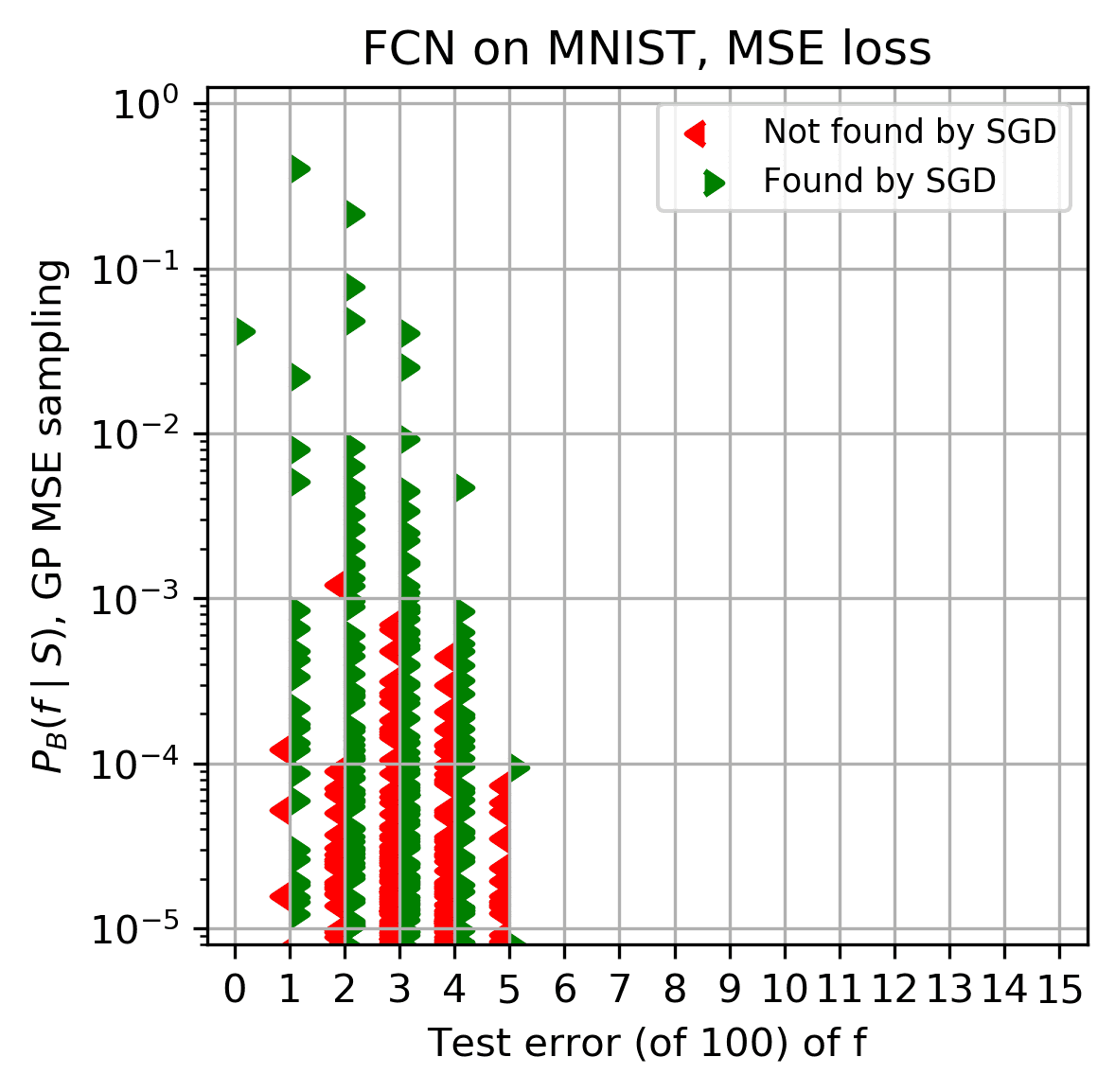}
    \caption{$|S|=10000$. 
    $\eg{GP}=1.61\%$ }
    \label{fig:mnist:train_10k_mse_3}
\end{subfigure}

\caption{{\bf $\mathbf{\pb}$ v.s. $\mathbf{\padagrad}$ for an FCN on MNIST with MSE loss, for different training set sizes. Further results for \cref{fig:train_size_main}.}
 [Test set size $|E|=100$ and batch size=128. Vertical dotted blue lines denote $90 \%$ probability boundary; solid blue lines are fit to guide the eye; dashed grey line is $x=y$.]
(a) and (b) show 1000 training examples, and (c) and (d) show 5000 training examples, and (e) and (f) show 10000. 
As expected, more training examples reduce the generalisation error and increase the correlation between $\padagrad$ and $\pb$.
For all the larger training sets, the $\sum \pb \approx 1$, but for $|S=1000|$ functions found by GP sampling only make up about $40\%$ of the probability of all functions found by Adagrad.
}\label{fig:train_mse}
\end{figure}

\subsection{Changing optimisers}\label{sec:optimisers}

\begin{figure}[H]
\centering

 \begin{subfigure}[b]{0.31\textwidth}
 	\includegraphics[width=\textwidth]{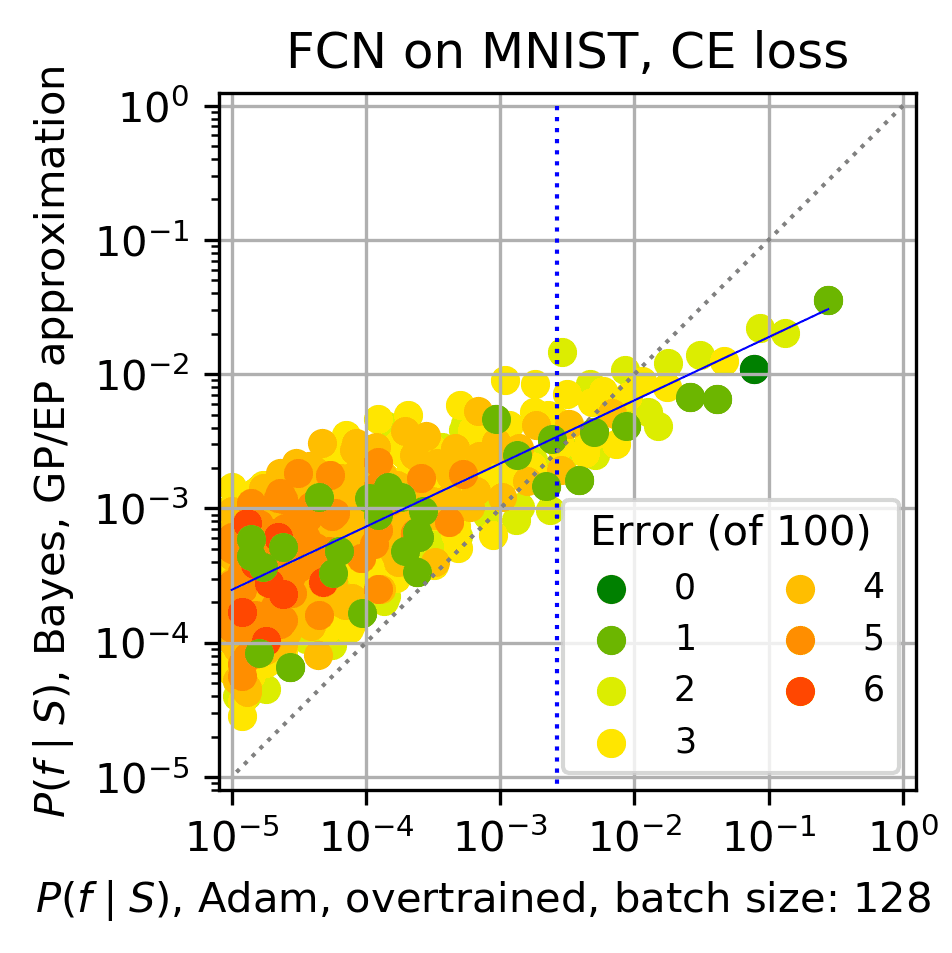}
    \caption{Adam with overtraining}
    \label{fig:optvsopt_adam_o}
\end{subfigure}
~~
\begin{subfigure}[b]{0.31\textwidth}
    \includegraphics[width=\textwidth]{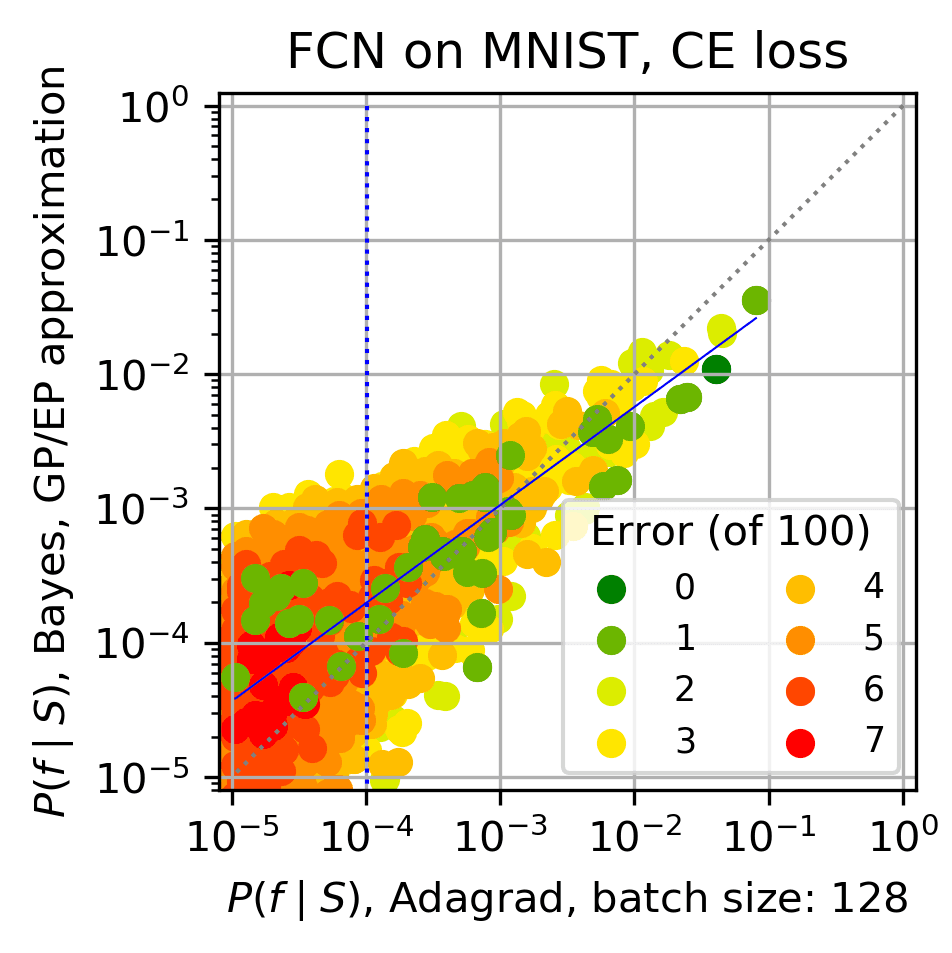}
    \caption{Adagrad, no overtraining}
    \label{fig:optvsopt_adagrad}
\end{subfigure}
~~
\begin{subfigure}[b]{0.31\textwidth}
    \includegraphics[width=\textwidth]{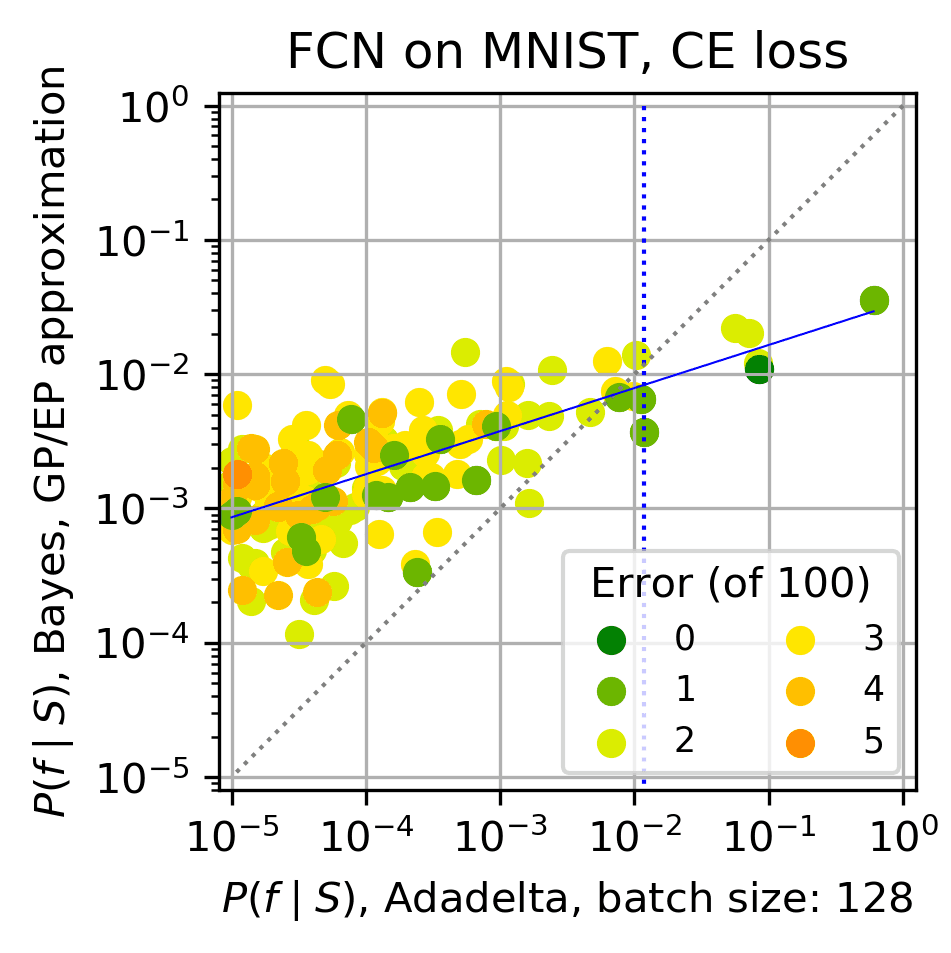}
    \caption{Adadelta, no overtraining}
    \label{fig:optvsopt_adadelta}
\end{subfigure}

\begin{subfigure}[b]{0.31\textwidth}
    \includegraphics[width=\textwidth]{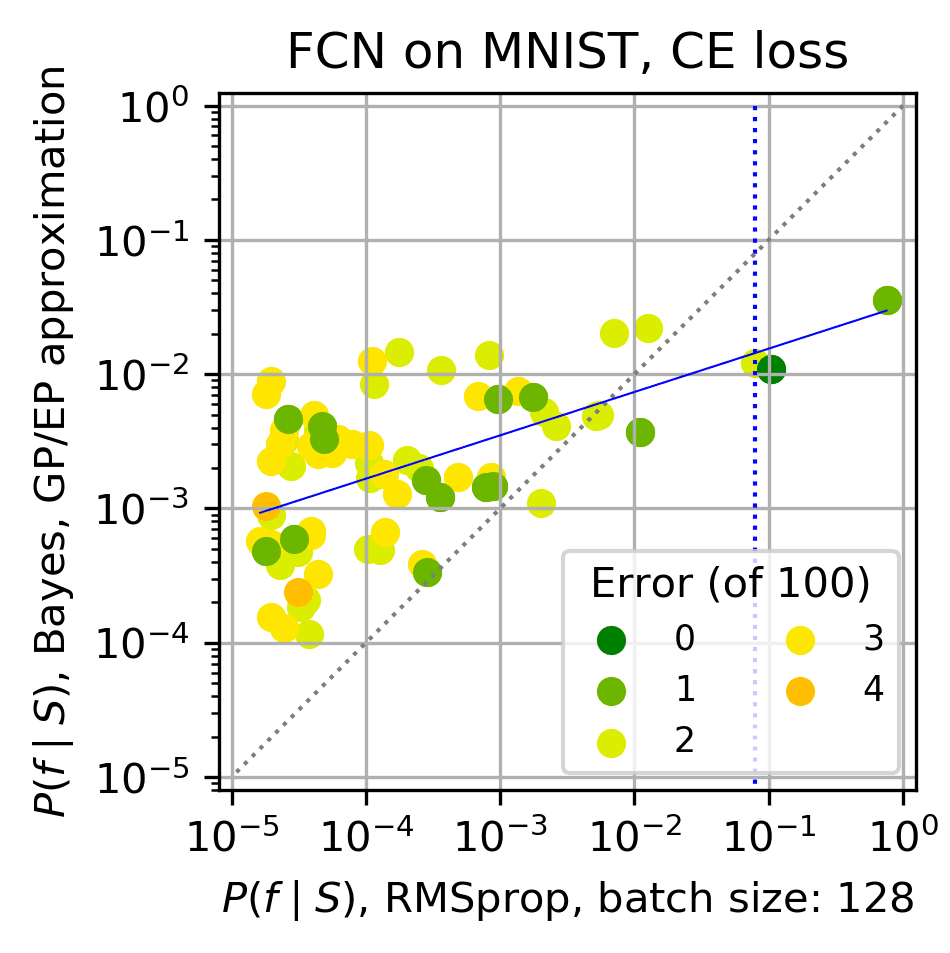}
    \caption{RMSprop, no overtraining}
    \label{fig:optvsopt_RMS}
\end{subfigure}
~~
\begin{subfigure}[b]{0.31\textwidth}
    \includegraphics[width=\textwidth]{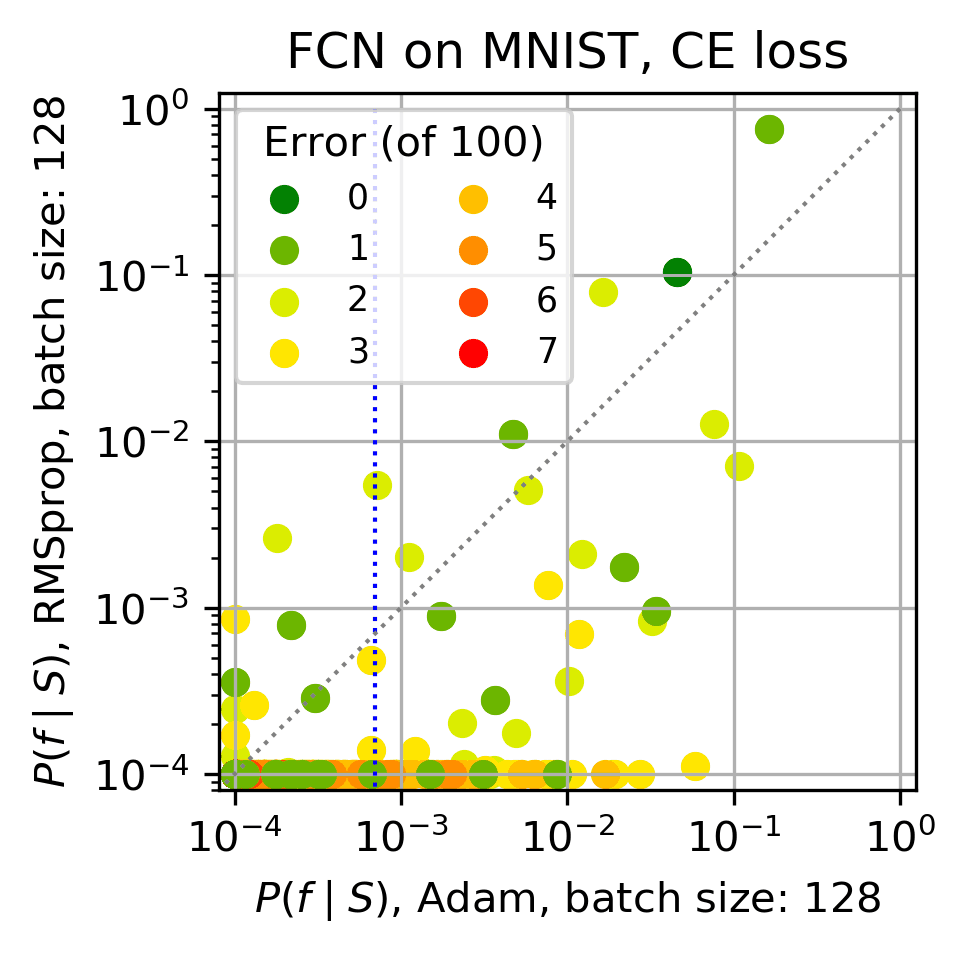}
    \caption{RMSprop vs Adam 128}
    \label{fig:optvsopt_rms-vs-adam128}
\end{subfigure}
~~
\begin{subfigure}[b]{0.31\textwidth}
    \includegraphics[width=\textwidth]{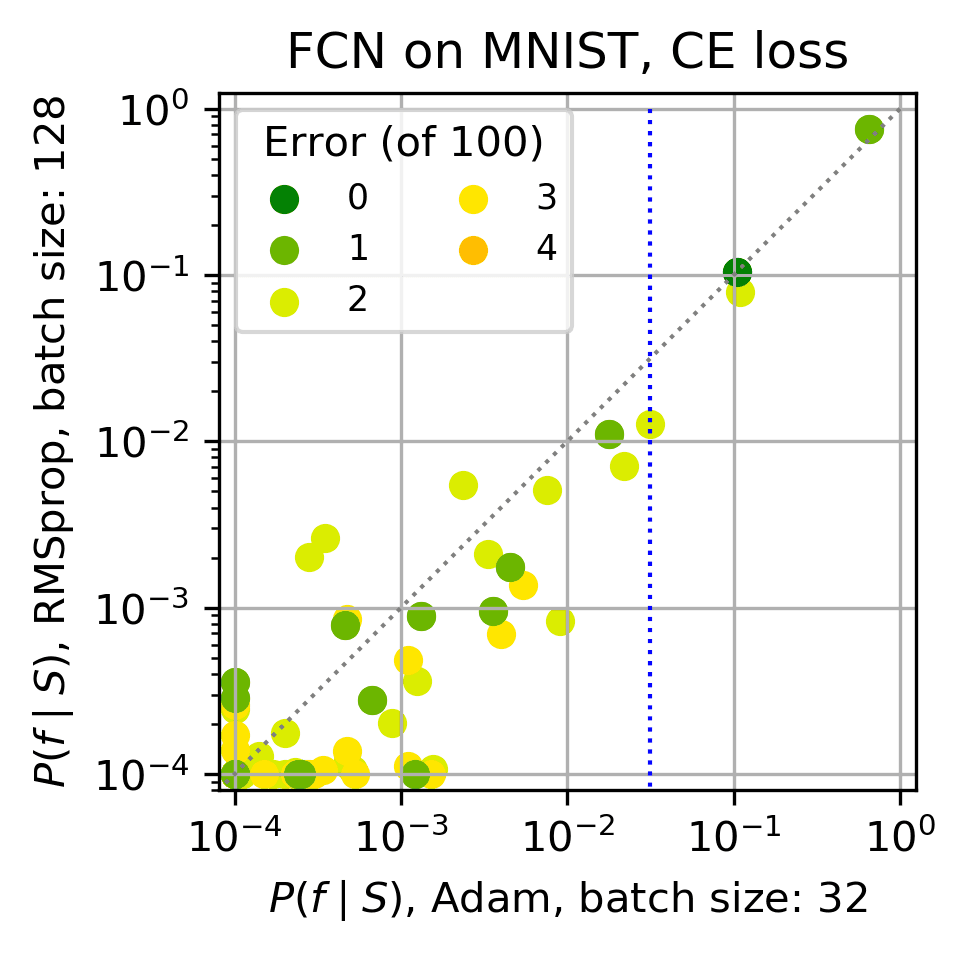}
    \caption{RMSprop vs Adam 32}
    \label{fig:optvsopt_rms-vs-adam32}
\end{subfigure}

\caption{{\bf Comparing  $\bf \pb$ to  $\bf \popt$ for an FCN on MNIST with CE loss for different optimisers.} [We use training/test set size 10,000$/$100 and batch size=128. Vertical dotted blue lines denote $90 \%$ probability boundary; dashed grey line is $x=y$.] \\
(a) Adam with overtraining for 64 epochs. $\eg=$ 1.73\%. \\ 
(b) Adagrad, $\eg=$ 2.63\%.\\
(c) Adadelta,  $\eg=$ 1.23\%. \\
(d) RMSprop,  $\eg=$  1.02\%. \\
(e) $\prmsprop$  v.s.\ $\padam$ both with  batch size 128. \\
(f) $\prmsprop$ with batch size 128  v.s.\ $\padam$ with batch size 32. \\
For Adam without overtraining, see \cref{fig:batch:128}. For Adagrad with overtraining see \cref{fig:optvsopt_adagrad_o_main}. For others with overtraining see \cref{fig:optvsopt_appendix}.   Overtraining has a milder effect than changing the optimiser does. 
(e) and (f) show that  there is a surprisingly close correspondence between RMSprop with batch size 128 and Adam with batch size 32.  By contrast, for the same batch size of 128,  Adam and Adagrad are similar to one another, as are Adadelta and RMSprop (see \cref{fig:4optcorr} for a direct comparison).  The latter two have better generalisation performance, which is reflected in higher probabilities for the lowest error functions.   See \cref{fig:optvsopt_appendix,fig:4optcorr} for further results. \\
In all these figures,  while 
$\popt$ is to first order determined by $\pb$, there are noticeable second order differences.
 }\label{fig:optvsopt}
\end{figure}

Much research effort has gone into adaptations of SGD.  One goal is to achieve more efficient optimisation, but another is to achieve better generalisation.  \cref{fig:optvsopt} illustrates the effect of changing the optimiser on the correlation between $\pb$ and $\popt$. (See also \cref{fig:batch:128,fig:optvsopt_adagrad_o_main,fig:optvsopt_appendix} to complete the set of optimisers with and without overtraining).   To first order this figure shows that $\pb$ and $\popt$ are remarkably closely correlated for all these optimisers, even taking into account that the EP approximation introduces errors, and likely leads to a slightly too small slope in $\pb$ v.s.\ $\popt$.  

What is perhaps more interesting here are second-order effects, since $\pb$ is identical in each plot. For example, RMSprop has the best generalisation performance, which is reflected in a stronger inductive bias towards a few key low error functions.  
Note also the similarity of batch size 128 RMSprop to Adam with smaller batch size of 32. 
We emphasise that this performance here doesn't mean that RMSprop is in general superior to the other SGD variants for FCNs on MNIST. To investigate that question, we would need to study other test and training sets, and need to do further hyperparameter tuning~\citep{choi2019empirical}.  
 
We can also compare the effect of overtraining.    In each case shown in \cref{fig:batch:128,fig:optvsopt_adagrad_o_main,fig:optvsopt_appendix},  overtraining brings a modest improvement in generalisation error on this test set (Adam from $\epsilon_G=2.2\%$ to $\epsilon_G=1.74\%$ and Adagrad from $\epsilon_G=2.63\%$  to $\epsilon_G=2.19\%$), for example.   From the graphs one can see a slight increase in the optimiser probability of the lowest error function with overtraining, but also a clear reduction in the scatter of the data for the whole range of probabilities, suggesting that  for CE loss, on average, overtraining brings $\popt$ closer to the Bayesian prediction.  This behaviour can possibly be rationalised in that overtraining allows the optimiser to sample functions with probabilities closer to the steady-state average (see also \cref{sec:heuristic_arguments}). 

Finally, in \cref{fig:optvsopt_rms-vs-adam32,fig:optvsopt_rms-vs-adam128,fig:4optcorr} we directly compare the $\popt$ to one another, in other words, without using $\pb$. A number of clear trends are visible, for example, with batch size 128, Adam and Adagrad are very similar to one another, as are RMSprop and Adadelta.  However, as can be seen in \cref{fig:optvsopt_rms-vs-adam32}, Adam with a smaller batch size of 32 is very similar to RMSprop with batch size of 128.  What these correlation plots show is that the behaviour of the different optimisers can depend on batch size in subtle ways that may not necessarily be picked up by the generalisation error. 
Further work (and significant computational resources) would be needed to completely compare these methods.

These examples show that studying the spectrum of function probabilities provides more fine-grained data than simply comparing generalisation error does.  Future studies on problems such as optimiser choice or hyperparameter tuning could exploit this fuller set of information to increase understanding and to improve DNN performance.

\section{Further results comparing $\popt$ to $\pb$.}\label{appendix:further_results}


\begin{figure}[H]
\centering

\begin{subfigure}[b]{0.4\textwidth}
    \includegraphics[width=\textwidth]{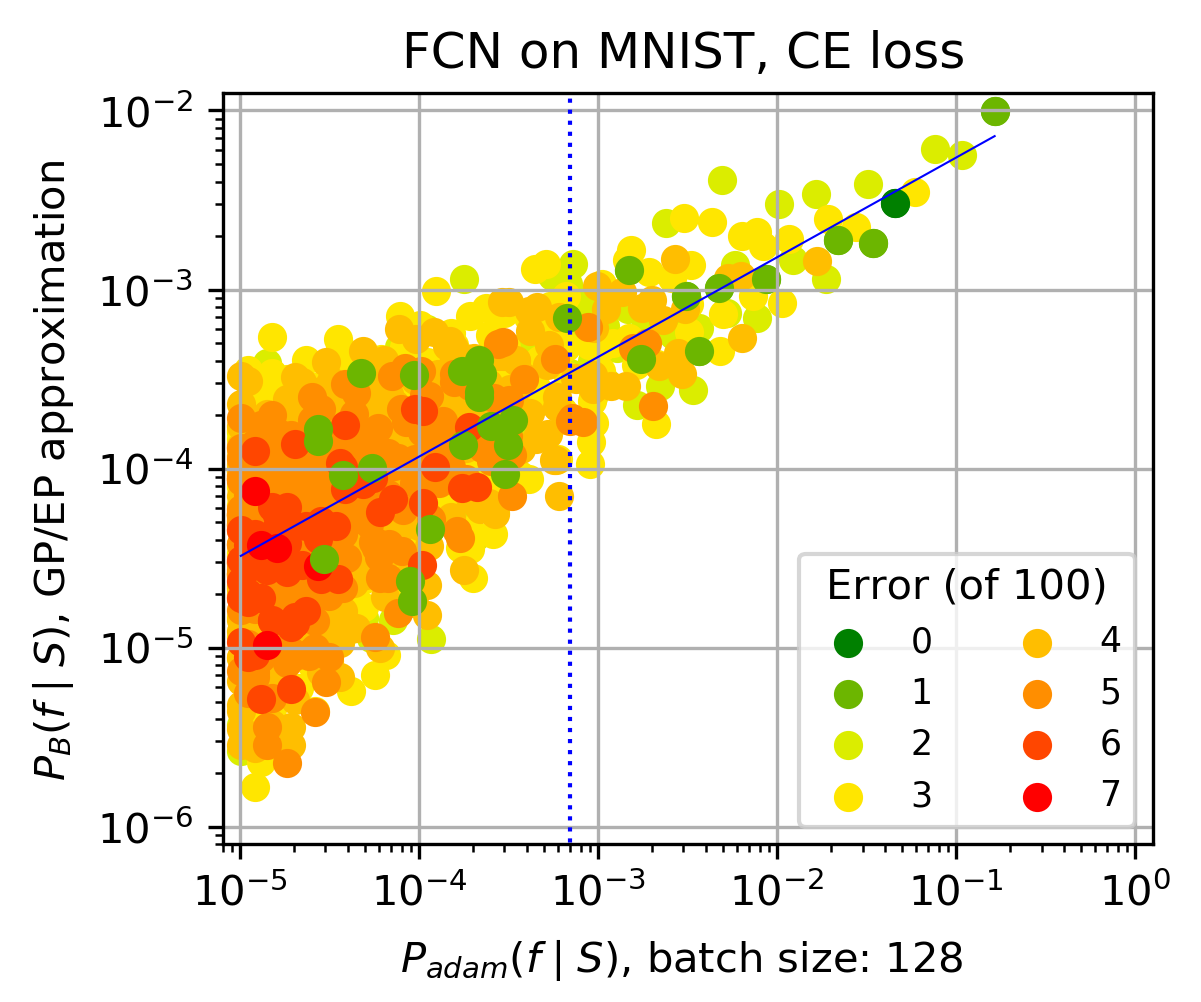}
    \caption{Test set 1: $\eg=$2.20\%.}
    \label{fig:test_norm}
\end{subfigure}
~~
\begin{subfigure}[b]{0.4\textwidth}
	\includegraphics[width=\textwidth]{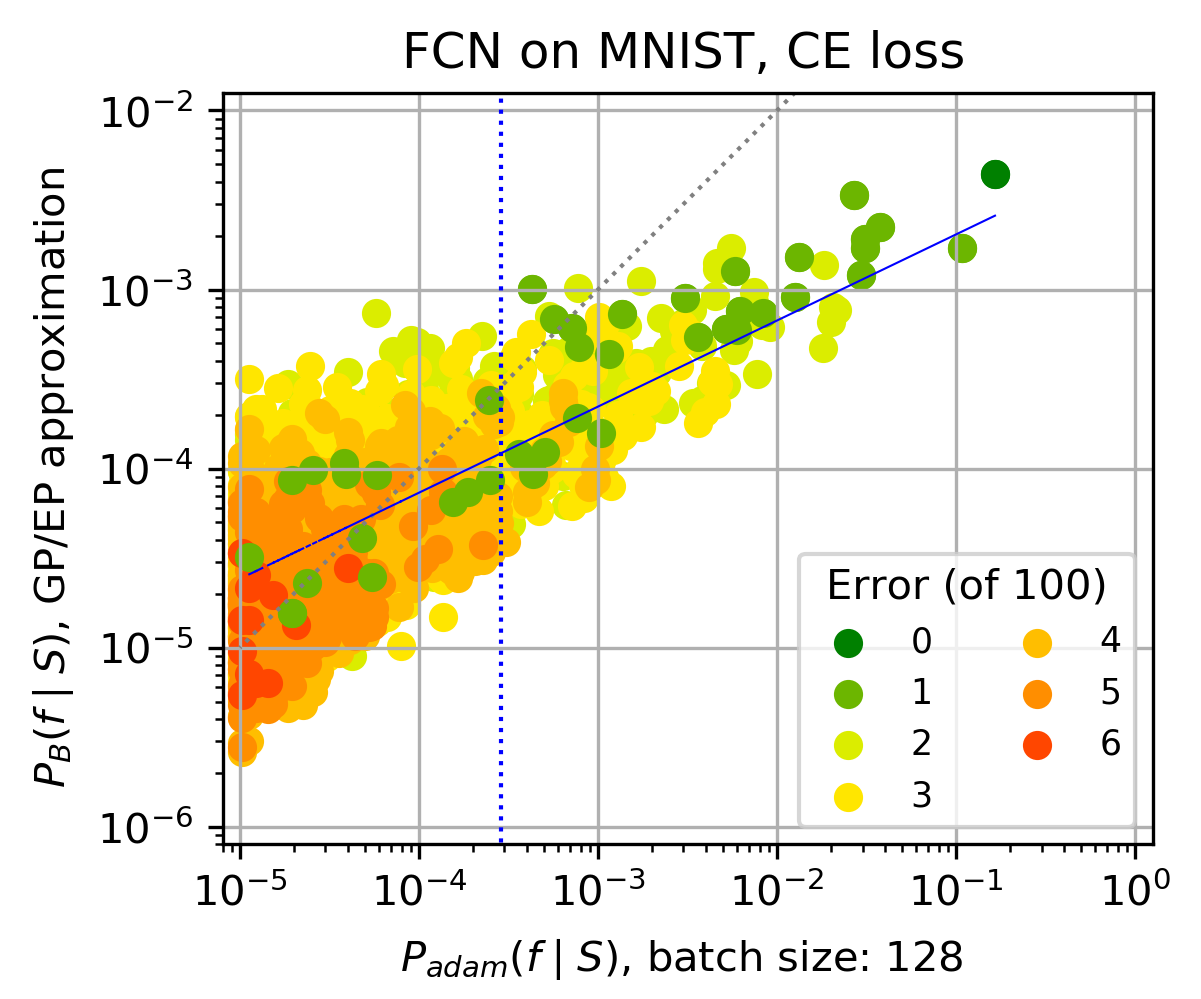}
    \caption{Test set 2: $\eg=$1.62\%}
    \label{fig:test_100-200}
\end{subfigure}
\caption{{\bf Comparing $\pb$ to $\padam$  for an FCN on MNIST with CE loss for two different test sets}
[We use training/test set size 10,000$/$100 and batch size=128. Vertical dotted blue lines denote $90 \%$ probability boundary; solid blue lines are fit to guide the eye; dashed grey line is $x=y$.]  (a) is the test set used throughout the paper and (b) is another test set (disjoint from both the first test set and the training set), chosen at random from MNIST.  To first order the two look very similar, but to second order small differences can be seen, not just in the function probabilities, but also in the slope of $\pb$ v.s.\ $\padam$, which may be affected by the EP approximation used here. No normalisation was applied to this figure so it depicts the raw EP approximation (see~\cref{app:GP_exp_0-1}).
}\label{fig:test_set_dependence}
\end{figure}



\begin{figure}[H]
\centering

\begin{subfigure}[b]{0.4\textwidth}
    \includegraphics[width=\textwidth]{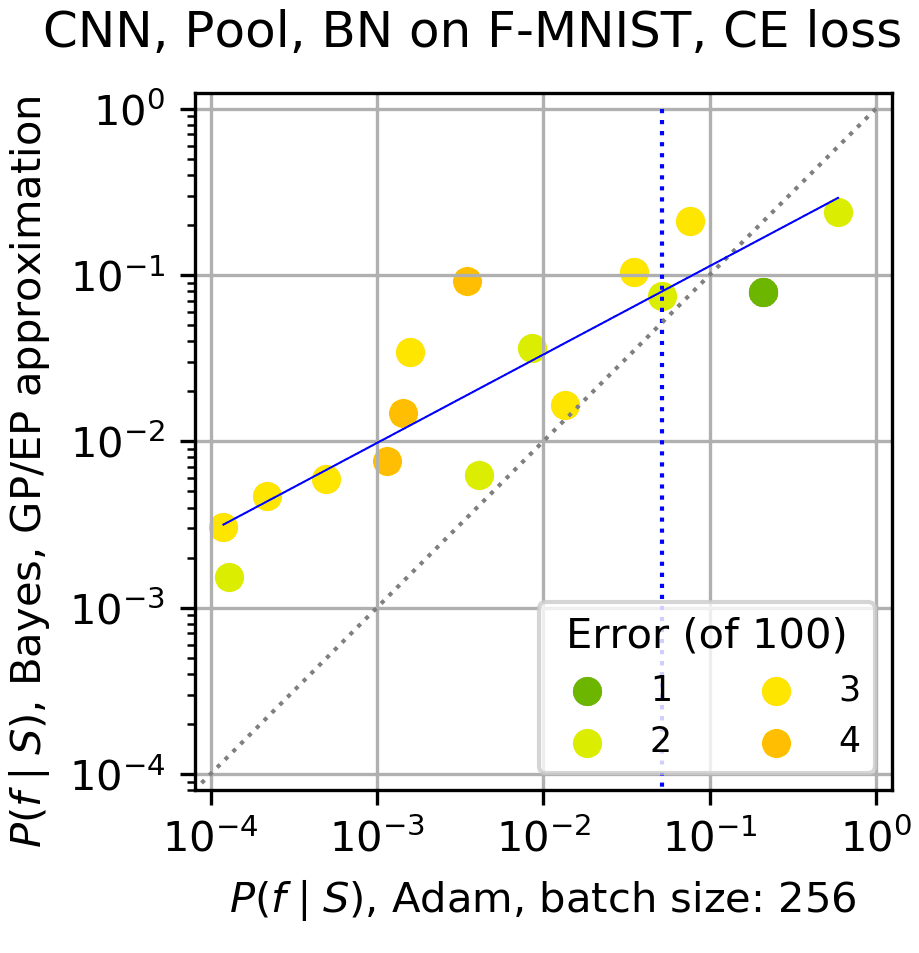}
    \caption{$\pb$ v.s.\ $\padam$}
    \label{fig:fashionMNIST_CP_3}
\end{subfigure}
\begin{subfigure}[b]{0.4\textwidth}
    \includegraphics[width=\textwidth]{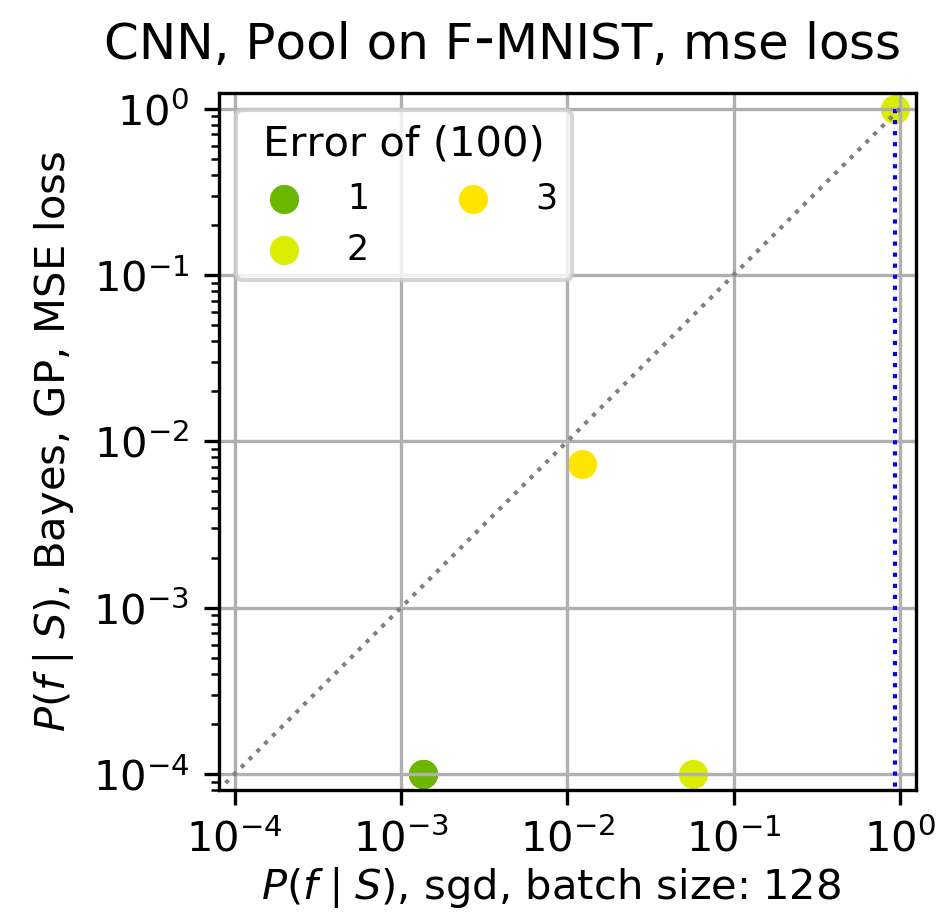}
    \caption{$\pb$ v.s.\ $\psgd$}
    \label{fig:fashionMNIST_MSE}
\end{subfigure}

\begin{subfigure}[b]{0.4\textwidth}
    \includegraphics[width=\textwidth]{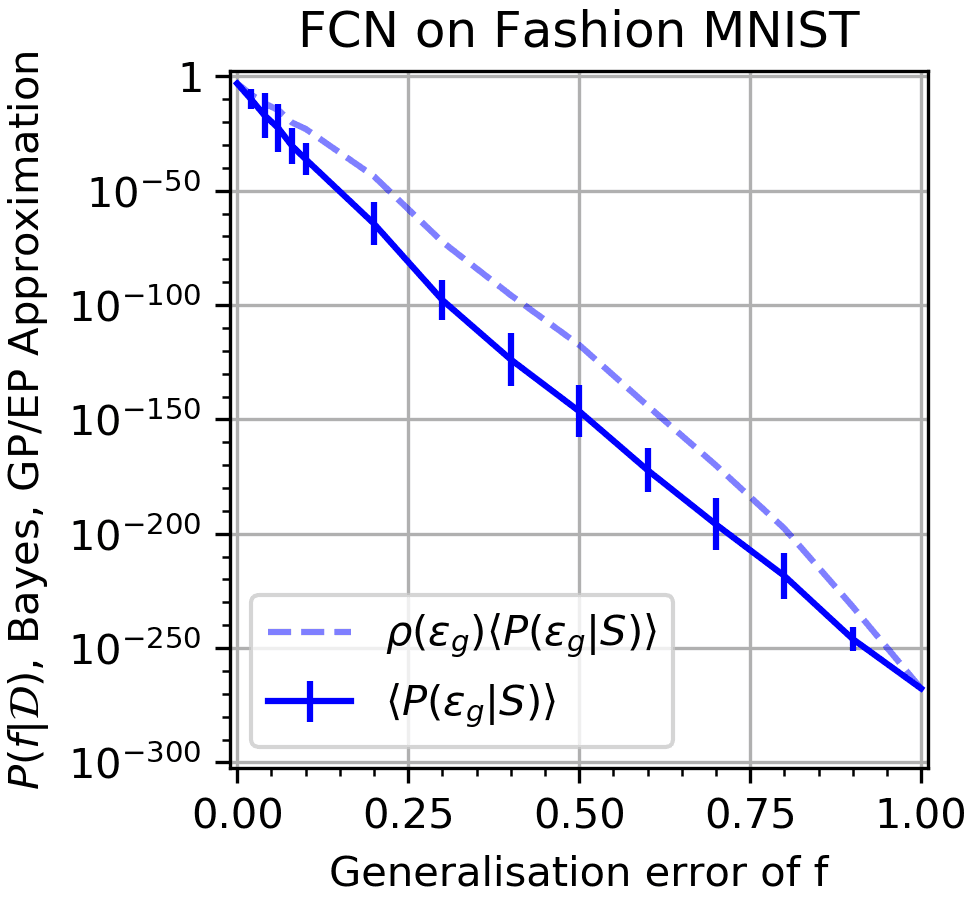}
    \caption{15 samples per $\eg$}
    \label{fig:fashionMNIST_FCN_3}
\end{subfigure}
\begin{subfigure}[b]{0.4\textwidth}
	\includegraphics[width=\textwidth]{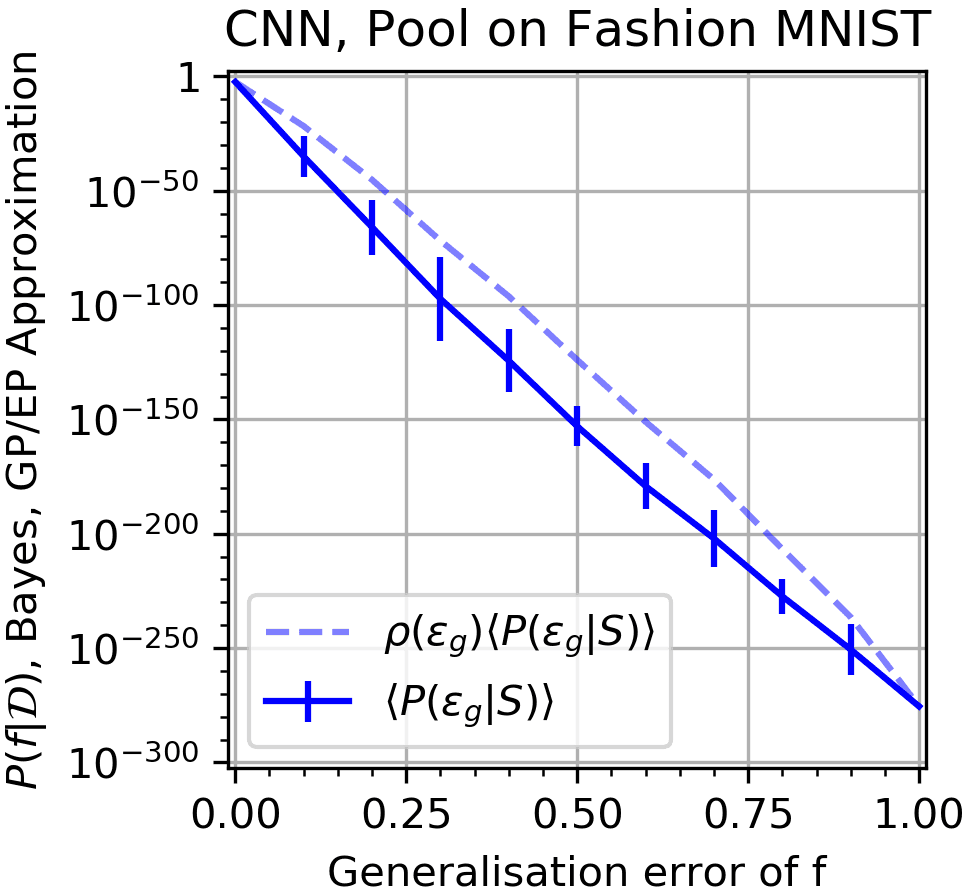}
    \caption{15 samples per $\eg$}
    \label{fig:fashionMNIST_CPBN}
\end{subfigure}

\caption{{\bf Comparing the Bayesian prediction $\bf P_{B}(f|S)$ to  $\bf P_{Adam}(f|\bf S)$ for an FCN and CNNs on Fashion-MNIST.  Further results for \cref{fig:fashionMNIST}.} [We use  train/test set size of 10,000$/$100; vertical dotted blue lines denote $90 \%$ probability boundary; solid blue lines are fit to guide the eye; dashed grey line is $x=y$.]
(a) CNN with max-pooling and batch normalisation on Fashion-MNIST; $\eg=2.11\%$ for Adam with CE loss. Note that the GP kernel used is the same as in \cref{fig:fashionMNIST_CP}, so with pooling but without batch normalisation. The effect of batch normalisation is relatively small on this system. 
(b) CNN with max-pooling on Fashion-MNIST and MSE loss. $\eg=2.01$ for SGD and $\eg=2.00$ for the GP MSE sampling. 2000 samples for SGD, $10^5$ samples from the GP. The same function is found over 99\% of the time by both the GP and SGD.
(c) and (d) show an average of $\pb$ versus error for an FCN and CNN with max pooling respectively.  }\label{fig:fashionMNIST_appendix}
\end{figure}
\vspace{-12pt}
\begin{figure}[H]
\centering

\begin{subfigure}[b]{0.31\textwidth}
    \includegraphics[width=\textwidth]{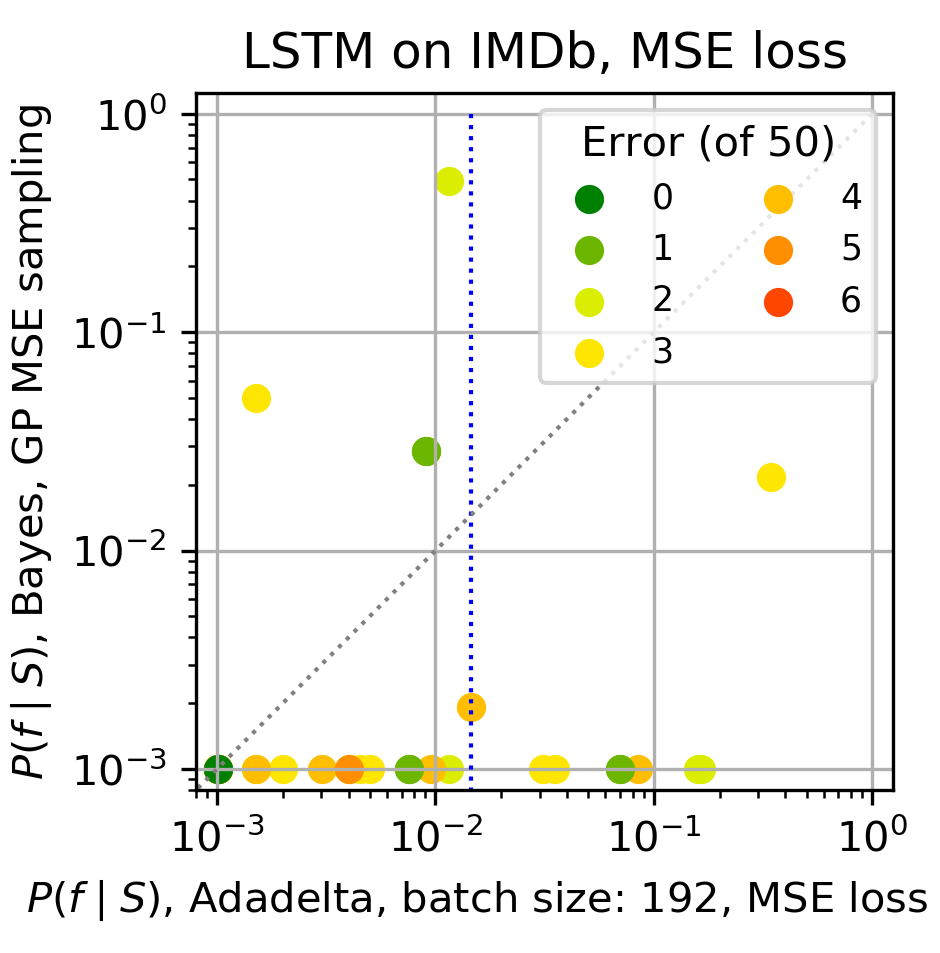}
    \caption{$\pb$ v.s.\ $\padadelta$}
    \label{fig:lstm_mse_1}
\end{subfigure}
~~
\begin{subfigure}[b]{0.31\textwidth}
	\includegraphics[width=\textwidth]{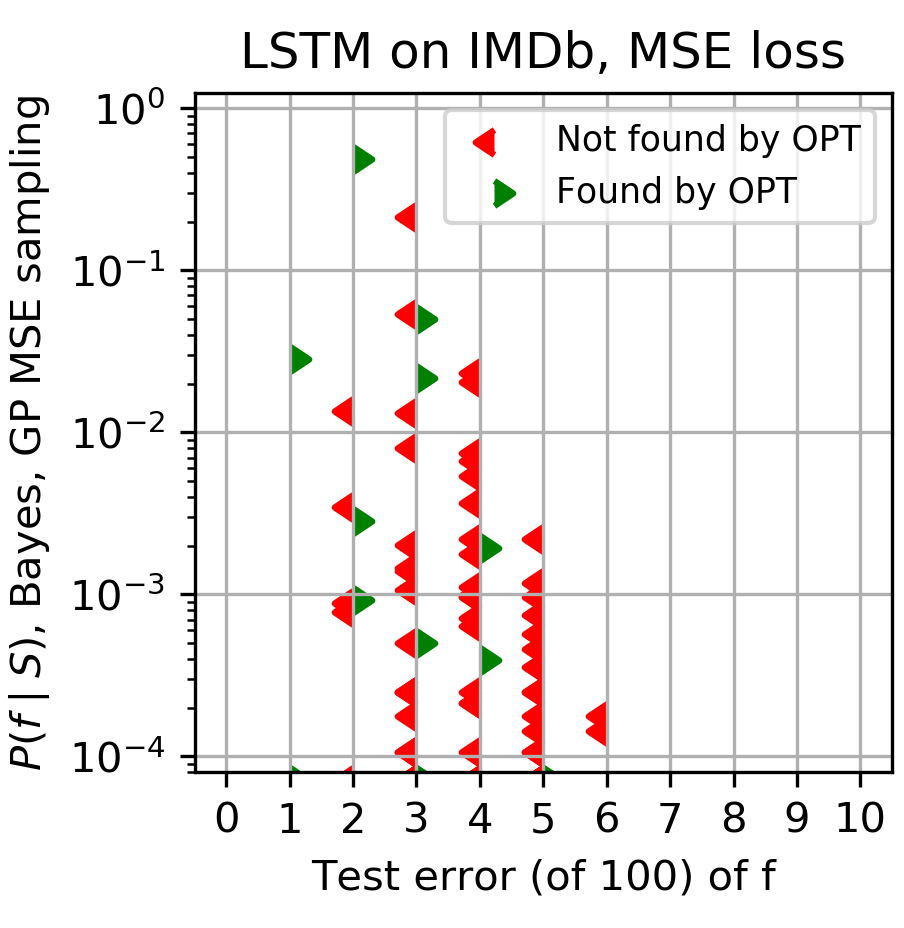}
    \caption{$f$ found by GP in (a)}
    \label{fig:lstm_mse_2}
\end{subfigure}
~~
\begin{subfigure}[b]{0.31\textwidth}
	\includegraphics[width=\textwidth]{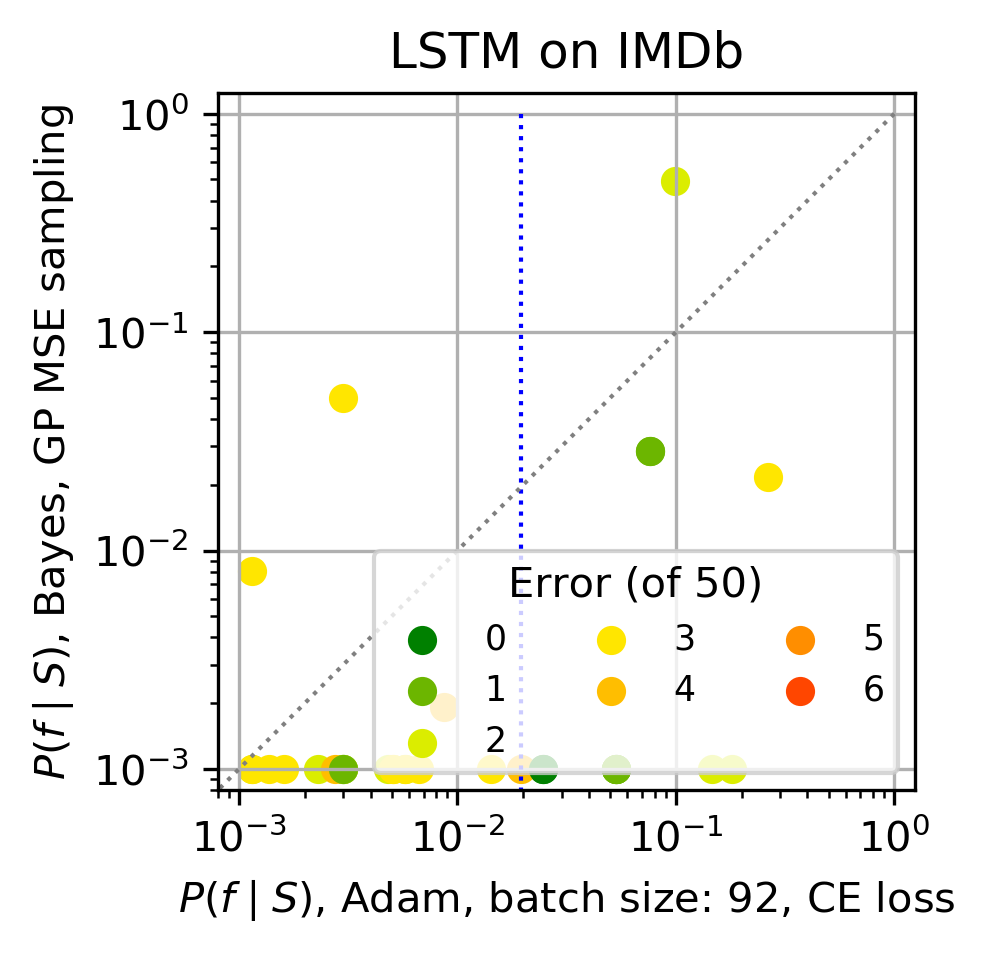}
    \caption{$\pb$ v.s.\ $\padam$}
    \label{fig:lstm_mse_ce}
\end{subfigure}

\caption{{\bf Comparing $\pb$ to $\popt$  for a LSTM on the IMDb movie review dataset. Further results for \cref{fig:lstm_main}.}
 [We use training/test set size 45,000$/$50 and batch size=192. Vertical dotted blue lines denote $90 \%$ probability boundary; solid blue lines are fit to guide the eye; dashed grey line is $x=y$.] \\
 (a) $\pb$ v.s. $\padadelta$ for MSE loss. 
 $n=2200$ for the optimiser and $n=2.6*10^4$ for the GP. 
 $\eg=5.22\%$  and $\eg_{GP}=5.04\%$  \\
 (b) Probability of all functions found by NNGP compared to those also found by the optimiser.  Green points are the set jointly found functions $F$. Red denotes functions found only by the GP,  $\sum_{f\in F}\pb=54.1\%$ and $\sum_{f\in F}\popt=77.0\%$.  \\
 (c) Compares the $\padam$ with CE loss from \cref{fig:lstm_exp1_main},  to the sampled $\pb$ using MSE loss for $n=10^4$ samples. \\
 While these results are for very limited sample numbers, they provide  evidence that for the LSTM,  $\pb$ has values on the same order of magnitude as $\psgd$.  The low values we find for the raw EP approximation estimates of $\pb$ are likely to be due to errors in the EP absolute values.  The fact that we still see correlations for the CE-trained LSTM with the renormalised EP approximations for $\pb$ suggests that the EP still does reasonably on relative errors.
 CE has the advantage that it is much faster to use than MSE.
}\label{fig:lstm}
\end{figure}


\begin{figure}[H]
\centering

\begin{subfigure}[b]{0.4\textwidth}
    \includegraphics[width=\textwidth]{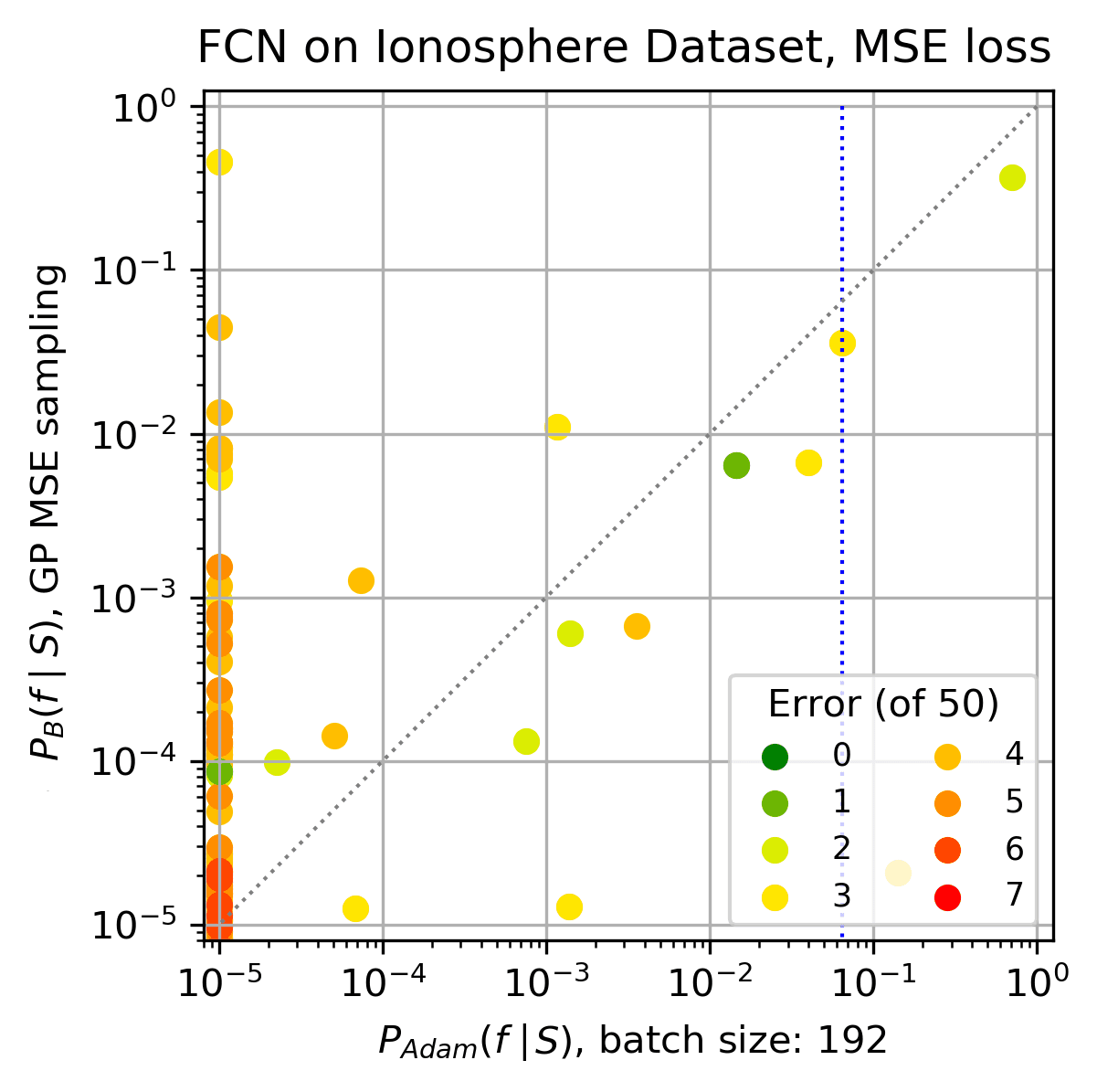}
    \caption{$\pb$ v.s.\ $\padam$}
    \label{fig:nonimage_ion_mse}
\end{subfigure}
~~
\begin{subfigure}[b]{0.4\textwidth}
	\includegraphics[width=\textwidth]{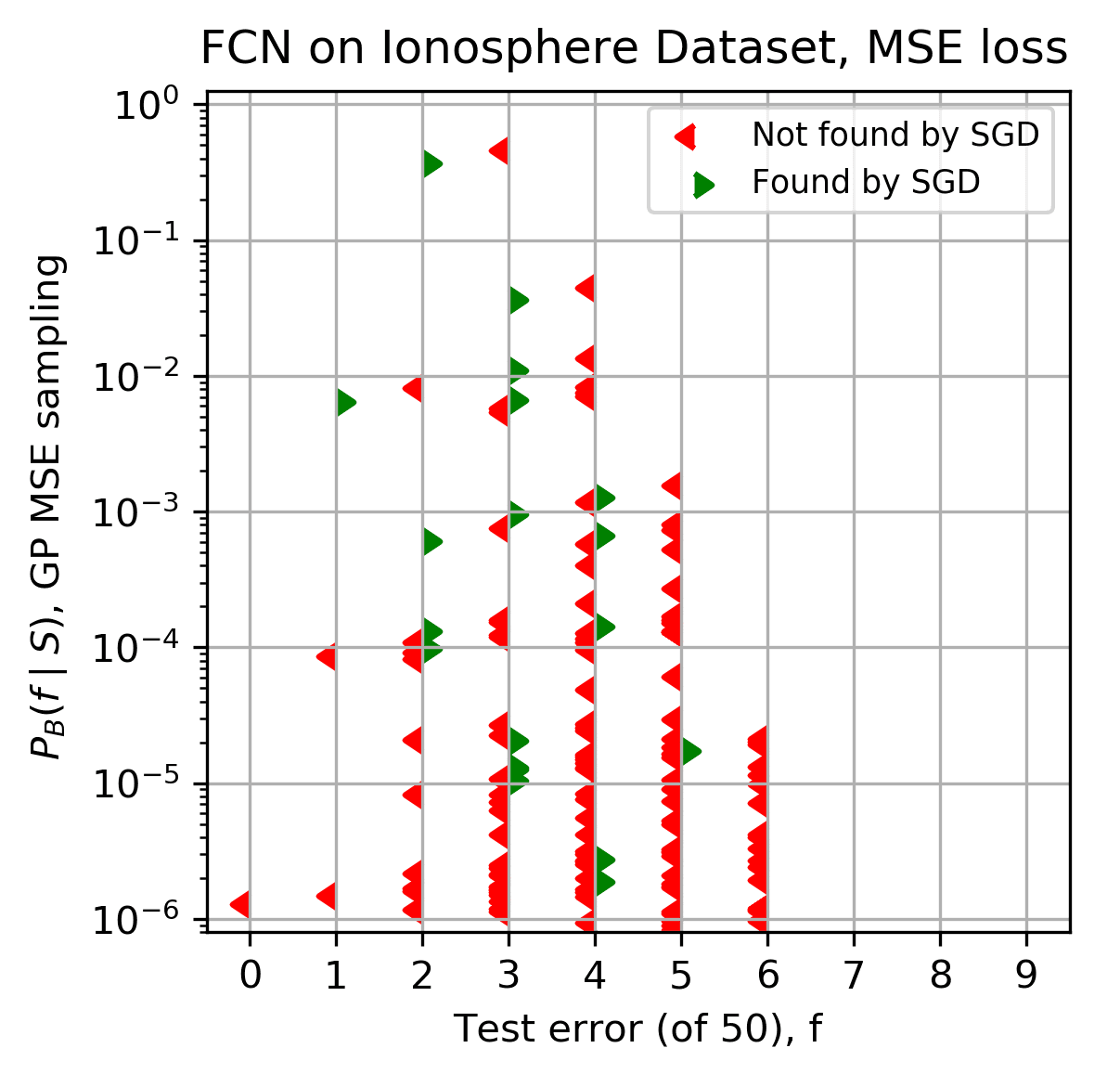}
    \caption{Functions found by GP}
    \label{fig:nonimage_ion_mse_gp_sampled}
\end{subfigure}

\begin{subfigure}[b]{0.4\textwidth}
    \includegraphics[width=\textwidth]{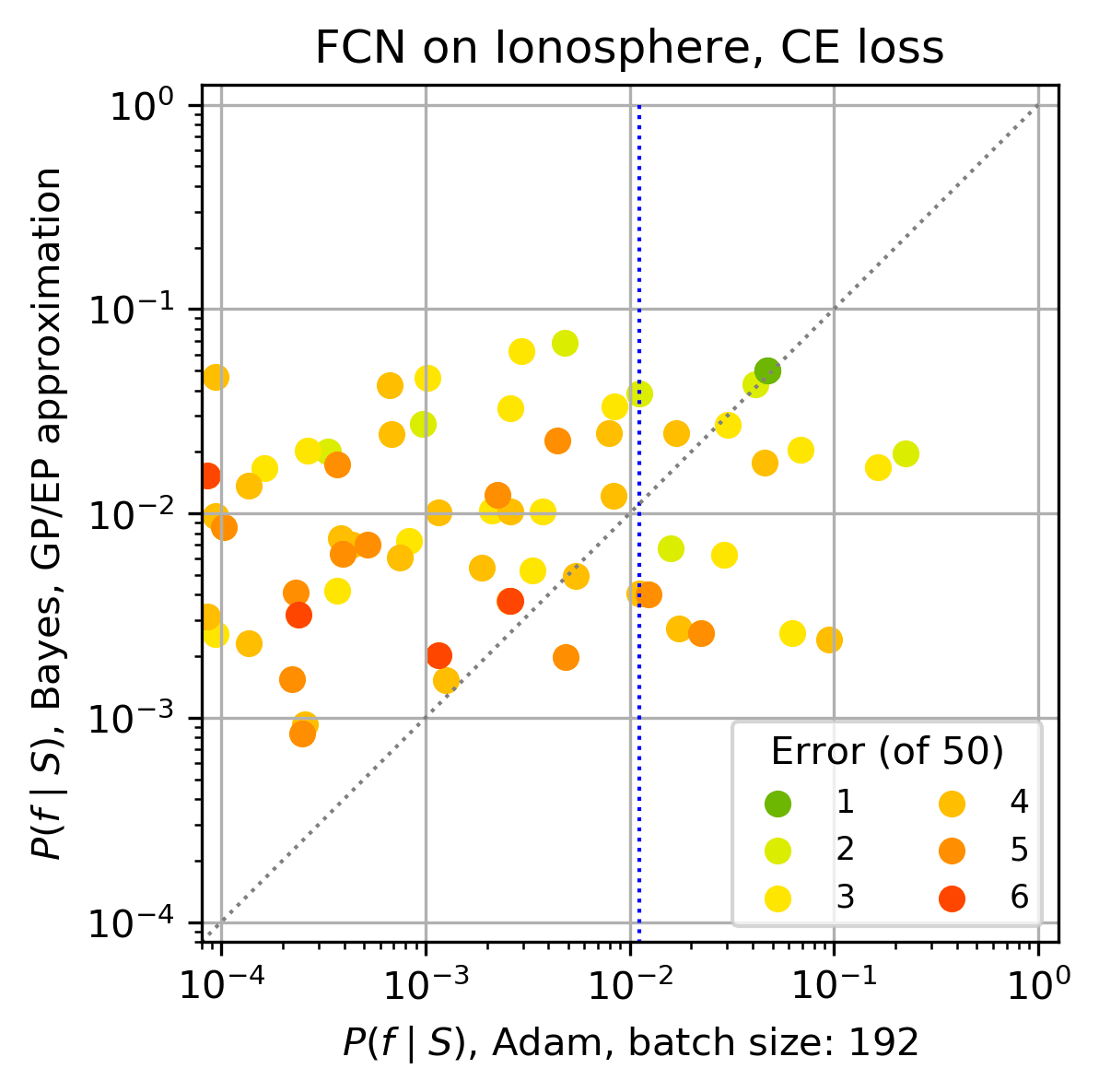}
    \caption{$\pb$ v.s.\ $\padam$}
    \label{fig:nonimage_ion_ce}
\end{subfigure}
~~
\begin{subfigure}[b]{0.4\textwidth}
	\includegraphics[width=\textwidth]{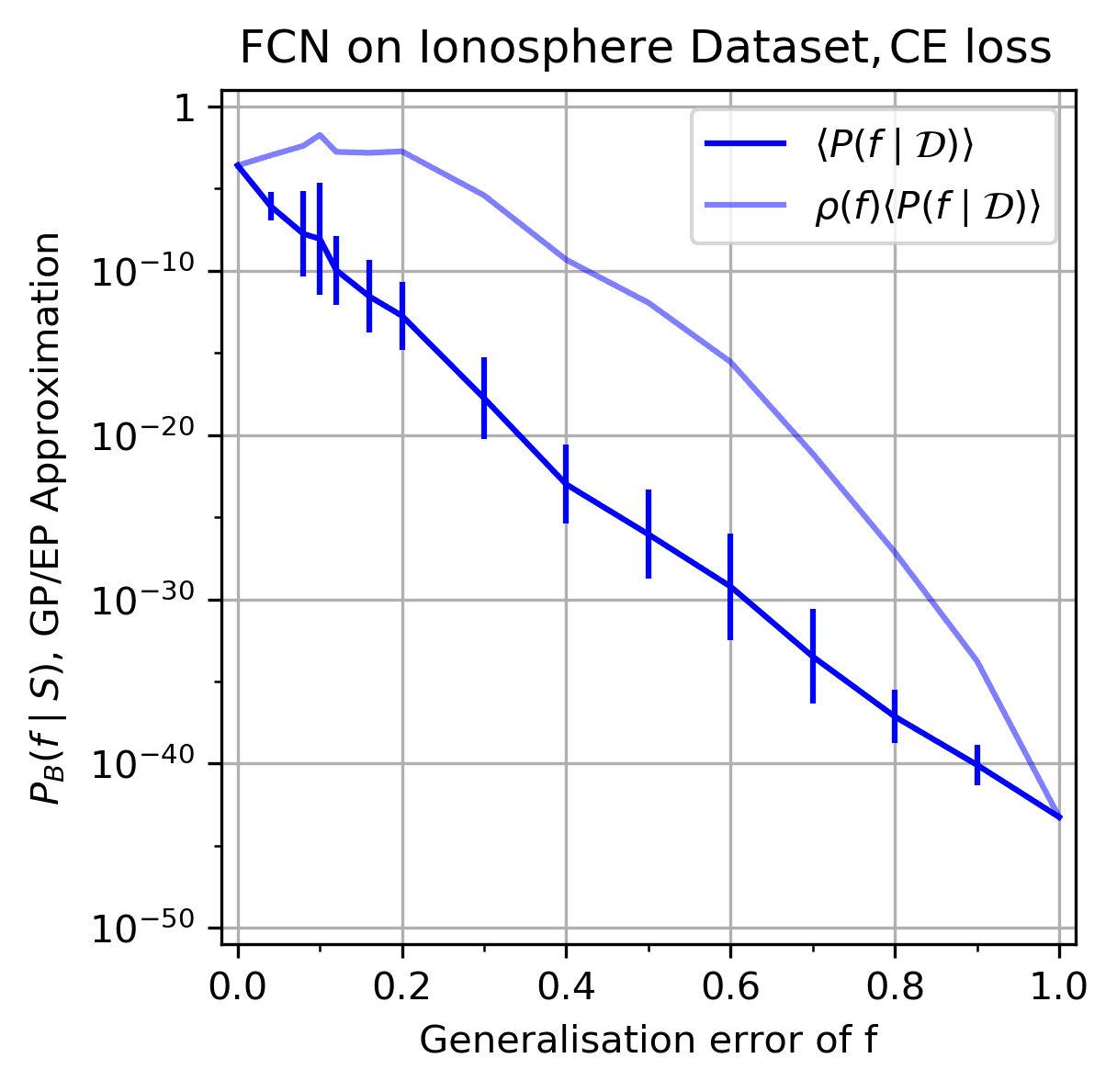}
    \caption{$\pb$ v.s.\ $\epsilon_G$}
    \label{fig:nonimage_ion_ce_gp}
\end{subfigure}

\caption{{\bf Comparing $\pb$ to $\padam$  for an FCN on the Ionosphere dataset with CE loss. Further results for \cref{fig:lstm_main}.}
 [We use training/test set size 301$/$50 and batch size=192. Vertical dotted blue lines denote $90 \%$ probability boundary; solid blue lines are fit to guide the eye; dashed grey line is $x=y$.] 
 \\ (a) $\pb$ v.s.\ $\padam$ for MSE (same as in \cref{fig:lstm_main}), put here ease of comparison). \\
 (b) Probability of all functions found by NNGP compared to those also found by the optimiser.  Green points are the set jointly found functions $F$.   $\sum_{f\in F}\pb=43.1\%$ and $\sum_{f\in F}\padam=99.8\%$ (in other words nearly all functions found by Adam are also found by the GP, but the GP also finds functions that Adam doesn't for this level of sampling).  For the optimiser, $\eg= 4.59\%$ and for the GP MSE sampling, $\eg_{GP}= 5.41\%.$ \\
 (c) $\pb$ v.s.\ $\padam$ for CE. $\eg=5.88\%$. \\
 (d) $<\pb>$ versus $\eg$ with CE, 20 samples per $\epsilon_G$.  The bias towards low error functions is less strong than what is found for MNIST or Fashion-MNIST.
}\label{fig:nonimage}
\end{figure}


\begin{figure}[H]
\centering

\begin{subfigure}[b]{0.4\textwidth}
    \includegraphics[width=\textwidth]{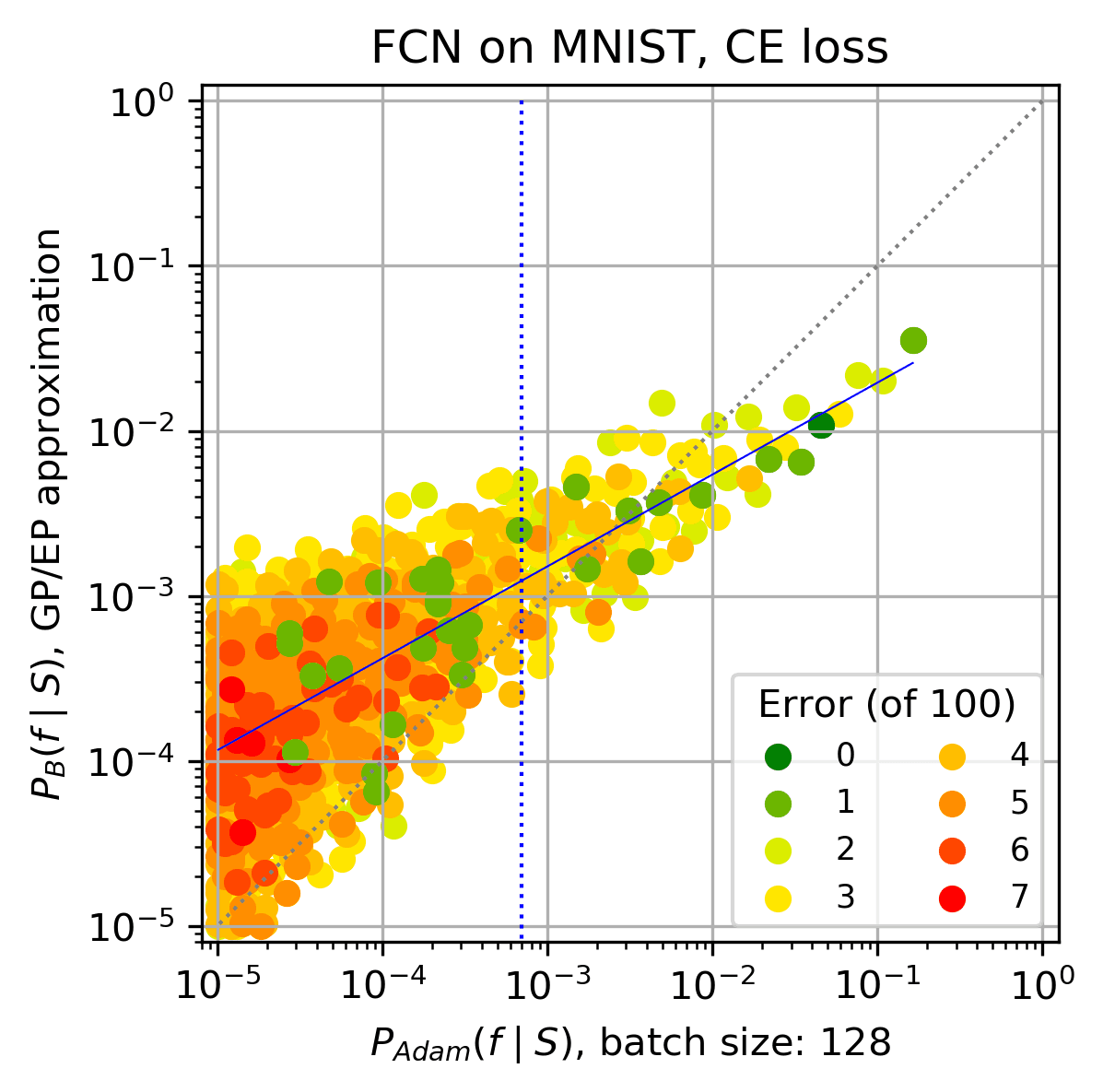}
    \caption{$\pb$ v.s.\ $\padam$}
    \label{fig:optvsopt_adam_app}
\end{subfigure}
~~
\begin{subfigure}[b]{0.4\textwidth}
	\includegraphics[width=\textwidth]{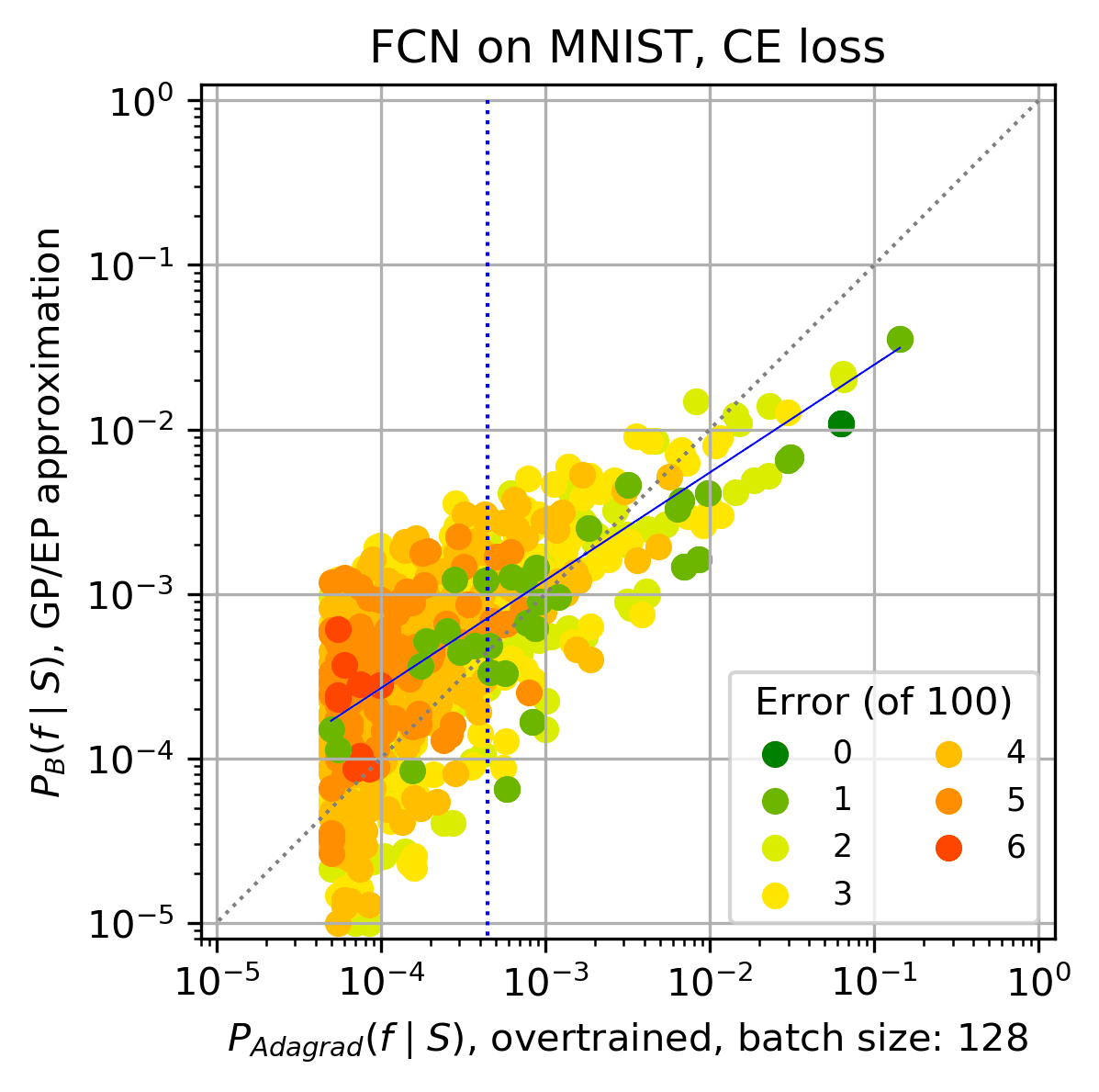}
    \caption{$\pb$ v.s.\ $\padagrad$}
    \label{fig:optvsopt_adagrad_o_app}
\end{subfigure}

\begin{subfigure}[b]{0.4\textwidth}
	\includegraphics[width=\textwidth]{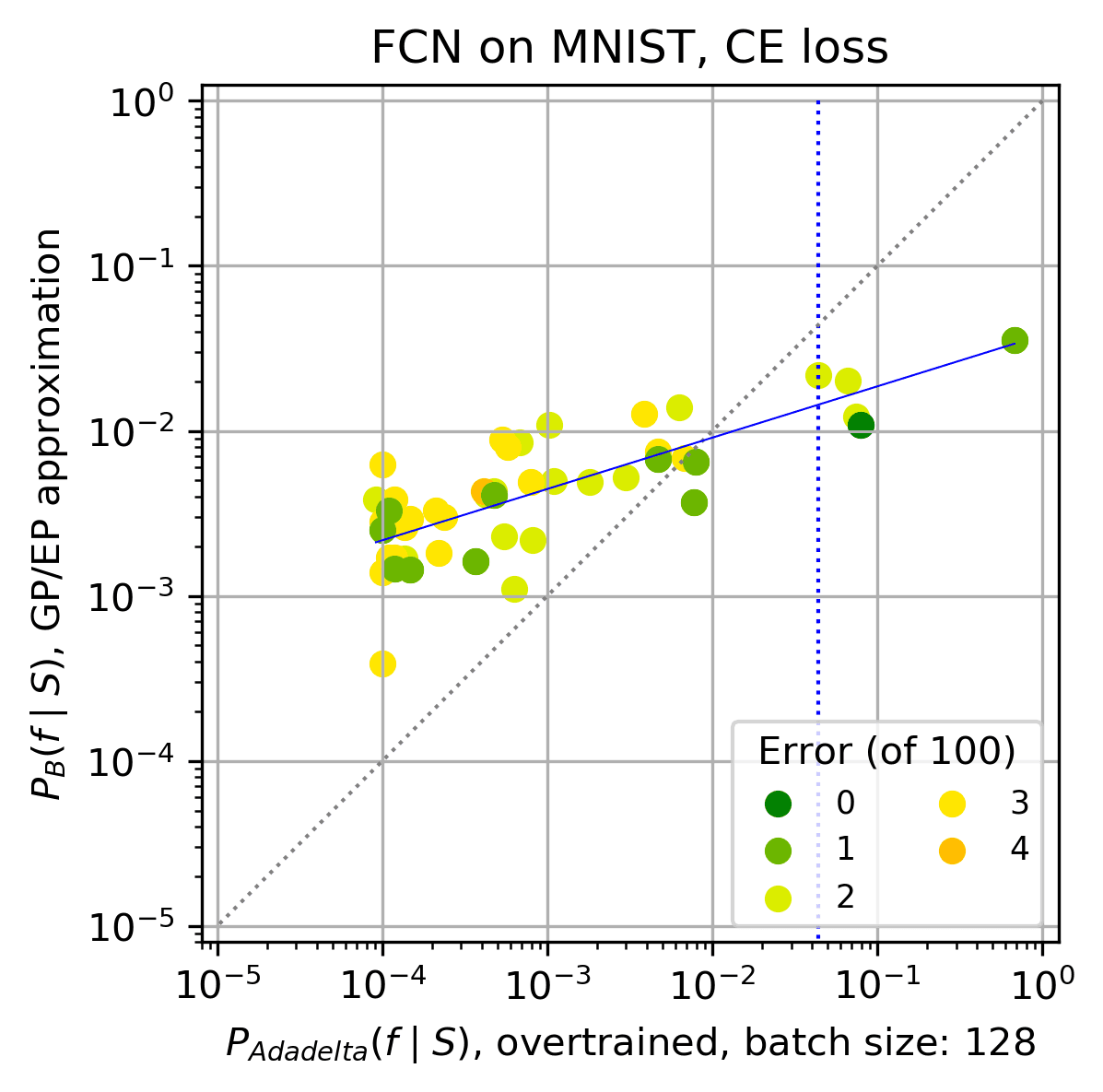}
   \caption{$\pb$ v.s.\ $\padadelta$}
    \label{fig:optvsopt_adadelta_o_app}
\end{subfigure}
~~
\begin{subfigure}[b]{0.4\textwidth}
	\includegraphics[width=\textwidth]{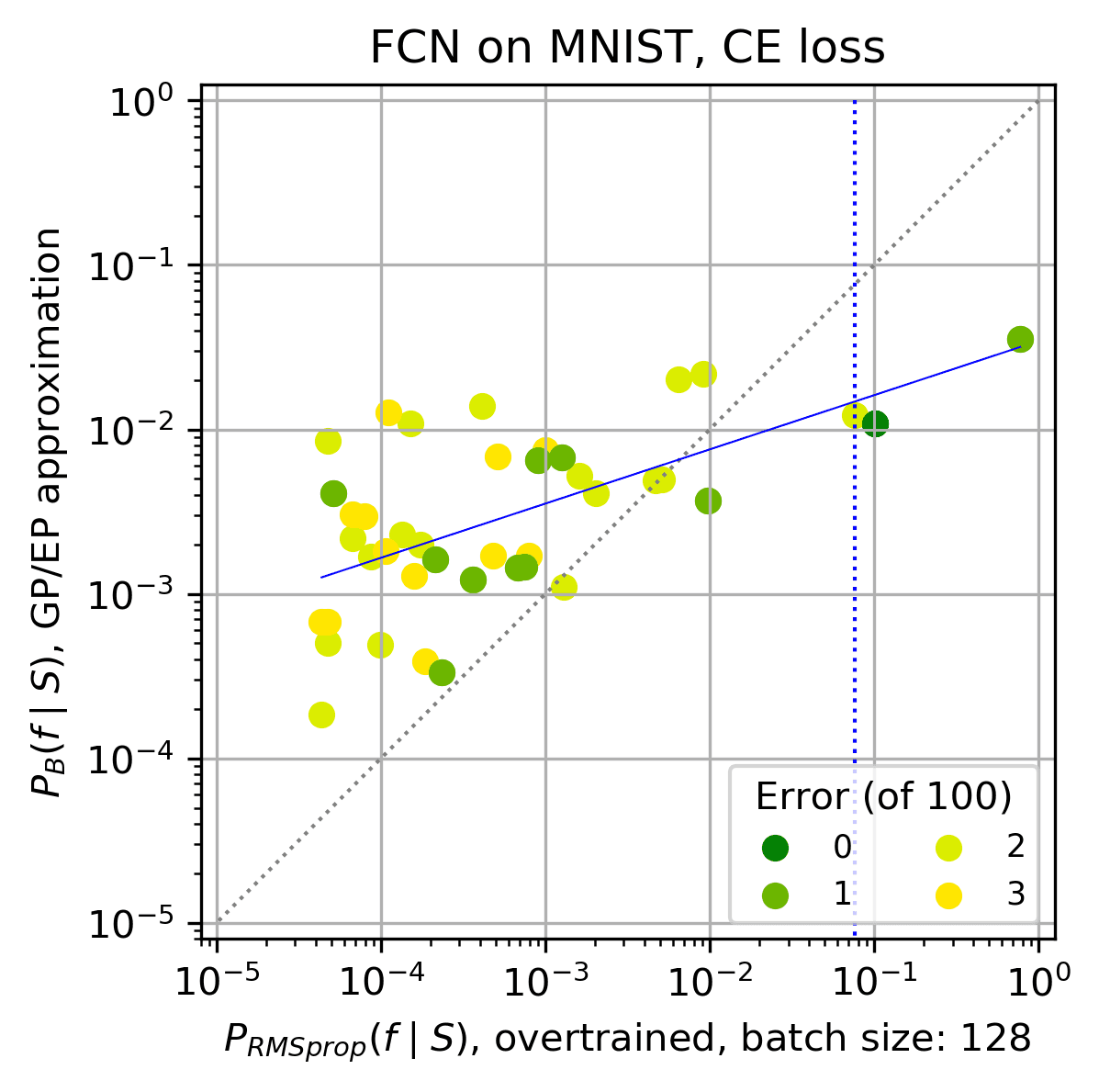}
    \caption{$\pb$ v.s.\ $\prmsprop$}
    \label{fig:optvsopt_RMS_o_app}
\end{subfigure}

\caption{{\bf Comparing  $\bf \pb$ to  $\bf \popt$ for an FCN on MNIST with CE loss for different optimisers. Further results for \cref{fig:optvsopt}.}  [We use training/test set size 10,000$/$100; batch size=128. Vertical dotted blue lines denote $90 \%$ probability boundary; solid blue lines are fit to guide the eye; dashed grey line is $x=y$.] \\
(a) Adam, no overtraining, \,\,\,\,\,\,\,\,\, $\eg=$ 2.20\%. Sample size: $n=10^6$.  \\
(b) Adagrad with overtraining,  $\eg=2.19\%$. Sample size: $n=2.1\times10^5$. \\ 
(c) Adadelta with overtraining,  $\eg=$ 1.17\%. Sample size: $n=10^5$.  \\
(d) RMSprop with overtraining,  $\eg=$ 1.01\%. Sample size: $n=2.5\times10^5$.
}
\label{fig:optvsopt_appendix}
\end{figure}

\begin{figure}[H]
\includegraphics[width=\textwidth]{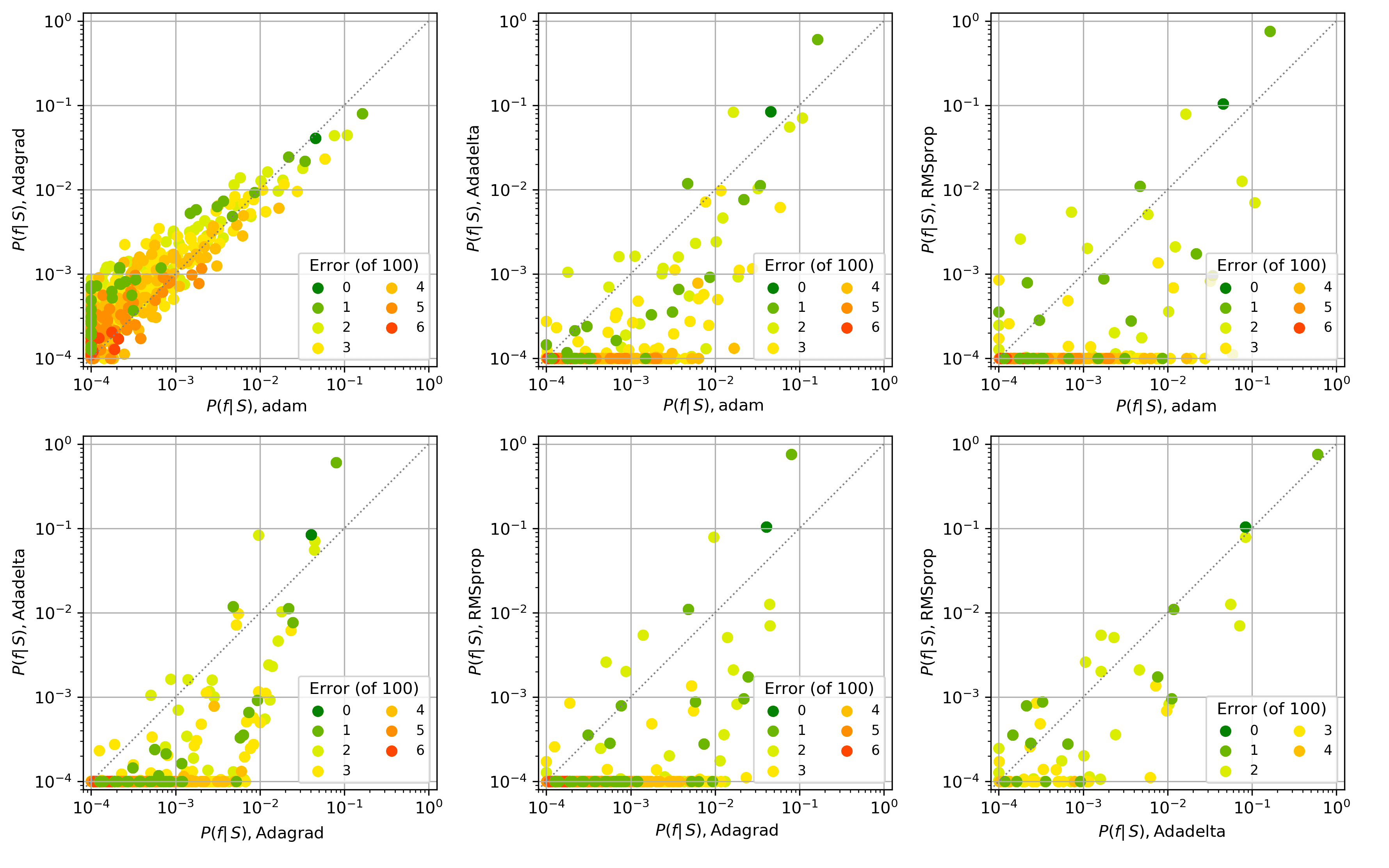}

\caption{{\bf Comparing $\popt$ with $\popt$ for different optimisers. Further results for \cref{fig:optvsopt}.} [We use the FCN architecture on MNIST, CE loss and a batch size of 128, training/test set size = 10,000$/$100, $y=x$ is denoted by a dashed line] \\
Adagrad and  Adam correlate very well as do Adadelta and RMSprop for these hyperparameters.  The other combinations do not correlate as well, suggesting that they sample the loss-function differently from one another.  
These plots do not rely on the GP or GP/EP approximation.  We believe that this sort of experiment may prove useful for understanding differences in the behaviour of the optimisers.}
\label{fig:4optcorr}
\end{figure}

\begin{figure}[H]
\centering

\begin{subfigure}[b]{0.4\textwidth}
    \includegraphics[width=\textwidth]{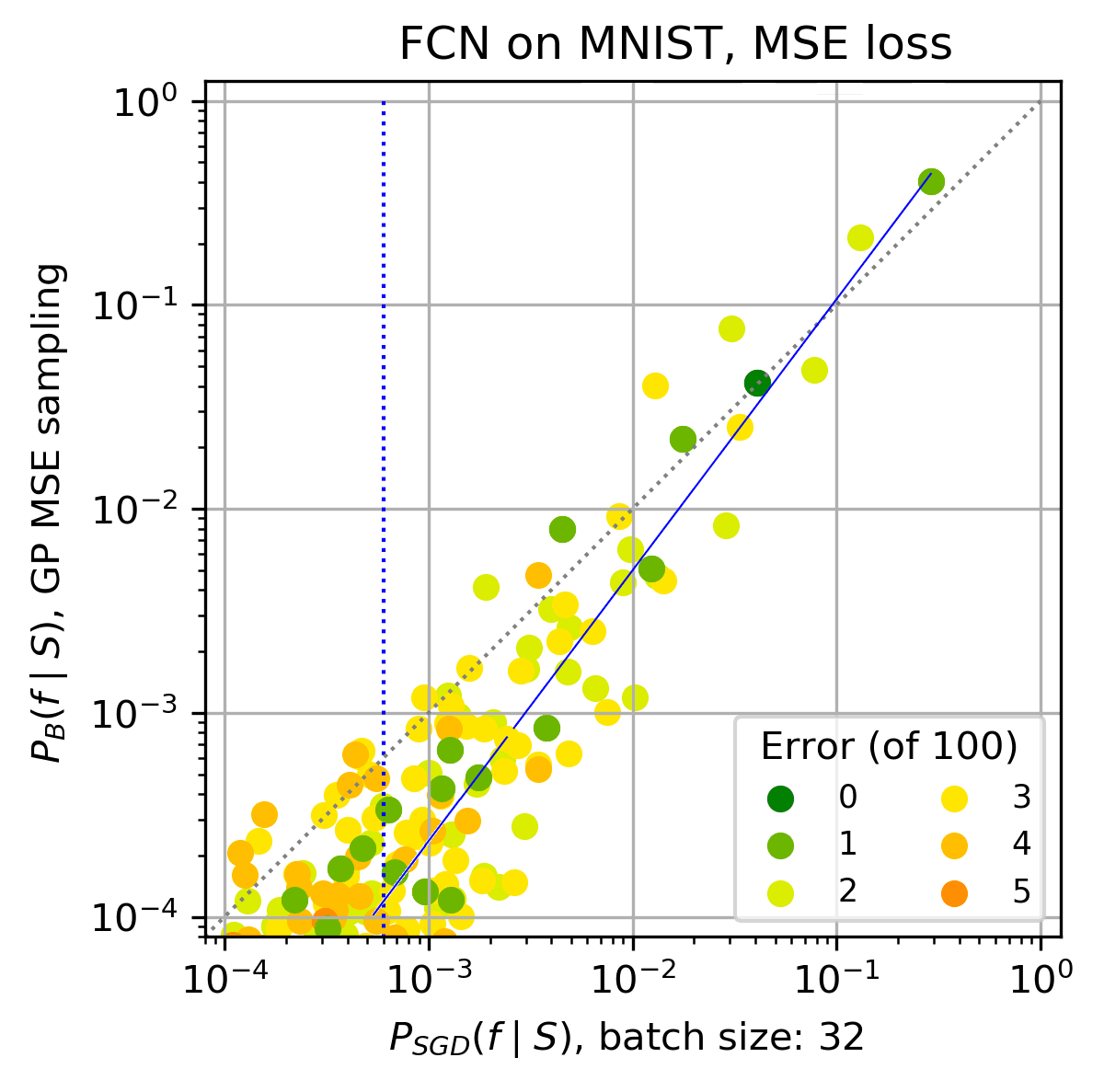}
    \caption{$\eg=$1.88\%, 0 Test error opt/gp: }
    \label{fig:batchvsbatchmse32}
\end{subfigure}
~~
\begin{subfigure}[b]{0.4\textwidth}
	\includegraphics[width=\textwidth]{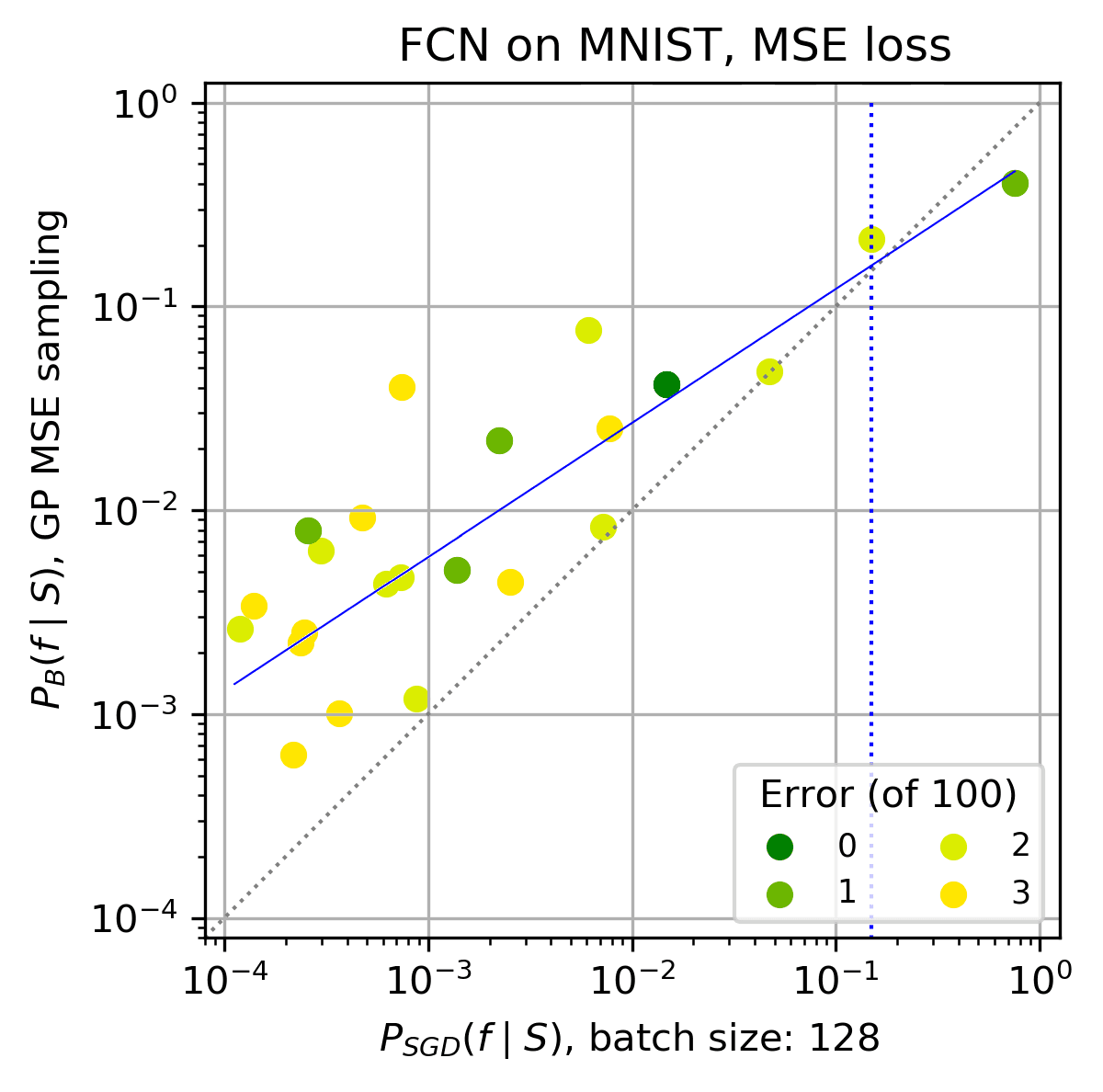}
    \caption{$\eg=$1.22\%, 0 Test error opt/gp:}
    \label{fig:batchvsbatchmse128}
\end{subfigure}

\caption{{\bf $\mathbf{\pb}$ v.s. $\mathbf{\padam}$ for an FCN on MNIST with the Adam optimiser and MSE loss with different batch sizes. Further results for \cref{sec:batch_sizes_main_text}.}
Batch sizes: (a) 32, and (b) 128. For this MSE loss function, we observe that increasing the batch size leads to better generalisation, and also to a shallower best fit because the highest probability functions are sampled with enhanced probability by SGD.  This trend with batch size is the opposite of what was observed for CE loss in  \cref{fig:sum_fig_batch}. }\label{fig:batchvsbatchmse}
\end{figure}

\begin{figure}[H]
\centering

\begin{subfigure}[b]{0.38\textwidth}
    \includegraphics[width=\textwidth]{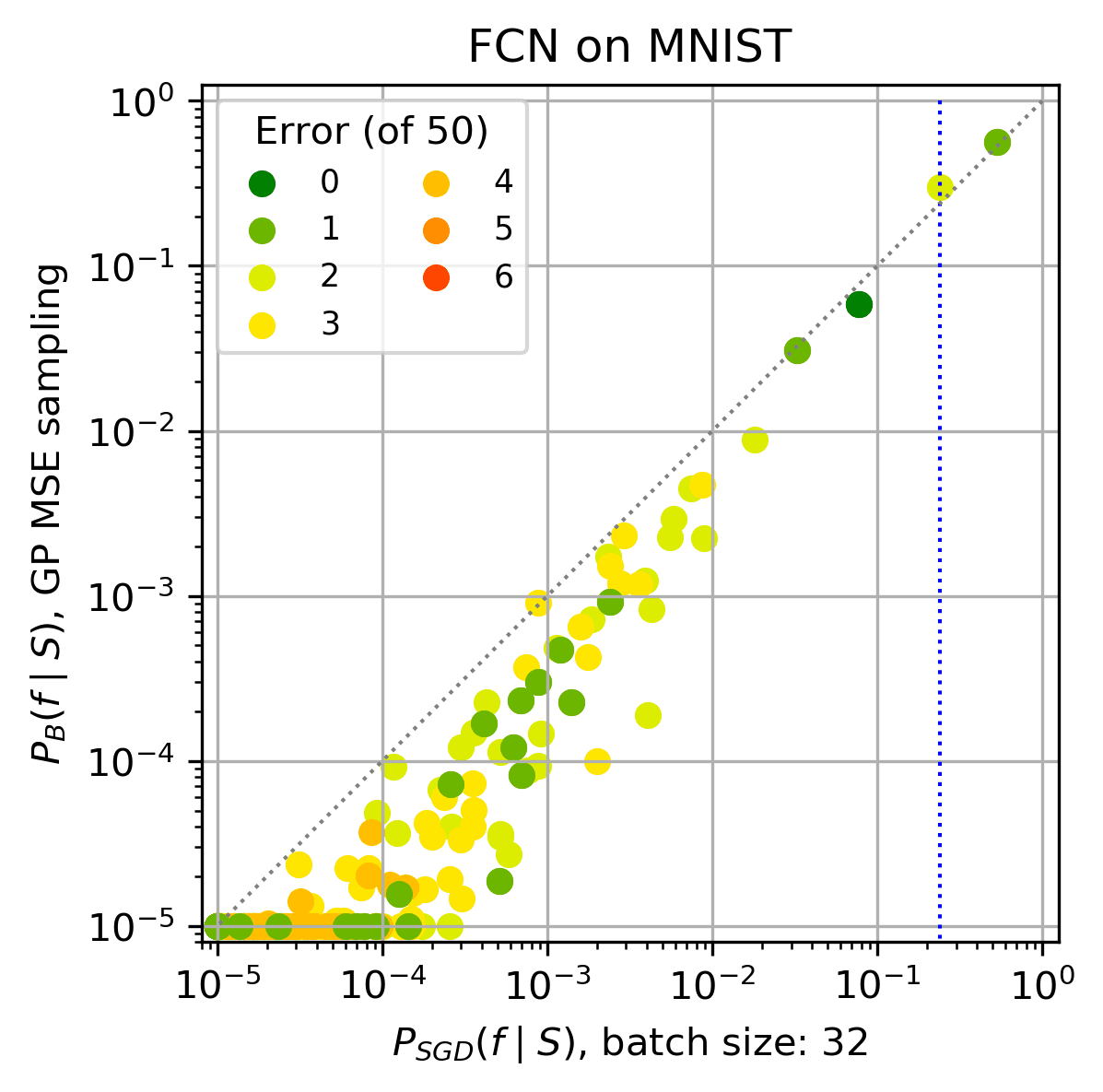}
    \caption{No label corruption}
    \label{sfig:4th_example_MSE_50}
\end{subfigure}
~~
\begin{subfigure}[b]{0.38\textwidth}
    \includegraphics[width=\textwidth]{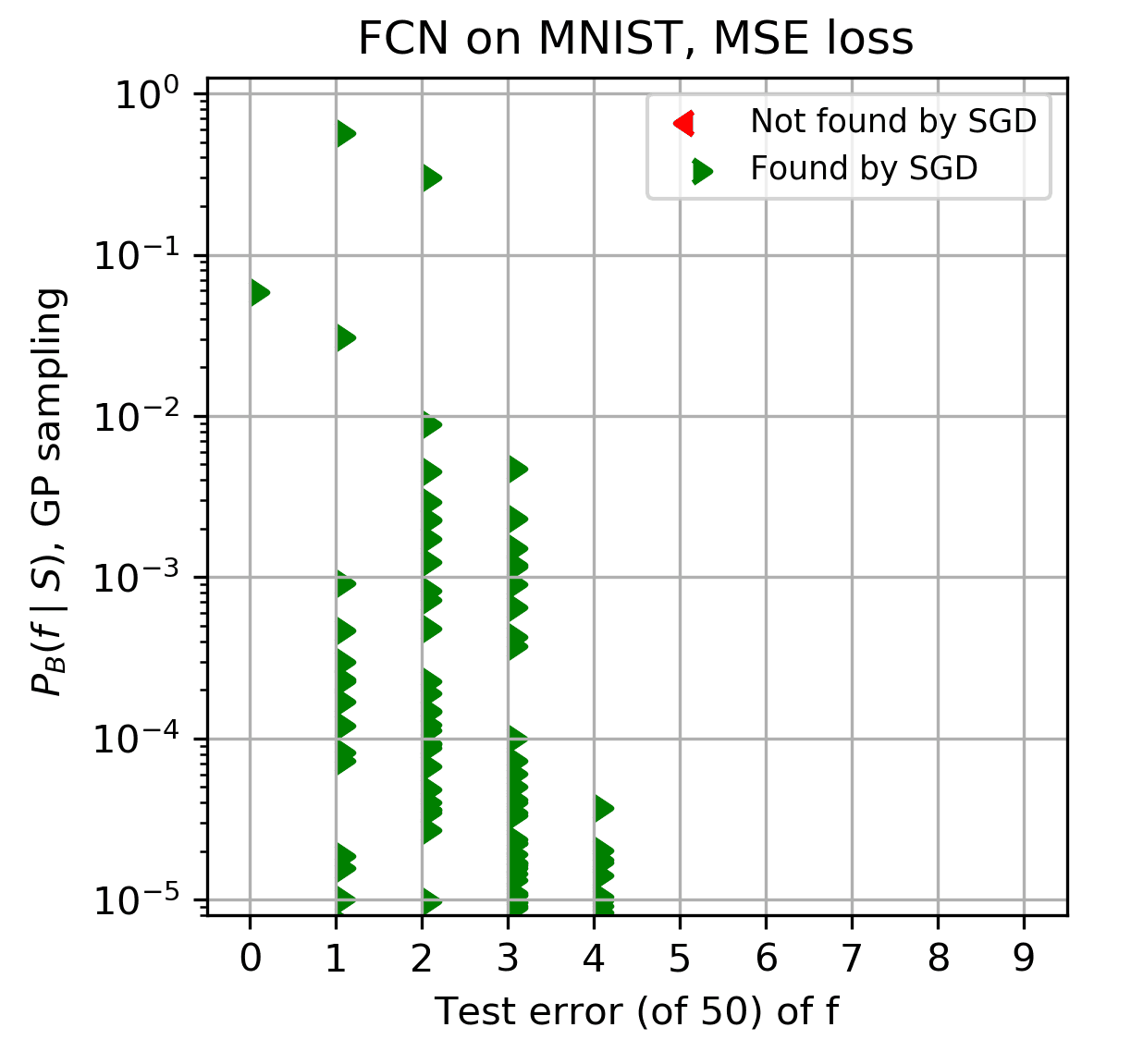}
    \caption{Functions found by GP in (a)}
    \label{sfig:4th_example_mse_50}
\end{subfigure}

\begin{subfigure}[b]{0.38\textwidth}
    \includegraphics[width=\textwidth]{fig/randomisation/sgdrand_4in.png}
    \caption{20\% label corruption.}
    \label{sfig:4th_example_MSE_20rand_50}
\end{subfigure}
~~
\begin{subfigure}[b]{0.38\textwidth}
    \includegraphics[width=\textwidth]{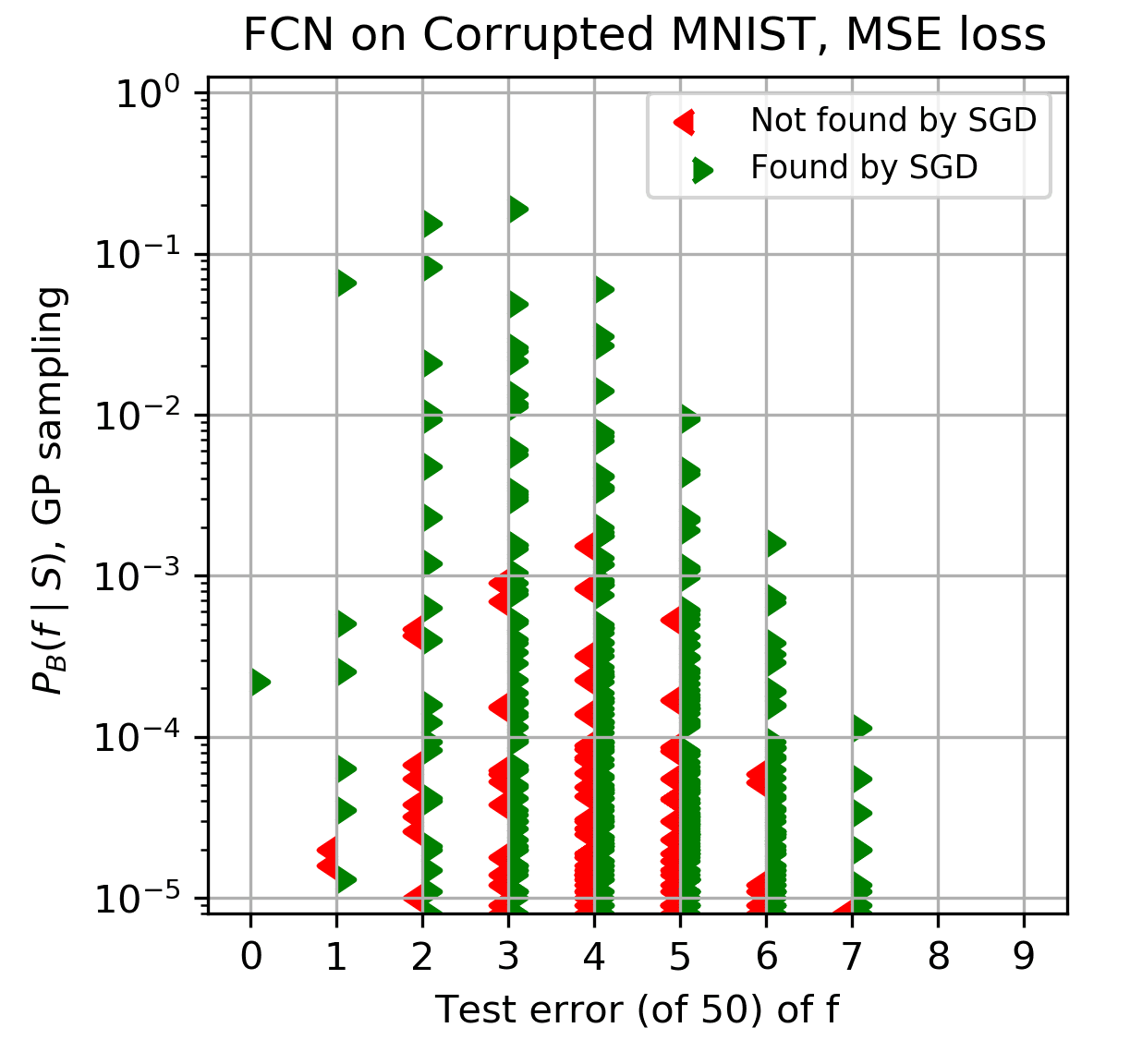}
    \caption{Functions found by GP in (c)}
    \label{sfig:4th_example_unclear}
\end{subfigure}

\caption{ {\bf Comparing $\psgd$ to $\pb$ for an FCN on MNIST with label corruption and MSE loss. Further results for \cref{fig:random}.}
[Training/test set size 10,000$/$50; batch size=128; vertical dotted blue lines denote $90 \%$ probability boundary; solid blue lines are fit to guide the eye; dashed grey line is $x=y$; points on the axes are found by one technique only.] \\
(a) $\pb$ v.s.\ $\psgd$ for no corruption,  $n=10^6$, $\eg=2.64\%$ for SGD and $\eg_{GP}=3.22\%$ for the GP. (Note this relative error is larger than for the $|E|=100$ test set because it includes a hard to classify image.  Such fluctuations are expected for small test sets.   \\
(b) Green dots denote functions in the set $F$, which are found by both SGD and the GP sampling.  Red dots are found by GP, but not by SGD. On this scale all GP functions are found by SGD. \\
(c) $\pb$ v.s. $\psgd$ for 20\% corruption (as \cref{fig:random}(c), but included for ease of comparison).
($\eg=13.4\%$ for SGD and $\eg_{GP}=5.80\%$ for the GP). Here we included functions with frequency $<10$. $n=10^5$ \\
(d) In contrast to (b), a considerable number of low probability functions are not found by SGD.  
 Here $\sum_{f\in F}\pb=99.3\%$, and $\sum_{f\in F}\psgd=24.3\%$, indicating that while SGD finds almost all functions with high $\pb$, it also finds many functions with low $\pb$. \\
 Comparing the $0\%$ and $20\%$ label corruption shows that the weaker bias in the latter  leads to less strong correlation between $\pb$ and $\psgd$.
}\label{fig:appendix_corrupted_mnist}
\end{figure}

\section{Critical Sample Ratio}\label{app:CSR}

A measure of the complexity of a function, the critical sample ratio (CSR), was introduced in \citep{krueger2017deep}.  It is defined with respect to a sample of inputs as the fraction of those samples which are critical samples, defined to be an input such that there is another input within a box of side $2r$ centred around the input, producing a different output (for discrete outputs).

\begin{figure}[H]
\centering

\begin{subfigure}[b]{0.4\textwidth}
    \includegraphics[width=\textwidth]{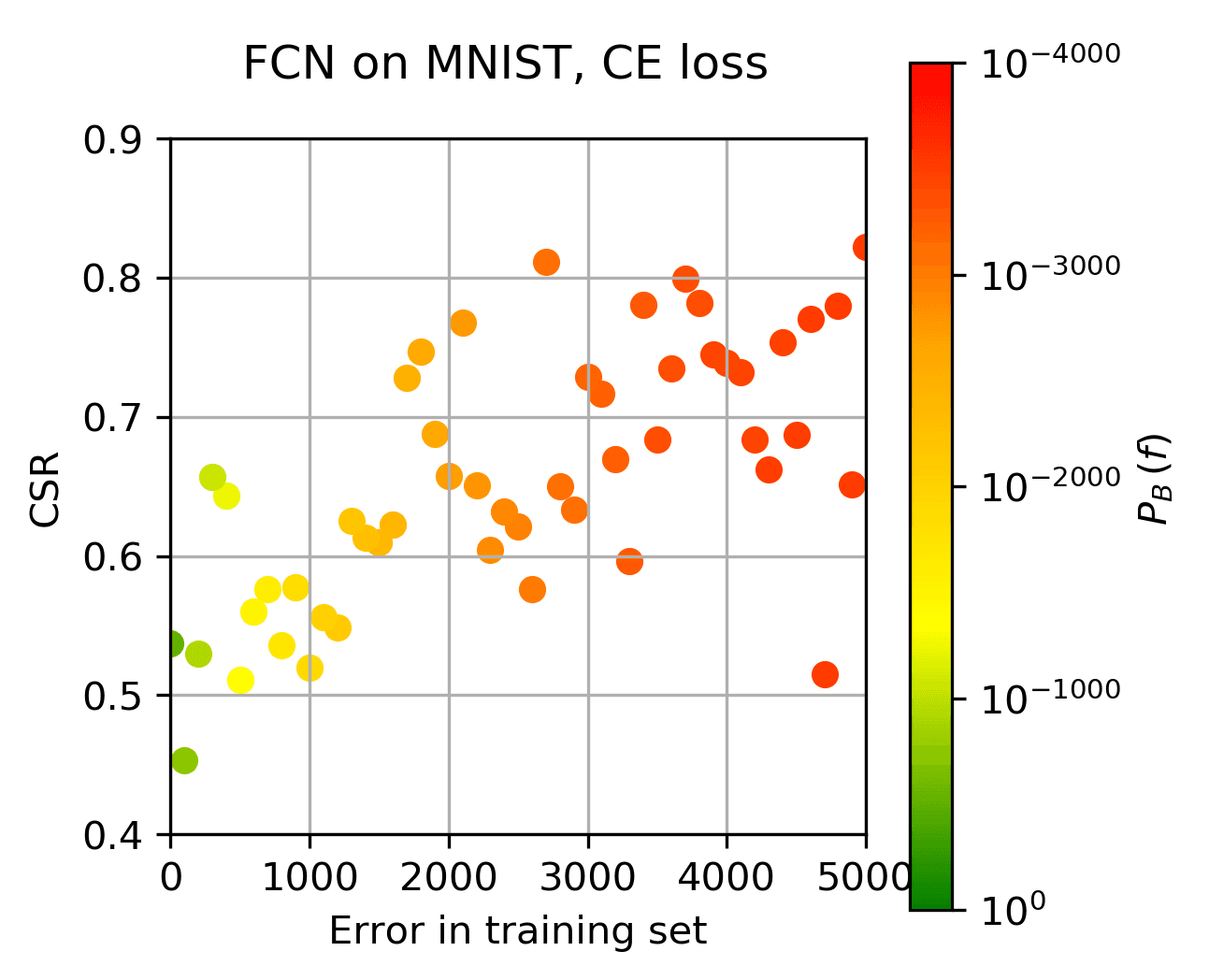}
    \caption{10 samples per $\langle \epsilon \rangle$}
    \label{fig:CSR_prior}
\end{subfigure}
~~
\begin{subfigure}[b]{0.4\textwidth}
	\includegraphics[width=\textwidth]{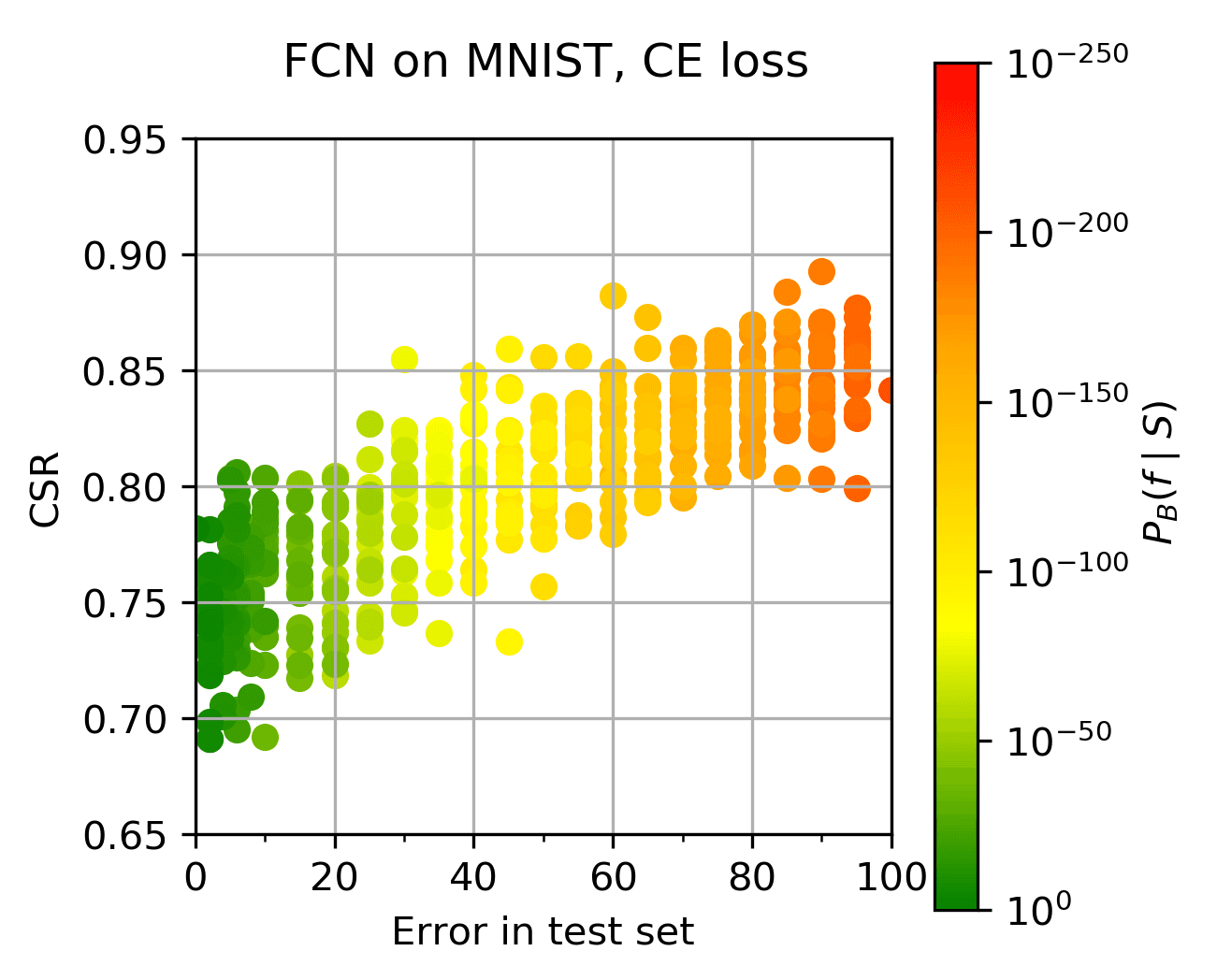}
    \caption{20 samples per $\langle \epsilon \rangle$}
    \label{fig:CSR_posterior}
\end{subfigure}

\caption{(a) shows the result of our experiment on the functions in the prior; (b) shows the result out our experiment on functions in the posterior. Clearly there is a strong correlation between the CSR complexity and the error of the function; between the CSR complexity and $\pb$; and between the error and $\pb$. Different hyperparameters were used in the experiment to obtain a suitable range of complexity.}
\label{fig:CSR}
\end{figure}

Following~\citep{valle2018deep}, we use CSR as a rough estimate of the complexity of functions found in the posterior (conditioning on $S$), and the prior (i.e.\ functions on $S$). In our experiments, we used binarised MNIST with a training set size of $10000$ and a test set of size $100$ (analogously to the majority of our other experiments).
For the prior, \cref{fig:CSR_prior}, we randomly generated 100 functions with errors ranging between 0 and 10000 on $S$. For each function, we recorded the error, $\pb$ and the CSR.
For the posterior, we generated 500 functions with a range of errors on the test set and concatenated them with the function correct on the training set. We then proceeded as with the prior.

To calculate the CSR, we trained a DNN to model the function in question. Clearly this induces effects not purely due to the parameter-function map -- although as we have seen in \cref{app:exampleresults}, the functions found by SGD are likely to be similar to those that would be found by training by random sampling of parameters. Therefore, we may expect this process to approximately give the average CSR of parameters producing the function of interest. In Figure 2a. of \citep{valle2018deep}, an example of a very similar experiment with CIFAR10 can be found, where the network was not trained. This produces very similar results to our experiments with network training.

\section{Bayesian inference by direct sampling on Boolean system.}\label{app:bool_func}

In this section we consider a much simpler system, with a smaller input space, and thus a much smaller space of functions. This  allows us to approximate Bayesian inference in the case of 0-1 likelihood (see \cref{app:GP_exp_0-1}) much more accurately, via direct sampling. We use the same DNN architecture and synthetic data studied in \citep{valle2018deep}. It consists of a two layer neural network with 7 Boolean inputs, two hidden layers of 40 ReLU-activated neurons, and a single Boolean output. The space of functions is thus the space  of $2^{128} \approx 3.4 \times 10^{38}$  Boolean functions of 7 inputs.

We perform `approximate Bayesian inference' (ABI) by sampling the parameters of the neural network i.i.d. from a Gaussian distribution with distribution parameters\footnote{Remember that, following standard convention, the actual weight variance is dividing by the number of input neurons} $\sigma_w=\sigma_b=1.0$, evaluating the neural network on the training set, and saving the samples for which the neural network achieves 100\% training accuracy. Each of these samples corresponds to a function, \textbf{sampled from the \emph{exact} Bayesian posterior}. We estimate the posterior probabilities by the empirical frequencies of individual functions (defined on all $2^7=128$ inputs), after sampling the parameters $10^{10}$ times. We used a random but fixed training set consisting of $32$ out of the $128$ inputs for a given target function.  Target functions were chosen among functions that appeared with reasonably high frequency, in a large sample obtained by randomly sampling the weights of the neural network, so as to ensure ABI would give enough samples. They were chosen to have a range of values of Lempel-Ziv complexity. See \citep{valle2018deep} for the definition of Lempel-Ziv complexity of Boolean functions used here.

Representative results are shown in Figures~\ref{boolean0},\ref{boolean2},\ref{boolean6} (results for the other 4 functions we tested look qualitatively similar). We empirically found that the ABI probabilities correlated and are of a similar order of magnitude to the SGD probabilities over the whole range of Boolean functions tried. SGD is consistently more biased towards the most likely functions for CE loss, although it only increased their probability by about a factor of two (a small amount relative to the whole range of probabilities, but because for this system this is the dominant function, this secondary effect can still have a significant effect on the average generalisation error). We also performed sampling using the EP approximation to the posterior with 0-1 loss, and the exact posterior using MSE loss to directly see the effects coming from the EP approximation. We found that for some functions (the simplest ones) GP/EP gave probabilities which were close to those found by ABI. However, for more complex functions\footnote{These are still rather simple w.r.t.\ the full range of possibilities, necessary to ensure that ABI sampling is feasible.}  GP/EP highly underestimates the probabilities In fact, for the most complex functions we studied, GP/EP didn't find a single function more than once in our sampling. This is in contrast to the GP/MSE sampling probabilities, which shows reasonable correlation with the ABI probabilities, as well as with the probabilities of SGD trained with MSE loss. 




These results support the main hypothesis of the paper that the Bayesian posterior probabilities correlate with the SGD probabilities. 
They also suggest that the EP approximation can sometimes heavily underestimate the probabilities. This agrees with the conjecture that the EP approximation is the main cause of the discrepancy in the magnitudes of the probabilities between GP/EP and SGD observed in the rest of the experiments in this paper, and that for CE loss, the true Bayesian prior probabilities may in fact match the SGD probabilities a lot more closely. Furthermore, the MSE results suggest that MSE may give a good approximation to the Bayesian posterior probabilities (even the ones based on 0-1 likelihood). However, we note that this is a small toy model, and so our analysis leaves as an open question how to understand the error induced by the EP approximation in more realistic systems. One approach could be to cross-validate some of our results with state-of-the-art Monte Carlo sampling techniques for Bayesian inference \citep{rasmussen2004gaussian}.




\textbf{Description of figures}. In Figures \ref{boolean0},\ref{boolean2} and \ref{boolean6} in this section, we show, for three representative target Boolean functions, data comparing $\psgd$ for CE or MSE loss versus different ways to estimate $\pb$, namely ABI, GP/EP for CE loss and GP/MSE for MSE loss.

In the \textbf{first column}, we show scatter plots comparing sampled probabilities (where functions not found in the sample are shown as if having a frequency of one).  The colours denote the number of errors for each function on the test set.

The \textbf{second column} shows probability versus rank of the different test-set functions (when ranked by probability) for the two sampling methods.

The \textbf{third column} shows  test accuracy histograms for the two sampling methods.

In the \textbf{first row}, we use sampling of parameters (and 0-1 loss) to estimate $\pb$, which we use as the gold standard method as it has controlled small errors.

In the \textbf{second row}, we estimate $\pb$ with the GP/EP approximation introduced in \cref{app:GP_exp_0-1}.

In the \textbf{third row}, we compare $\pb$ estimated from sampling of the exact MSE posterior (explained in \cref{app:GP_exaplanation:MSE}) versus the ABI sampling (for 0-1 loss).

In the \textbf{fourth row}, we compare $\psgd$ when training with MSE loss versus ABI sampling (for 0-1 loss). 


\foreach \n in {0,2,6} {
\begin{figure}[H]
\centering
\begin{subfigure}[b]{0.31\textwidth}
    \includegraphics[width=\textwidth]{fig/boolean/abi_sgd/\n_scatter_ABI_vs_SGD_0__1.png}
    \label{fig:\n_prob_scatter}
\end{subfigure}
~~
\begin{subfigure}[b]{0.31\textwidth}
    \includegraphics[width=\textwidth]{fig/boolean/abi_sgd/\n_ABI_vs_SGD_rank_plots_fc_32_0___1_1_7_2x40_1.png}
    \label{fig:\n_rank_plot_abisgd}
\end{subfigure}
~~
\begin{subfigure}[b]{0.31\textwidth}
    \includegraphics[width=\textwidth]{fig/boolean/abi_sgd/\n_ABI_vs_SGD_test_error_histograms_fc_32_0___1_1_7_2x40_1.png}
    \label{fig:\n_error_hist_abisgd}
\end{subfigure}

\begin{subfigure}[b]{0.31\textwidth}
    \includegraphics[width=\textwidth]{fig/boolean/gpep_sgd/\n_scatter_GP_EP_vs_SGD_0__1.png}
    \label{fig:\n_prob_scatter_gpep}
\end{subfigure}
~~
\begin{subfigure}[b]{0.31\textwidth}
    \includegraphics[width=\textwidth]{fig/boolean/gpep_sgd/\n_GP_EP_vs_SGD_rank_plots_fc_32_0___1_1_7_2x40_1.png}
    \label{fig:\n_rank_plot_gpep}
\end{subfigure}
~~
\begin{subfigure}[b]{0.31\textwidth}
    \includegraphics[width=\textwidth]{fig/boolean/gpep_sgd/\n_GP_EP_vs_SGD_test_error_histograms_fc_32_0___1_1_7_2x40_1.png}
    \label{fig:\n_error_hist_gpep}
\end{subfigure}

\begin{subfigure}[b]{0.31\textwidth}
    \includegraphics[width=\textwidth]{fig/boolean/msesample_abi/\n_scatter_ABI_vs_GP_MSE_0__0.001_1_gpmse.png}
    \label{fig:\n_prob_scatter_mse_abi}
\end{subfigure}
~~
\begin{subfigure}[b]{0.31\textwidth}
    \includegraphics[width=\textwidth]{fig/boolean/msesample_abi/\n_ABI_vs_GP_MSE_rank_plots_fc_32_0__0.001__1_1_7_2x40_1_gpmse.png}
    \label{fig:\n_rank_plot_mseabi}
\end{subfigure}
~~
\begin{subfigure}[b]{0.31\textwidth}
    \includegraphics[width=\textwidth]{fig/boolean/msesample_abi/\n_ABI_vs_GP_MSE_test_error_histograms_fc_32_0__0.001__1_1_7_2x40_1_gpmse.png}
    \label{fig:\n_error_hist_mseabi}
\end{subfigure}

\begin{subfigure}[b]{0.31\textwidth}
    \includegraphics[width=\textwidth]{fig/boolean/msesgd_msesample/\n_scatter_SGD_vs_GP_MSE_0__0.001_1_mse.png}
    \label{fig:\n_prob_scatter_mse_mse}
\end{subfigure}
~~
\begin{subfigure}[b]{0.31\textwidth}
    \includegraphics[width=\textwidth]{fig/boolean/msesgd_msesample/\n_SGD_vs_GP_MSE_rank_plots_fc_32_0__0.001__1_1_7_2x40_1_mse.png}
    \label{fig:\n_rank_plot_mse_mse}
\end{subfigure}
~~
\begin{subfigure}[b]{0.31\textwidth}
    \includegraphics[width=\textwidth]{fig/boolean/msesgd_msesample/\n_SGD_vs_GP_MSE_test_error_histograms_fc_32_0__0.001__1_1_7_2x40_1_mse.png}
    \label{fig:\n_error_hist_mse_mse}
\end{subfigure}
\ifnum\n=0 \caption{\label{boolean0}Results for \cref{app:bool_func} for target function with LZ complexity 35.0. Refer to text for detailed description.}\fi
\ifnum\n=1 \caption{\label{boolean1}Results for \cref{app:bool_func} for target function with LZ complexity 38.5. Refer to text for detailed description.}\fi
\ifnum\n=2 \caption{\label{boolean2}Results for \cref{app:bool_func} for target function with LZ complexity 28.0. Refer to text for detailed description.}\fi
\ifnum\n=3 \caption{\label{boolean3}Results for \cref{app:bool_func} for target function with LZ complexity 31.5. Refer to text for detailed description.}\fi
\ifnum\n=4 \caption{\label{boolean4}Results for \cref{app:bool_func} for target function with LZ complexity 42.0. Refer to text for detailed description.}\fi
\ifnum\n=5 \caption{\label{boolean5}Results for \cref{app:bool_func} for target function with LZ complexity 31.5. Refer to text for detailed description.}\fi
\ifnum\n=6 \caption{\label{boolean6}Results for \cref{app:bool_func} for a (different) target function of LZ complexity 28.0. Refer to text for detailed description.}\fi
\end{figure}
}